\documentclass{article}

\usepackage{arxiv}

\usepackage{cite}

\usepackage{amsmath}
\DeclareMathOperator*{\argmax}{arg\,max}

\DeclareMathOperator*{\pluseq}{+\!\!=}

\usepackage{url}

\usepackage{amsmath}
\usepackage[utf8]{inputenc} 
\usepackage[T1]{fontenc}    
\usepackage[nolinks=true]{hyperref}       
\usepackage{url}            
\usepackage{booktabs}       
\usepackage{amsfonts}       
\usepackage{nicefrac}       
\usepackage{microtype}      
\usepackage{lipsum}		
\usepackage{enumitem} 
\setlist[itemize]{noitemsep, topsep=0pt}
\usepackage[nolinks=true]{hyperref}
\usepackage{array}
\usepackage{subfig}
\usepackage{amsmath}
\usepackage{bbm}
\usepackage{tikz}
\usetikzlibrary{decorations.pathreplacing}
\usepackage{siunitx}
\usepackage{algorithm}
\usepackage{algpseudocode}
\usepackage{todonotes}
\usepackage{array}

\algnewcommand{\LineComment}[1]{\State \(\triangleright\) #1}
\newcolumntype{C}[1]{>{\centering\arraybackslash}m{#1}}

\usepackage{mathrsfs} 
\usetikzlibrary{3d,decorations.markings,calc}
\usepackage{tikz-3dplot}
\makeatletter 
\tikzoption{canvas is xy plane at z}[]{%
  \def\tikz@plane@origin{\pgfpointxyz{0}{0}{#1}}%
  \def\tikz@plane@x{\pgfpointxyz{1}{0}{#1}}%
  \def\tikz@plane@y{\pgfpointxyz{0}{1}{#1}}%
  \tikz@canvas@is@plane
}

\newsavebox{\algbox}
\usepackage{varwidth}

\begin{document}

\title{Quality-Diversity Meta-Evolution: customising behaviour spaces to a meta-objective}

\author{David M. Bossens and Danesh Tarapore$^{}$
\thanks{$^{}$Authors are with the School of Electronics and Computer Science, University of Southampton, SO17 1BJ Southampton, U.K.
        {\tt\small d.m.bossens@soton.ac.uk}}%
}

\maketitle

\begin{abstract}
Quality-Diversity (QD) algorithms evolve behaviourally diverse and high-performing solutions. To illuminate the elite solutions for a space of behaviours, QD algorithms require the definition of a suitable behaviour space. If the behaviour space is high-dimensional, a suitable dimensionality reduction technique is required to maintain a limited number of behavioural niches. While current methodologies for automated behaviour spaces focus on changing the geometry or on unsupervised learning, there remains a need for customising behavioural diversity to a particular meta-objective specified by the end-user. In the newly emerging framework of QD Meta-Evolution, or QD-Meta for short, one evolves a population of QD algorithms, each with different algorithmic and representational characteristics, to optimise the algorithms and their resulting archives to a user-defined meta-objective. Despite promising results compared to traditional QD algorithms, QD-Meta has yet to be compared to state-of-the-art behaviour space automation methods such as Centroidal Voronoi Tessellations Multi-dimensional Archive of Phenotypic Elites (CVT-MAP-Elites) and Autonomous Robots Realising their Abilities (AURORA). This paper performs an empirical study of QD-Meta on function optimisation and multilegged robot locomotion benchmarks. Results demonstrate that QD-Meta archives provide improved average performance and faster adaptation to a priori unknown changes to the environment when compared to CVT-MAP-Elites and AURORA. A qualitative analysis shows how the resulting archives are tailored to the meta-objectives provided by the end-user.
\end{abstract}


\section{Introduction}
Historically, most evolutionary algorithms (EAs) optimised a fitness function without considerations for generalisation to unseen problems or robustness to perturbations to the evaluation environment. However, it was widely known that successfully converging to the maximum of that fitness function requires maintaining genetic diversity in the population of solutions (e.g., \cite{Laumanns2002,Gupta2012,Ursem2002,Ginley2011}). Moreover, the use of niching demonstrated how  maintaining subpopulations could help find multiple solutions to a single problem \cite{Mahfoud1995}. Some studies included genetic diversity as one of the objectives of the EA \cite{Toffolo2003}. Approaches in evolutionary robotics, artificial life, and neuro-evolution realised that genetic diversity does not necessarily imply a diversity of solutions, since (i) different genotypes may encode the same behaviour and vice versa; and (ii) many genotypes may encode unsafe or undesirable solutions that should be discarded during evolution (e.g., when a robot crashes into an obstacle). Such approaches began to emphasise \textit{behavioural diversity} \cite{Mouret2009a,Gomez2009,Mouret2009,Mouret2012a}, not only as a driver for objective-based evolution but also as the enabler for diversity- or novelty-driven evolution \cite{Lehman2011}.

In \textit{quality-diversity} (QD) algorithms such as MAP-Elites  
\cite{Mouret2015} and Novelty Search with Local Competition \cite{LehmanStanley2011}, the behavioural diversity approach is combined with local competition such that the best solution for each local region in the behaviour space is stored, forming a large archive of high-quality solutions. The development of quality-diversity algorithms has allowed a plethora of applications. In robotics, this includes the design of robot morphologies and controllers \cite{Mouret2015,NordmoenEllefsen2018} as well as providing the behaviour space for behaviour adaptation algorithms \cite{Cully2015b,Bossens2020c}, which search for high-performing behaviours in the evolved archive to help robots recover rapidly from environmental changes or damages to their sensory-motor systems.

A growing number of approaches has started to explore automatically defined behaviour spaces for QD algorithms. For Novelty Search, Meyerson et al. (2016) \cite{Meyerson2016} proposed to learn a behavioural distance function to suit a particular domain of problems by increasing weights of those behavioural features that are critical to success on the target domain. For MAP-Elites, low-dimensional behaviour spaces are particularly desirable since the number of niches grows exponentially with the number of dimensions in the behaviour space. In CVT-MAP-Elites \cite{Vassiliades2018b}, Centroidal Voronoi Tesselations provide a different geometry to MAP-Elites such that each niche is defined by the regions around a centroid in the behaviour space. Since the user can pre-specify the number of centroids, this avoids the exponential increase in the number of niches. In AURORA \cite{Cully2019}, standard dimensionality reduction algorithms such as Principal Component Analysis (PCA) and auto-encoders are applied to automating the behaviour space in a distance-based archive as in Novelty Search with Local Competition. In a first study, AURORA outperformed CVT-MAP-Elites and also demonstrated the benefit of auto-encoders over PCA \cite{Cully2019}. A follow-up study \cite{Grillotti2021a} further showed the fitness advantage over TAXONS \cite{Paolo2020}, a comparable behaviour space automation method which also uses auto-encoders but which emphasises diversity but not quality.

In a variety of applications, end-users may be particularly interested in behaviour spaces that are custom-made to a particular \textit{meta-objective}, which expresses desirable properties for the final archive, such as a high number of solutions, generalisation towards particular problems, ability to form meaningful behavioural sequences, etc. In this context, we propose the newly emerging framework of \textit{quality-diversity meta-evolution} \cite{Bossens2020a,Bossens2021a}, or QD-Meta for short, to evolve a population of QD algorithms, each with their own behaviour space and optionally some other representational or algorithmic properties. The framework adapts a feature-map to define the behaviour space and through the use of a large database with the solutions generated so far, it allows to rapidly construct new archives based on the new behaviour space. The proposed approach also controls dynamically, based on the meta-population statistics, how many generations the QD archives will evolve before they are evaluated on the meta-objective. This paper aims to further demonstrate QD-Meta as a unique framework for QD optimisation by investigating the following research questions:
\begin{itemize}
\item How does QD-Meta compare to state-of-the-art methods AURORA and CVT-MAP-Elites on standard QD metrics as well as on behaviour adaptation to dynamic changes to the fitness landscape, for instance, due to changes in the environment?
\item What are the unique properties of the archives, in terms of quantitative metrics but also qualitative behaviour characterisations, as they are tailored to different meta-objectives?
\item What is the benefit of the population-based methodology of QD-Meta, in which multiple feature-maps are developed, and how does the meta-population size affect QD and meta-fitness measures of QD-Meta?
\item Previous QD-Meta work has used feature-sets, which are sets of behavioural descriptors selected by the user. Are such hand-crafted feature-sets a requirement for QD-Meta's performance?
\end{itemize}
These questions are investigated on two distinct benchmarking domains of high importance for the evolutionary computation community, namely Rastrigin function optimisation and multi-legged robot locomotion.
\section{Quality-diversity meta-evolution}
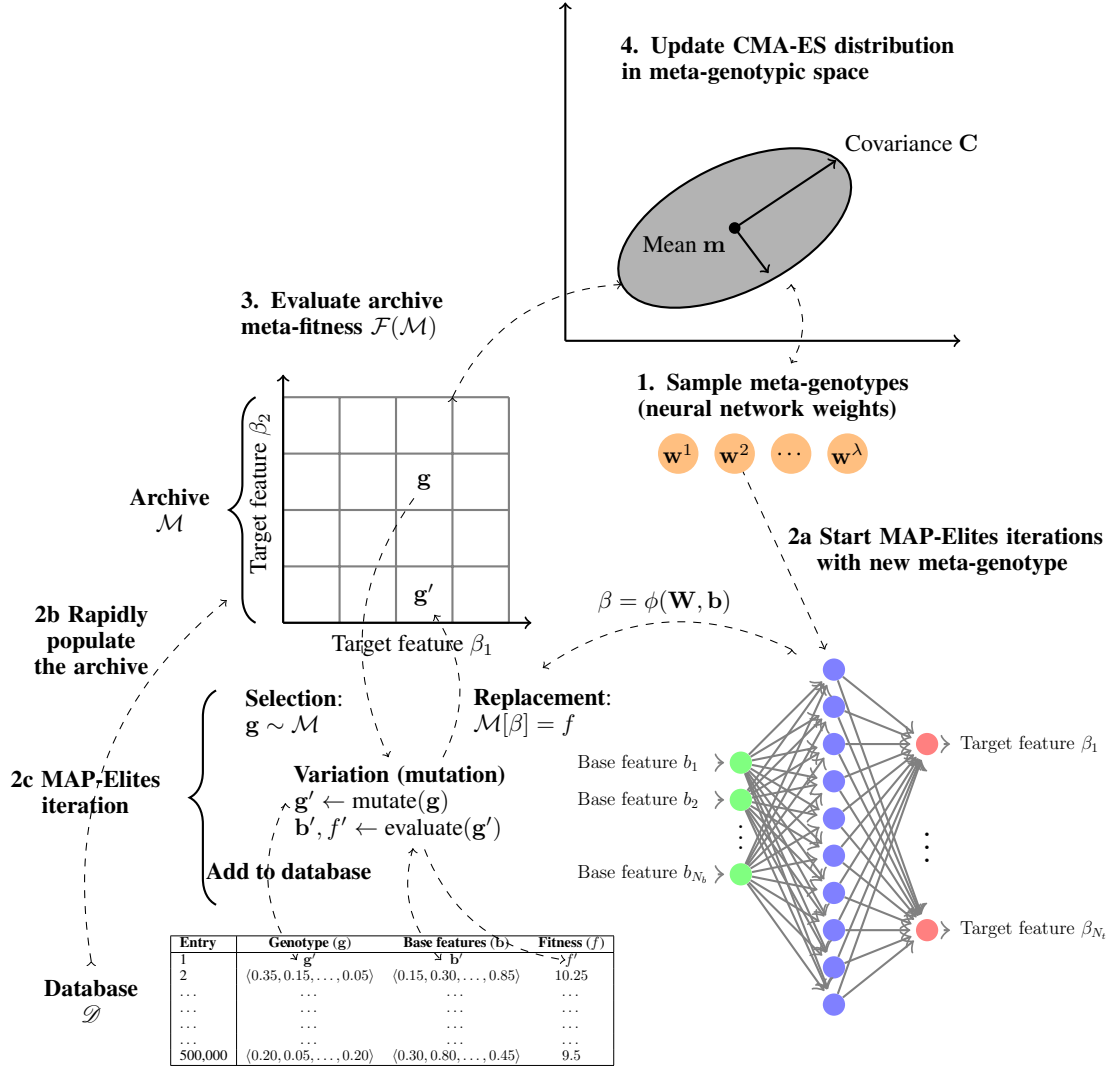
\begin{figure}
\begin{tikzpicture}[thick,scale=1.5, every node/.style={scale=0.90}]
\filldraw[] (2,2) coordinate (p0);
\filldraw[] (2,3) coordinate (q0);
\filldraw[] (4,3) coordinate (qt);
\draw[->]  (1.5,1.5) --  (1.5,4.5);
\draw[->]  (1.5,1.5) --  (5,1.5);

\begin{scope}[shift={(p0))},x={(qt)},y={($(p0)!1!90:(qt)$)}]
  \draw[fill=gray!60] (.5,0) ellipse (.5 and .25);
\end{scope}
\node[above,text width=5cm] at (3.5,3.7){\textbf{4. Update CMA-ES distribution \\ in meta-genotypic space}};
\draw[fill=black] (3,2.5) circle (1.2pt) node[below left]{Mean $\mathbf{m}$};
\draw[->] (3,2.5) --  (3.9,3.1) node[above right]{Covariance $\mathbf{C}$};
\draw[->] (3,2.5) --  (3.3,2.1) node[above right]{};

\draw[->, dashed, bend left, thin] (3.5,2.0) edge (3.5,1.3);

\tikzstyle{metageno}=[circle,fill=orange!50,minimum size=17pt,inner sep=0pt]
\draw[step=0.5cm,color=gray] (-1,-1) grid (1,1);
\draw[->]  (-1,-1) --  (1.2,-1);
\draw[->]  (-1,-1) --  (-1,1.2);
\node[name=A] at (-0.75,+0.75) { };
\node[name=B] at (-0.25,+0.75) { };
\node[name=C] at (+0.25,+0.75) { };
\node[name=D] at (+0.75,+0.75) { };
\node[name=E] at (-0.75,+0.25) { };
\node[name=G] at (+0.25,+0.25) {$\mathbf{g}$};
\node[name=H] at (+0.75,+0.25) { };
\node[name=G'M] at (+0.25,-0.75) {$\mathbf{g}'$};
\draw[->, dashed,bend left, thin](0.5,1) edge node[left,text width=4.0cm]{\textbf{3. Evaluate archive} \newline \textbf{meta-fitness} $\mathcal{F}(\mathcal{M})$}(2,2);
\node at (-2.0,0.0)[text width=2cm,text centered]{\textbf{Archive} \\$\mathcal{M}$};

\draw [decorate,decoration={brace,amplitude=10pt,mirror,raise=4pt},yshift=0pt]
(-1.15,1) -- (-1.15,-1);
\node[name=Phi] at (+3.8,-2.9) {\includegraphics[scale=2.2]{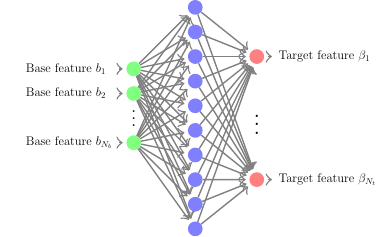}};
% ...

\draw [decorate,decoration={brace,amplitude=10pt,mirror,raise=4pt},yshift=0pt]
(-1.5,-1.6) -- (-1.5,-3.5);
\node[text width=2.3cm, text centered] at (-2.75,-2.5){\textbf{2c MAP-Elites \\ iteration}};
\node[text width=4.5cm] at (+3.5,+1.0) {\textbf{1. Sample meta-genotypes \\ (neural network weights)}};
\draw[->, dashed, thin] (+3.0,+0.5) edge node[right,text width=6.0cm]{\textbf{2a Start MAP-Elites iterations \\ \hspace{0.3cm} with  new meta-genotype}} (Phi.90);
\node[metageno] at (+2.5,+0.50) {$\mathbf{w}^1$};
\node[metageno] at (+3.0,+0.5) {$\mathbf{w}^2$};
\node[metageno] at (+3.50,+0.5) {\dots};
\node[metageno] at (+4.0,+0.5) {$\mathbf{w}^{\lambda}$};
\node at (+0.75,-0.75) { };
\node at (-1.2,0.15)[rotate=90]{Target feature $\beta_2$};
\node at (0.15,-1.2)[rotate=0]{Target feature $\beta_1$};
\node[name=G',text width=3.5cm] at (0.15,-2.6){\textbf{Variation (mutation)}\\$\mathbf{g}' \gets \text{mutate}(\mathbf{g})$ \\ $\mathbf{b}', f' \gets \text{evaluate}(\mathbf{g}')$};
\draw[->, dashed, bend right, thin] (G') edge (1.5,-3.99);
\draw[->, dashed, bend right, thin] (Phi.100) edge node[above]{$\beta = \phi(\mathbf{W},\mathbf{b})$} (1.3,-1.4);
\draw[->,dashed,bend right, thin] (G) edge (G');
\draw[->,dashed,bend right, thin] (G') edge (G'M);
\node[align=left] at (-0.9,-1.8) {\textbf{Selection}:\\ $\mathbf{g} \sim \mathcal{M}$};
\node[align=left] at (1.3,-1.8) {\textbf{Replacement}:\\ $\mathcal{M}[\beta]=f$};
\node[align=left] at (-0.95,-3.2) {\textbf{Add to database}};
\draw[->,dashed,bend right, thin] (0.15,-3.0) edge (0.40,-3.99);
\draw[->,dashed,bend right, thin] (G'.180) edge (-0.88,-3.99);
\node at (-2.7,-4.35)[text width=2cm,text centered]{\textbf{Database} \\$\mathscr{D}$};
\node at (-0.05,-4.35)[]{
\resizebox{0.40\textwidth}{!}{
\begin{tabular}{|l | c c c|}
\hline
\textbf{Entry} & \textbf{Genotype} ($\mathbf{g}$) & \textbf{Base features} ($\mathbf{b}$)   &  \textbf{Fitness} ($f$) \\
\hline
1 & $\mathbf{g}'$ & $\mathbf{b}'$  &  $f'$ \\
2 & $\langle 0.35, 0.15, \dots, 0.05 \rangle$ & $\langle 0.15, 0.30, \dots, 0.85 \rangle$  &  $10.25$ \\
$\dots$ & $\dots$ & $\dots$ & $\dots$\\
$\dots$ & $\dots$ & $\dots$ & $\dots$\\
$\dots$ & $\dots$ & $\dots$ & $\dots$\\
$\dots$ & $\dots$ & $\dots$ & $\dots$\\
500,000 & $\langle 0.20, 0.05, \dots, 0.20 \rangle$ & $\langle 0.30, 0.80, \dots, 0.45 \rangle$  &  $9.5$ \\
\hline
\end{tabular}\\
}
};
\draw[->,dashed,bend left, thin] (-2.7,-4.00) edge (-1.5,-0.8);
\node at (-2.7,-1.15)[text width=2cm,text centered]{\textbf{2b Rapidly populate the archive}};
\end{tikzpicture}
\caption{Flow diagram of QD-Meta, which repeats a four-step cycle: \textbf{1.} CMA-ES samples the meta-population from its distribution in the meta-genotypic space. \textbf{2.} Each of the meta-genotypes $\mathbf{w}^i$ for $i \in \{1,\dots,\lambda\}$ then independently applies MAP-Elites with its own feature-map: \textbf{2a} construct the feature-map $\phi(\mathbf{w}^i,\cdot)$; \textbf{2b} rapidly populate the archive  $\mathcal{M}^i$ with entries in the database $\mathscr{D}$. \textbf{2c} perform repeated MAP-Elites iterations, selecting a genotype from the archive, mutating it, evaluating its fitness $f'$ and its base-features $\mathbf{b}'$ from the observed behaviour, computing the target-features $\beta = \phi(\mathbf{w}^i,\mathbf{b}')$, adding the solution to the database, and adding it in the archive as $\mathcal{M}[\beta])$ -- if it is an elite for the region around $\beta$. \textbf{2d} the archive's meta-fitness is computed. \textbf{3.} The archives evolved by the meta-genotypes are evaluated on their meta-fitness.  \textbf{4.} Using the meta-genotypes and their meta-fitness scores, CMA-ES updates its distribution in the meta-genotypic space to find archives with the highest meta-fitness.} \label{fig: diagram}
\end{figure}

This section details our quality-diversity meta-evolution algorithm. In this study it is implemented using Covariance Matrix Adaptation Evolutionary Strategy (CMA-ES) \cite{Hansen2007,Hansen2016} to evolve a population of MAP-Elites \cite{Mouret2015} algorithms  (see Fig.~\ref{fig: diagram} for an illustration).

\subsection{MAP-Elites algorithm}
The MAP-Elites (ME) algorithm   \cite{Mouret2015} discretises the behaviour space into behavioural bins, which are equally-sized hypercubes, and then maintains for each behavioural bin the elite solution (i.e., the solution with the highest fitness), leading to quality-diversity.

ME first randomly generates an initial population of genotypes. Then, each genotype in the initial population is evaluated, resulting in a fitness score $f$ and a behavioural descriptor $\mathbf{\beta}$. Each genotype is then added to the behaviour-performance map $\mathcal{M}$ based on the following replacement rule: if the behavioural bin for $\mathbf{\beta}$ is empty (i.e., $\mathcal{M}[\mathbf{\beta}]=\emptyset$) or if the fitness is higher than the current genotype in that bin (i.e., $ f > f(\mathcal{M}[\mathbf{\beta}])$), then place the genotype $\mathbf{g}$ in that bin of the behaviour-performance map (i.e., $\mathcal{M}[\mathbf{\beta}] \gets \mathbf{g}$).

After initialisation, the algorithm applies repeated cycles of random selection, genetic variation, evaluation, and replacement. Random selection is implemented by randomly selecting genotypes from non-empty behavioural bins in the behaviour-performance maps. Genetic variation is based on mutations to the genotypes. Evaluation of genotypes is based on a user-defined fitness function $f(\cdot)$. Replacement is based on the above-mentioned replacement rule. After many repetitions of this cycle, the behaviour-performance map is gradually filled with behaviourally diverse and high-quality solutions.

\subsection{QD-Meta algorithm}
There are two key limitations to ME: 1) it is not suited to high-dimensional behaviour spaces; and 2) since the elites in the resulting behaviour-performance maps are not at any time evaluated in changed environments, they are not necessarily selected for their generalisation potential. Quality-diversity meta-evolution provides a promising perspective to overcome these issues by using feature-maps to formulate low-dimensional behaviour spaces and by evaluating its elites on a meta-objective that includes changed environments. 

The QD-Meta algorithm implementation uses Covariance Matrix Adaptation Evolutionary Strategy (CMA-ES) \cite{Hansen2007,Hansen2016} to evolve a population of MEs -- the meta-population -- and thereby optimise behaviour-performance maps with generalisation as a meta-objective. The algorithm, described in detail in Algorithm~\ref{alg: meta-CMAES}, first applies an initialisation phase to populate the behaviour-performance maps in which a large number of random genotypes are sampled, evaluated, and then added to the database $\mathcal{D}$ (see l. 1--6, and Section \ref{subsec: database} for further details on the database). After initialisation, the meta-evolution algorithm performs a large number of \textit{meta-generations}, consisting of the following steps:
\begin{enumerate}
\item Sample new meta-genotypes $\mathbf{W}^1,\dots,\mathbf{W}^{\lambda}$ from the multivariate normal distribution defined by CMA-ES (see l. 10). 
\item For $i \in \{1,\dots,\lambda \}$, use meta-genotype $\mathbf{W}^i$ to construct a new map $\mathcal{M}^i$ based on existing solutions in the database (l. 11--12). 
\item Get the meta-population state $s'$ and select the number of ME iterations $a'$ based on the $Q$-table defined by SARSA (l. 15--16).
\item For $i \in \{1,\dots,\lambda \}$, $\text{ME}(a',\mathcal{M}^i,\mathbf{W}^i)$ further evolves $\mathcal{M}^i$ (l. 18 and l. 36--42). Newly generated solutions, including their genotype, base-features, and fitness, are stored in the database to be used in step 2) of the next meta-generation.
\item Evaluate each meta-genotype $i \in \{1,\dots,\lambda\}$ on the meta-fitness $\mathcal{F}(\mathbf{W}^i)$ (l. 19; see Section~\ref{sec: experimental-setup} for its definition).  
\item SARSA($\lambda$) uses the new experience $\langle s, a, r , s', a' \rangle$ to update $Q$ (see Section~\ref{sec: SARSA}).
\item CMA-ES updates the mean, covariance, and step size, applying the $(\mu/\mu_W, \lambda)$-CMA Evolution Strategy \cite{Hansen2016} (l. 23--25). 
\end{enumerate}

\begin{algorithm}
  \caption{QD-Meta}
  \fontsize{10}{15}
  \label{alg: meta-CMAES}
    \begin{algorithmic}[1] 
      \State $\mathcal{D} \gets \emptyset$. \Comment{Create empty database.}
      \For{$i=1$ to $p$} \Comment{Create initial database.}
      \State $\mathbf{g} \gets \texttt{random-genotype()}$. 
      \State $\mathbf{b}, f \gets \texttt{eval}(\mathbf{g})$.  \Comment{Base-features and fitness.}
      \State Insert $\langle \mathbf{g}, \mathbf{b}, f \rangle$ into $\mathcal{D}$. \Comment{Fill the database (see Section~\ref{subsec: database}).}
      \EndFor
      \For{$j=1$ to $G$ } \Comment{Loop over meta-generations.}
	
      	\For {$i=1$ to $\lambda$}
      		\State Set $\mathcal{M}^i \gets \emptyset$. \Comment{Empty the map.}
      		\State $\mathbf{w} \sim \mathcal{N} \! \left( \mathbf{m}, \sigma \mathbf{C} \right)$. \Comment{Sample meta-genotype.}
      		\For { $\langle \mathbf{g},\mathbf{b},f \rangle \in \mathcal{D}$ } \Comment{Construct map from database.}
      		\State \texttt{add-to-map}($\mathcal{M}^i$, $\mathbf{w}$, $\mathbf{g}$, $\mathbf{b}$, $f$).
      		\EndFor
      	\EndFor
      	\State Get meta-population state $s'$. \Comment{see Section~\ref{sec: SARSA}}	    
      	\State $a' \gets \texttt{epsilon-greedy}(s')$. \Comment{Control ME  iterations.}
      	\For {$i=1$ to $\lambda$}
      		\State Perform \texttt{ME-iterations}($a'$,$\mathcal{M}^i$,$\mathbf{w}^i$).
      		\State $\mathcal{F}_i \gets$ \texttt{Meta-fitness}($\mathcal{M}^i$). 
      	\EndFor
      \State SARSA($\lambda$) update on $\langle s,a,r,s',a'\rangle$.  \Comment{see Eq.~\ref{eq: Q}}
        \State $s \gets s'$; $a \gets a'$.
      	\State $\mathbf{m} \gets $ \texttt{Update-mean}().
      	\State $\mathbf{C} \gets $ \texttt{Update-covariance}().
      	\State $\sigma   \gets $ \texttt{Update-step}().
      \EndFor 
      \Procedure{add-to-map}{$\mathcal{M}$, $\mathbf{w}$, $\mathbf{g}$, $\mathbf{b}$, $f$}
      \State $\mathbf{W} \gets \texttt{transform}(\mathbf{w})$. \Comment{Transform meta-genotype (see Section~\ref{subsec: featuremaps}).}
      \State $\mathbf{\beta} \gets \phi(\mathbf{W},\mathbf{b})$. \Comment{Apply feature-map to get target features (see Eq.~\ref{eq: nonlinfm}).}
      \If {$\mathcal{M}[\mathbf{\beta}] = \emptyset$ \textbf{ or } $f > f(\mathcal{M}[\mathbf{\beta}])$}
      \State $\mathcal{M}[\beta] \gets \mathbf{g}$. \Comment{Add genotype $\mathbf{g}$ to the map $\mathcal{M}$.}      
      \EndIf
      \EndProcedure
      \Procedure{ME-iterations}{$I$,$\mathcal{M}$, $\mathbf{w}$}
      \For {$i=1$ to $I$} \Comment{$I$ is the number of iterations.}
      	\State $\mathbf{g} \sim \mathcal{M}$. \Comment{Sample genotype randomly from map.}
      	\State $\mathbf{g}' \gets \texttt{mutate}(\mathbf{g})$. \Comment{Mutation.}
      \State $\mathbf{b}, f \gets \texttt{eval}(\mathbf{g}')$.  \Comment{Base-features and fitness.}
	  \State \texttt{add-to-map}($\mathcal{M}$, $\mathbf{w}$, $\mathbf{g}'$, $\mathbf{b}$, $f$).
      \State Insert $\langle \mathbf{g}', \mathbf{b}, f \rangle$ into $\mathscr{D}$. \Comment{Fill the database.}
      \EndFor
      \EndProcedure
    \end{algorithmic}
\end{algorithm}
\subsection{Database}
\label{subsec: database}
The database $\mathscr{D}$ stores a large number of previously found solutions to enable rapidly generating new behaviour-performance maps. Each such solution is a tuple $\langle \mathbf{g}, \mathbf{b}, f  \rangle$, where $\mathbf{g}$ is the bottom-level genotype (e.g., the parameters of a controller), $\mathbf{b}$ is an extended behavioural description of the solution according to a large number of $N_b$ user-defined behavioural base-features, and $f$ is the fitness (e.g., the performance of a controller). 

Two database types have been explored in prior work, a circular buffer \cite{Bossens2020a} and the $k$-best database designed to preserve high-performing solutions with high behavioural diversity \cite{Bossens2021a}. We opt for the former, as the latter is not scalable to high-dimensional base-behavioural spaces.
\subsection{Non-linear feature-maps}
\label{subsec: featuremaps}
In QD-Meta, feature-maps are used to transform the base-behavioural features $\mathbf{b} \in [0,1]^{N_b}$ to a low-dimensional behavioural descriptor $\mathbf{\beta} \in [0,1]^{N_t}$. We use non-linear feature-maps based on a feed-forward neural network with scaled sigmoid function,
\begin{equation}
\label{eq: nonlinfm}
\begin{split}
n(\mathbf{W}, \mathbf{b}) & = \mathbf{W}^{2} S_{N_b}(\mathbf{W}^{1}  \mathbf{b} + B^1) + B^2 \\
\phi(\mathbf{W}, \mathbf{b}) & = S_{N_h}(n(\mathbf{W}, \mathbf{b})) \,,
\end{split}
\end{equation}
 where $S_N(\mathbf{x}) = 1 / \left( 1 + \exp\left(-\alpha_{s} \mathbf{x}/(N+1)\right)\right)$ is an elementwise sigmoid function that scales the incoming activation based on the number of incoming units, $N$;\footnote{Hidden units receive a weighted activation based on $N=N_b$ input units while output units receive a weighted activation based on $N=N_h$ hidden units} $\alpha_s$ is an empirically defined 
 scaling factor; and the transformed meta-genotype, $\mathbf{W}$, is composed of a weight matrix from input to hidden layer, $\mathbf{W}^{1} \in \mathbb{R}^{N_h \times N_b}$, a weight matrix from hidden layer to output, $\mathbf{W}^{2} \in \mathbb{R}^{N_t \times N_h}$, and the corresponding bias units $B^1, B^2 \in \mathbb{R}$. For networks such as $\phi(\mathbf{W}, \mathbf{b})$, universal approximation theorems (e.g., \cite{Leshno1993,Hornik1989a}) imply that in principle all multi-variate functions over closed and bounded intervals can be represented to arbitrary precision, assuming a sufficient number of neurons. 
 
Such feature-maps strongly outperform other linear and feature-selector feature-maps in practice (see \cite{Bossens2021a}). Beyond the ability to represent arbitary input-output mappings, the output sigmoid activation in Eq.~\ref{eq: nonlinfm} also has a favourable statistical profile. It accounts for the high frequencies of near-zero values of $n(\mathbf{W}, \mathbf{b})$ as it increases steeply for values close to zero and slowly for extreme values. This ensures that for a sizeable proportion of mappings, each bin in $[0,1]^{N_t}$ is frequently represented, leading to diversity and local competition in the target-feature space -- the key requirements for quality-diversity.
\subsection{Dynamic parameter control}
\label{sec: SARSA}
An additional feature of the algorithm is dynamic parameter control to find high-performing dynamic schedules of parameters such as the number of generations per meta-generation and the mutation rate. Compared to other dynamic parameter control schedules, reinforcement learning (RL) using the SARSA($\lambda$) implemetation  of Karafotias et al. (2014) \cite{Karafotias2014} has been demonstrated as the method of choice in QD-Meta \cite{Bossens2021a}. This paper will use the same RL technique for dynamic control of the number of generations per meta-generations to solve the trade-off between quality-diversity (base-generations) and the evaluation of the quality-diversity archive in terms of the meta-fitness (meta-generations). An advantage of RL over other adaptive schedules, such as annealing, endogenous control,  multi-armed bandits, etc. (see \cite{Nordmoen2018} for annealing and endogenous control in QD and  \cite{Karafotias2015} for an overview), is that RL makes use of a rich observation history with various indicators of meta-evolutionary progress.

In the RL setup for QD-Meta parameter control, the algorithm uses intervals of the parameter setting as ``actions'' and then learns which actions result in meta-fitness improvements given observations on the meta-evolutionary progress. Observations are the maximum, mean, and standard-deviation of meta-fitness, the meta-genotypic diversity, the number of consequent meta-generations the maximal meta-fitness has not improved, and the reward. The reward is the ratio improvement in maximal meta-fitness divided by the function evaluations performed in the latest meta-generation. To ensure the state-space is not too large with such continuous observations, the state of the RL agent is formed using a tree-based discretisation \cite{Uther1998}, in which states represent different partitions with Q-values indicated as significantly distinct according to a Kolmogorov-Smirnoff test. 

The RL parameter control follows the algorithm by Karafotias et al. (2014) \cite{Karafotias2014}. It uses the SARSA algorithm \cite{Rummery1994,Sutton2018b}, an on-policy version of Q-learning, with eligibility traces (see SARSA($\lambda$) \cite{Sutton2018b}). The algorithm performs $\epsilon$-greedy action selection, selecting the best action according to $a^*=\argmax_{a} Q(s,a)$ with probability $1-\epsilon$ and selecting a random action with probability $\epsilon$. After each experienced transition of state, action, reward, state, and action, $\langle s_t,a_t,r_{t},s_{t+1},a_{t+1}\rangle$, the Q-table is updated for all eligible state-action pairs as follows:
\begin{equation}
\label{eq: Q}
Q(s,a) \pluseq \alpha e(s,a) \big( r_t + \gamma Q(s_{t+1},a_{t+1})  - Q(s_t,a_t)  \big) \,,
\end{equation}
where $\gamma$ is the discount factor (discounting the future rewards); $\alpha$ is the learning rate; and the eligibility $e(s,a)$ gives more weight to recently experienced state-action pairs, which addresses the temporal credit assignment problem  in a way that works well for online learning in a non-episodic environments.\footnote{The dynamic parameter control problem is non-episodic since there is only a single lifetime with no terminal states} This is done by assigning $e(s_t,a_t) \gets 1$ and $e(s,a) \gets e(s,a) \lambda \gamma$ for all other state-action-pairs $(s,a) \in \mathcal{S} \times \mathcal{A}$ .

\section{Experiment setup}
\label{sec: experimental-setup}
To illustrate the principles of QD meta-evolution on a variety of applications, experiments are conducted on two benchmarks, dynamic function optimisation and hexapod robot locomotion.\footnote{Open-source code for replicating the experiments is available on the repository \url{https://github.com/resilient-swarms/RHex_experiments}.} In both benchmarks, there are two phases: 
\begin{enumerate}
\item \textbf{Evolution phase}: the behavioural archives are evolved using QD algorithms; a summary of evolutionary parameters is given in Table~\ref{tab: evolutionparameters}.
\item \textbf{Test phase}: the behavioural archives are assessed on fitness landscapes not experienced during evolution. 
\end{enumerate}
The key hypothesis is that if robustness to dynamic changes is incorporated to the meta-fitness, then archives evolved with QD meta-evolution will have improved performance on the test phase when compared to other QD algorithms.  

\subsection{Baseline algorithms}
To evaluate QD meta-evolution compared to the state-of-the-art for low-dimensional behaviour spaces in QD, we  compare to two baseline algorithms:
\begin{itemize}
\item \textbf{AURORA} \cite{Grillotti2021a}. To represent unsupervised automation of low-dimensional behaviour spaces, we include the popular AURORA algorithm mentioned in the introduction. Based on its superior overall performance over other AURORA variants, we choose AURORA-CSC-Uniform, which uses  Container Size Control (CSC), a proportional control technique to match a desired pre-specified container size, and uniform selection over the archive. Out of the two neural network architectures for the feature-maps experimented with in the literature, Multi-Layer Perceptron (MLP) and Convolutional Neural Networks (CNN), we select the MLP as the base-features are not spatially correlated pixel data as is typical in CNN applications. The implementation is based on the source code provided in Grillotti \& Cully (2021) \cite{Grillotti2021a} (see \url{https://github.com/adaptive-intelligent-robotics/AURORA}).
\item \textbf{CVT-MAP-Elites} \cite{Vassiliades2018b}. An alternative methodology for automated low-dimensional behaviour spaces, without use of feature-maps, is to formulate the problem in one dimension based on a few prototype behaviours. To represent this approach for dimensionality reduction in QD algorithms, we include Centroidal Voronoi Tesselations MAP-Elites, in which the niches are determined by a behaviour's distance to a limited number of pre-defined centroids spread evenly across the behaviour space. The implementation is based on the CVT-MAP-Elites module for the \texttt{sferes2} framework (see \url{https://github.com/sferes2/cvt_map_elites}), which was used in Vassiliades et al. (2021) \cite{Vassiliades2018b}.
\end{itemize}

In the experiments, the baseline algorithms and QD-Meta algorithms have the same base-behavioural features, denoted as $\mathbf{b} \in [0,1]^{N_b}$. For AURORA and QD-Meta, base-features are projected onto a set of target-features, $\mathbf{\beta} \in [0,1]^{N_t}$ by means of their feature-maps. For CVT-MAP-Elites, the centroids are formulated within the same base-feature space.

\begin{table}[htbp!]
\centering
\caption{Parameter settings for evolution. Top half shows settings common to all conditions while bottom half shows settings for QD-Meta conditions.}  \label{tab: evolutionparameters}
\begin{tabular}{l  l l}
\toprule
\textbf{Parameter}  & \textbf{Setting for Rastrigin} & \textbf{Setting for RHex} \\ \hline
Genotype  ($\mathbf{g}$) & $[0,1]^{20}$  & discretised in $[0,1]^{24}$  \\
Base-behaviour space & genotype & trajectory in $[0,1]^{50}$ or feature-sets in $[0,1]^{15}$\\
Mutation rate   & $0.10$ & $0.125$ \\
Mutation type &  Gaussian with SD $\sigma = 0.05$ & random increment or decrement with step of $0.025$ \\
Maximal number of solutions & 10,000 solutions & 4,096 solutions \\
Function evaluations & 100,000,000 & 12,000,000 \\ 
Batch size per generation & 400 &   400 \\
Initial population ($p$) & 2,000 & 2,000 \\
\hline
Meta-population size ($\lambda$) & 10 & 10 \\
Number of target-features ($N_t$) & 2 & 4 \\
Number of hidden units ($N_h$) & 10 & 10 \\
Meta-genotype    ($\mathbf{w}$) & $[-1,1]^{222}$ & $[-1,1]^{542}$ \\
Sigmoid scaling factor ($\alpha_s$)  & 30 & 30\\
Database capacity & 500,000 & 500,000 \\
\bottomrule
\end{tabular}
\end{table}
\subsection{Rastrigin function optimisation benchmark}
In a first set of experiments, quality-diversity archives are evolved on a 20-dimensional Rastrigin function and then assessed on a set of test-perturbations, which represent a dynamic change of the fitness landscape. While standard genetic algorithms can solve the 20-D Rastrigin near-optimally, the archives evolved by QD algorithms have a unique benefit of covering various peaks of a fitness landscape; therefore, QD algorithms can potentially provide a near-optimal solution after a dynamic landscape change. The following two independent QD-Meta conditions are run, QD-Meta Dimension and QD-Meta Translation, corresponding to two different  meta-objectives: 1)  dimensionality decreases which remove one input-dimension from the Rastrigin computation; and 2) linear transformations over the fitness landscape. The test phase  runs two independent test scenarios, each of which are applied to all the algorithms. The Dimension test consists of injecting sinusoidal noise to selected pairs of dimensions while the Translation test consists of performing linear transformations. The resulting fitness landscapes of the tests are out-of-distribution even for the QD-Meta conditions.
\paragraph{Evolution phase}
The 20-dimensional Rastrigin function is highly multi-modal and non-linear, posing a challenging optimisation problem. Formulated as a maximisation problem it takes the following form:
\begin{equation}
\label{eq: Rastrigin}
f(\mathbf{g}) =  - 10 N_{g}  \Big( \sum_{i=1}^{N_g} g_i^2 - 10 \cos(2 \pi g_i)  \Big) \,,
\end{equation}
where $N_g$ is the number of genes in the genotype, $\mathbf{g} \in [-5.12,5.12]^{N_g}$.  

The following settings are common to all algorithms. The Rastrigin with its continuous search space is optimised by a Gaussian mutation operator with standard deviation of $\sigma=0.05$ and a mutation rate of 0.10. As the Rastrigin experiments have low-cost function evaluations, a large budget of 100 million function evaluations and a large number of 10,000 solutions are allowed. The base-behavioural features of a solution are equal to its genotype. 

Further algorithm-specific settings are as follows. For generating its centroids, CVT-MAP-Elites uses the k-means algorithm on randomly generated data points; here, we use 1 million data points of randomly generated points in $[0,1]^{20}$. For QD-Meta,  the meta-population size is increased to 10 compared to 5 in prior works, mainly to discourage getting stuck in local optima of the meta-fitness. The meta-fitness is implemented in two distinct conditions. In a first condition, called \textbf{QD-Meta Dimension}, the meta-fitness aims to provide behaviourally diverse and high-quality solutions despite changes to the dimensionality of the Rastrigin function. For QD-Meta Dimension, the meta-fitness of an archive $\mathcal{M}$ combines the archive-summed fitness, averaged over different dimensionality decrements, and the summed pairwise distance in the genotypic space (here also the base-behavioural space):
\begin{equation}
\mathcal{F}_{\text{d}}(\mathcal{M}) = - \frac{1}{|\mathcal{J}|}   \Big( \sum_{j \in \mathcal{J}} \sum_{\mathbf{g} \in \hat{\mathcal{M}}} M + \bar{f}_{\{j\}}(\mathbf{g}) + \alpha \sum_{\mathbf{g}' \neq \mathbf{g}} || \mathbf{g} - \mathbf{g}' ||_2   \Big)\,,
\end{equation}
where $\mathcal{J} \subset \{1,\dots, 20\}$ is a set of 10 indices randomly selected before the start of evolution; $\hat{\mathcal{M}}$ is a random subset of 10\% of the solutions of the full archive, selected anew for each meta-fitness evaluation; an upper bound $M$ normalises  $\tilde{f}_j(\mathbf{g})$ to the range $[0,M]$ to obtain a positive score; $\bar{f}_{\{j\}}(\mathbf{g}) =  - 10 (N_{g} - 1) \sum_{i \neq j} g_i^2 - 10 \cos(2 \pi g_i)$, thereby removing one selected dimension; and $\alpha=\frac{\sqrt{N_{g}} M}{|\hat{\mathcal{M}}| - 1}$ is scaled such that diversity and quality have comparable importance. Taking the archive-summed fitness rather than archive-averaged fitness implies a bonus for larger archives. In a second condition, called \textbf{QD-Meta Translation}, the meta-fitness aims to provide high-quality solutions despite various translations to the genotype changing the location of the global optimum. For QD-Meta Translation, the meta-fitness of an archive $\mathcal{M}$ is the average of the archive-summed fitness across different translations:
\begin{equation}
\mathcal{F}_{\text{t}}(\mathcal{M}) = - \frac{10}{|\mathcal{T}| N_{g}}   \Big( \sum_{T \in \mathcal{T}} \sum_{g \in \hat{\mathcal{M}}} \sum_{i=1}^{N_g} T(g_i)^2 - 10 \cos(2 \pi T(g_i))  \Big) \,,
\end{equation}
where $\mathcal{T}$ is a set of randomly generated linear transformations of the type $T(x) \gets ax + b$ where $b \in [-0.50,0.50]$ represents a small to medium-sized shift of the fitness landscape and $a \in [-1.10,-0.91] \cup [0.91,1.10]$ represents a 10\% expansion or shrinkage of the landscape, with or without a sign change. To reduce variance across meta-evolution while sampling different values of $a$ and $b$, we sample transformations based on 8 non-overlapping ranges determined by three boolean features: the sign of $a$, whether $a$ shrinks or expands $x$, and the sign of $b$. So, for example, $T_1$ has $a \sim U(-1.10,-1), b \sim U(-0.50,0)$, $T_2$ has $a \sim U(-1.10,-1), b \sim U(0,0.50)$, $T_3$ has $a \sim U(1,1.10), b \sim U(-0.50,0)$, etc. For each range there are two repetitions, yielding 16 random translations in each meta-fitness evaluation.

\paragraph{Test phase}
The test phase assess rapid adaptation to fitness landscape changes in no more than 100 function evaluations of random search over the behavioural archive, randomly selecting solutions without replacement. Two distinct and independently run tests are set up which compared to the meta-fitness functions are conceptually similar but have a completely different distribution of optimal solutions.\\ 
\indent In the \textbf{Dimension test}, one removes two selected dimensions of the genotype from the Rastrigin computation and replaces it with high-frequency sinusoidal noise with the same amplitude (i.e. 10) as the cosine in the original Rastrigin):
\begin{equation}
\label{eq: dim-reduced}
\begin{split}
\tilde{f}_{\mathcal{J}}(\mathbf{g}) =  - 10 (N_{g} - |\mathcal{J}|) \Big( \sum_{i \not \in \mathcal{J}} g_i^2 - 10 \cos(2 \pi g_i) - \sum_{i \in \mathcal{J}} 10 \sin(6 \pi g_i) \Big) \,,
\end{split}
\end{equation}
where $\mathcal{J}$ is a chosen index set. While conceptually similar to QD-Meta Dimension, the key difference is that two dimensions are altered and that the altered dimensions are replaced by a high-amplitude sine wave rather than completely being removed.\\
In the \textbf{Translation test}, one transforms the genotype using a linear transformation $g_i \gets a g_i + b$ for all $i \in \{1,\dots,N_g\}$ before computing the 20-dimensional Rastrigin (see Eq.~\ref{eq: Rastrigin}). In the test set, 120 unique index sets are generated, each of size $|\mathcal{J}|=2$. Twelve settings of the slope, $a$, are generated by applying, or not applying, shrinkage and reversal to $a'=\{0.01,0.25,0.5\}$. The settings using $a'=0.25$ and $a'=0.50$, with or without shrinkage or reversal, are far out-of-distribution compared to the QD-Meta Translation's meta-fitness. The intercept, $b$, is chosen in $\pm \{0.01,0.25,0.50,0.75,1.0\}$. Therefore, only 8 out of 120 $(a,b)$-values are in-distribution for QD-Meta Translation and values close to these are only rarely chosen during meta-fitness evaluation.

\subsection{Hexapod robot locomotion benchmark}
\begin{figure}
\centering
\subfloat[RHex robot]{\includegraphics[height=0.25\linewidth]{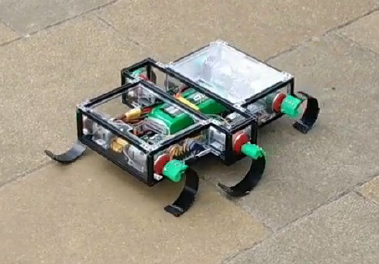} }
\subfloat[Down-and-up stairs]{\includegraphics[width=0.30\linewidth]{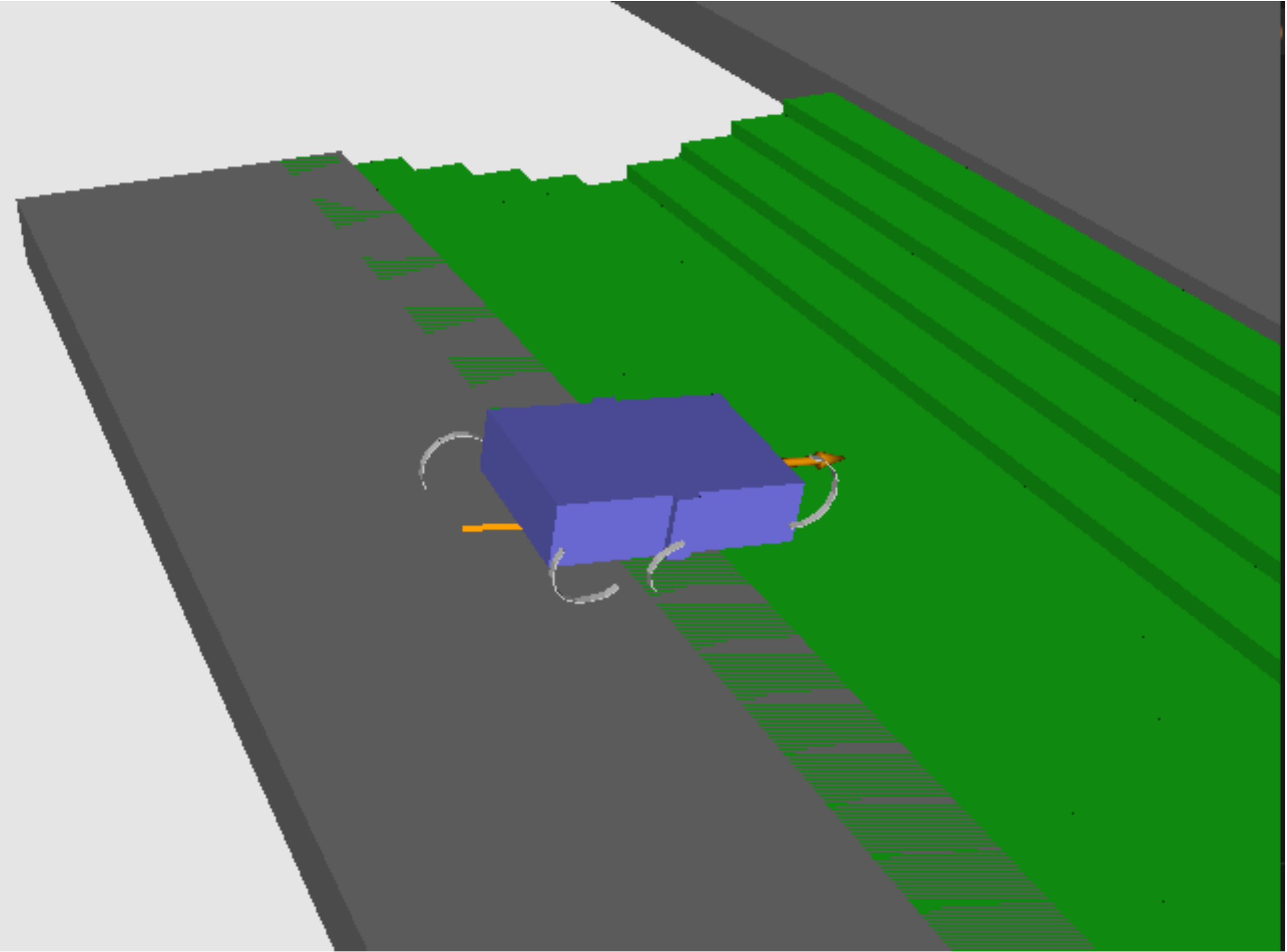}} \hfill
\subfloat[Thick pipe]{\includegraphics[width=0.30\linewidth]{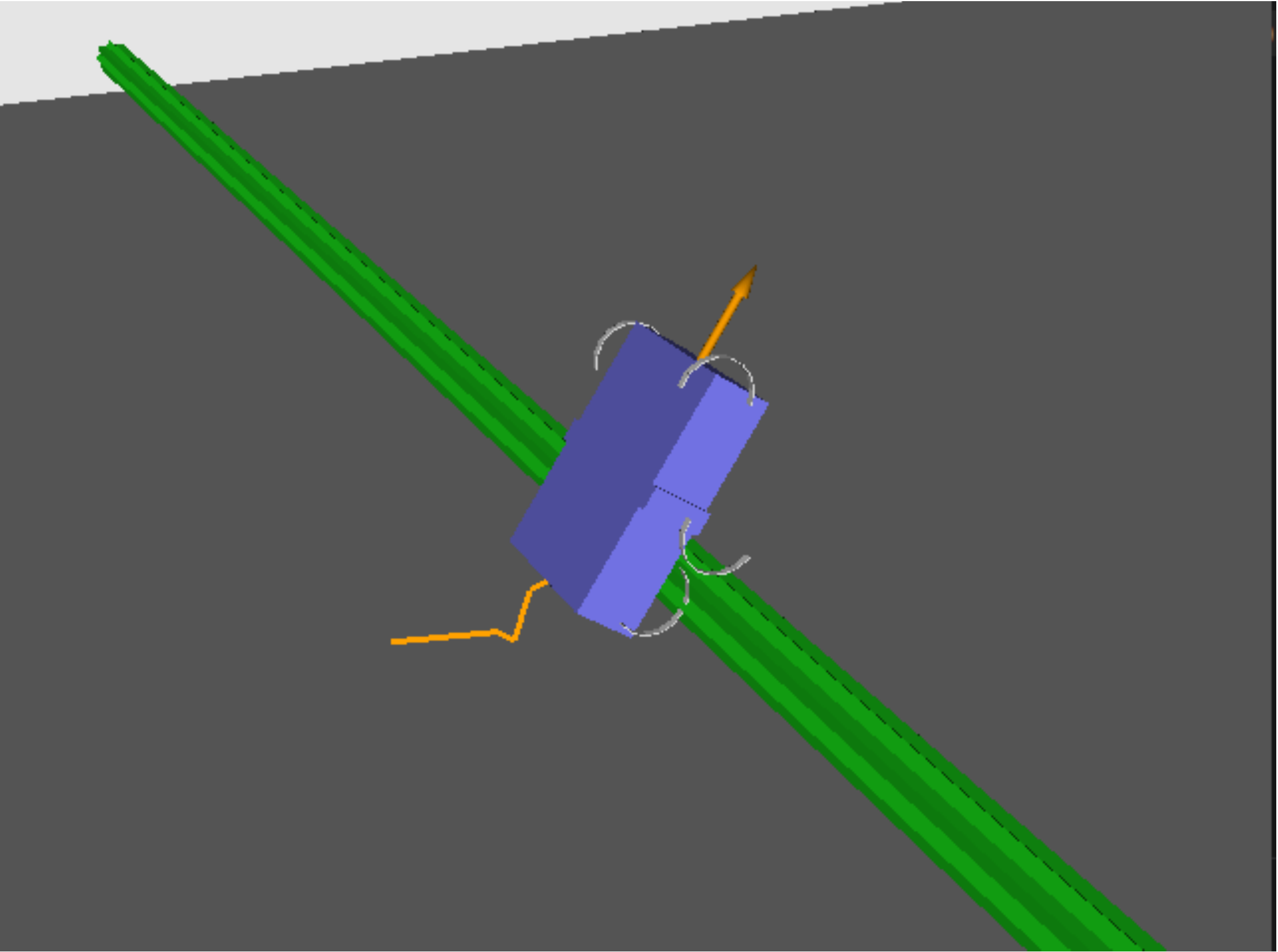}} \\
\caption{The RHex hexapod robot platform: \textbf{(a)} the physical robot, on which the simulation is based; \textbf{(b)} the down-and-up stairs obstacle course; \textbf{(c)} the thick pipe obstacle course. For the full set of obstacle courses, see Fig.~S2 of the Supplemental Materials.} \label{fig: RHex}
\end{figure}

To evaluate the QD systems on a realistic application, they are compared on the RHex  hexapod robot platform (see Figure \ref{fig: RHex}) \cite{Saranli2001}, which is of interest because RHex can move across difficult terrains at high speeds. For QD-Meta, we independently run 4 different conditions based on two meta-objectives and two types of base-behavioural spaces per meta-objective. The meta-objectives are based on the performance after damaging  one of the RHex robot's legs or after introducing an obstacle into the environment. The base-behavioural space is based on either feature-sets, which are traditionally used in quality-diversity meta-evolution, or the trajectory of the robot, as is traditionally used in AURORA and CVT-MAP-Elites. In the test phase, we devise two independent test scenarios, namely damages to the robot's legs  as well as obstacle-courses. These test scenarios are not presented during the evolutionary phase and therefore represent adaptation to a priori unknown events.
\paragraph{Evolution phase}
The RHex robot is simulated using the DART (Dynamic Animation and Robotics Toolkit) physics engine \cite{Lee2018}. The task of the Rhex robot is to walk in a straight line and the fitness function $f$ is the total distance the robot has moved forward within a time span of \SI{5}{s}, on a flat plain surface where the robot faces no damages or obstacles. The control cycle of the locomotion controller is \SI{5}{ms}. 

Each leg of the robot is controlled by a Buehler clock, which alternates between a stance phase where the robot leg touches the ground and a swing phase where the leg rotates above ground \cite{Seipel2007,Saranli2001}. The genotype $\mathbf{g}$ is comprised of 24 parameters, including: the period of the clocks, $T$, between 0.33-\SI{1}{s} (clock speed within 1-\SI{3}{Hz}; parameter 1); for each leg its duty-factor, the proportion of time that the leg touches the ground (parameters 2-7); the stance angle in $[0,\pi]$, the angle in which the robot leg touches the ground (parameters 8-13); the stance offset in $[-\pi/4,\pi/4]$, the angular offset to the leg's stance phase (parameters 14-19); the phase offset in $[0,T/2]$, the time lag with which the leg touches the ground when compared to leg 1 (parameters 20-24; reference leg 1 is excluded as parameter). The resulting 24-dimensional genotypic space is discretised with a step size of $0.025$ and therefore whenever gene is mutated, the mutation operator generates random increments or decrements of one step. The RHex robot experiments involve computationally expensive simulations,\footnote{Experiments are conducted on Intel Xeon Gold 6138 (2.00GHz) CPUs. While for Rastrigin, a single run requires only one CPU run for at most 20h, the RHex robot experiment runs require 40 CPUs for around 300h.} so the budget is limited to 12 million function evaluations and only 4,096 solutions are allowed.

Settings common to all algorithms are as follows. The base-behavioural space is a 50-dimensional space in $[0,1]^{50}$ representing 10-step trajectories of the robot in terms of Cartesian $xy$-coordinates and the roll-pitch-yaw coordinates\footnote{A tight normalisation in $[0,1]^{50}$ is obtained by dividing at each step the relative change from the previous step by the maximal deviations; for example, to compute $x_2$ as a feature, $x_2 \gets 0.5 + 0.5\frac{x_2 - x_1}{\Delta x}$ is computed where $\Delta x$ is an empirical estimate of the maximal absolute value of $x$.}. This choice serves two purposes: 1) these features are comparable to the typical AURORA setup thereby allowing a fair comparison; and 2) while previous work on QD-Meta used hand-crafted feature-sets, it is not clear whether such feature-sets are required or whether trajectories similar to those used in AURORA are sufficient. This would be important to know because the construction of feature-sets often requires additional prior knowledge (as mentioned in \cite{Grillotti2021a}), resulting in a potential trade-off between applicability and the customisation.

Settings specific to particular algorithms are as follows. For CVT-MAP-Elites, feasible trajectories must be sampled before forming centroids since a large part of the base-behavioural space may not provide a feasible solution; unlike the Rastrigin function optimisation, where all behaviours are feasible, the RHex robot may have a validly formed genotype without adhering to safety criteria. To obtain a suitable dataset for the feasible behaviour space, we sample genotypes until there are 100,000 valid robot trajectories that do not violate the safety measure (i.e. the RHex robot did not turn over). For QD-Meta, we additionally include the original feature-sets (see \cite{Bossens2020a}), which combine: 1) the linear velocity in the 3 Cartesian axes (3D); 2) the frequencies of the body orientation being significantly negative or positve in the 3 orientation axes of roll, pitch, and yaw (6D); and 3) for each leg, the duty factor, i.e. the proportion of time it touches the ground. The different base-behaviour spaces will be distinguished by the suffix \textbf{Feature-sets} versus \textbf{Trajectory}. The meta-fitness is implemented in two distinct ways, namely in terms of damages to one of the robot's legs (as in \cite{Bossens2020a}) and in terms of an obstacle course making the way of the robot more challenging. In the \textbf{QD-Meta Damage} condition, the meta-fitness of a map $\mathcal{M}$ is defined as
\begin{equation}
\label{eq: meta-damage}
\mathcal{F}_{\text{damage}}(\mathcal{M}) =  \frac{1}{6|D|  |\mathcal{M}|} \sum_{\mathbf{g} \in \mathcal{M}} \sum_{d \in D} \sum_{l=1}^{6} f(\mathbf{g};d(l))  \,,
\end{equation}
where $D$ defines a set of two damage-types, selected randomly from the full damage set at the start of the evolutionary experiment; and $f(\mathbf{g};d(l))$ computes the fitness of the genotype $\mathbf{g}$ when a damage of type $d$ is applied to leg $l$ of the RHex robot. The full damage set, from which the meta-fitness is constructed, comprises four types of damages to the RHex robot's legs: 1) leg-removal, which completely removes one leg; 2) leg-shortening, which shortens one leg significantly, usually preventing it from touching the floor; 3) blocked-joint, in which a joint cannot move; or 4) passive-joint, in which the affected joint cannot be controlled but can move passively. In the \textbf{QD-Meta Obstacle} condition, the meta-fitness of a map $\mathcal{M}$ is defined as
\begin{equation}
\label{eq: meta-obstacle}
\mathcal{F}_{\text{obstacle}}(\mathcal{M}) =  \frac{1}{|O| |\mathcal{M}|} \sum_{\mathbf{g} \in \mathcal{M}} \sum_{o \in O} f(\mathbf{g}; o )  \,,
\end{equation}
where $O$ defines a set of 5 obstacles selected at random out of the full obstacle set at the start of the evolutionary experiment. The full obstacle set comprises 9 obstacle courses (see Figure~S2 in Supplemental Materials): a) a large sphere interrupting the normal trajectory; b) a continuing upward stairs; c) a slope; d) a large pile of rubble forming a rough terrain; e) downward stairs followed by upward stairs; f) several small pipes evenly spaced across the forward trajectory ; g) a ditch; h) a thick pipe; and i) a thin pipe. To avoid selecting problems that are not solvable, the obstacle courses were adjusted to meet a target optimal performance between $\SI{3}{m}$ and $\SI{5}{m}$, which was confirmed by obstacle-specific runs with a traditional EA. 

\paragraph{Test phase}
In the test phase, the archives are assessed on obstacles and damages not known a priori by any of the algorithms. The \textbf{Obstacle test} assesses rapid performance receovery on obstacles not experienced during meta-evolution while the \textbf{Damage test} assesses rapid performance recovery on damages not experienced during meta-evolution. In both cases, recovery is done by performing random search (without replacement) in no more than 50 function evaluations. To assess the overall robustness of solutions, we also perform a \textbf{Generalisation test} in which \textit{all} controllers are assessed for their average performance over \textit{all} environments.

\section{Results}
This section analyses the comparative performance of QD-Meta and the baseline algorithms on  the function optimisation and RHex robot experiments. For each benchmark, we evaluate the evolutionary development, including the number of solutions in the archive and their fitness over time, as well as their performance on the consequent test. A further qualitative analysis probes into the behavioural differences observed when comparing different meta-objectives.

\subsection{Rastrigin function optimisation}
\begin{figure}
\centering
\includegraphics[width=0.25\linewidth]{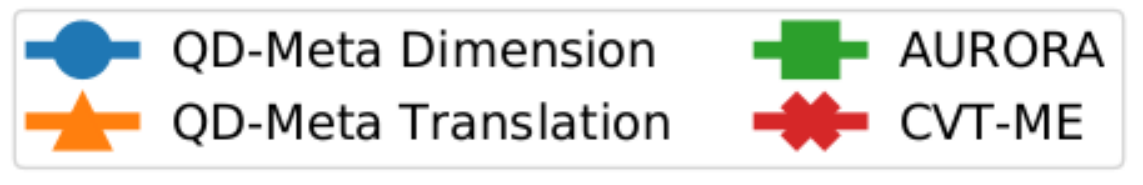} \\
\subfloat[Number of solutions]{\includegraphics[width=0.30\linewidth]{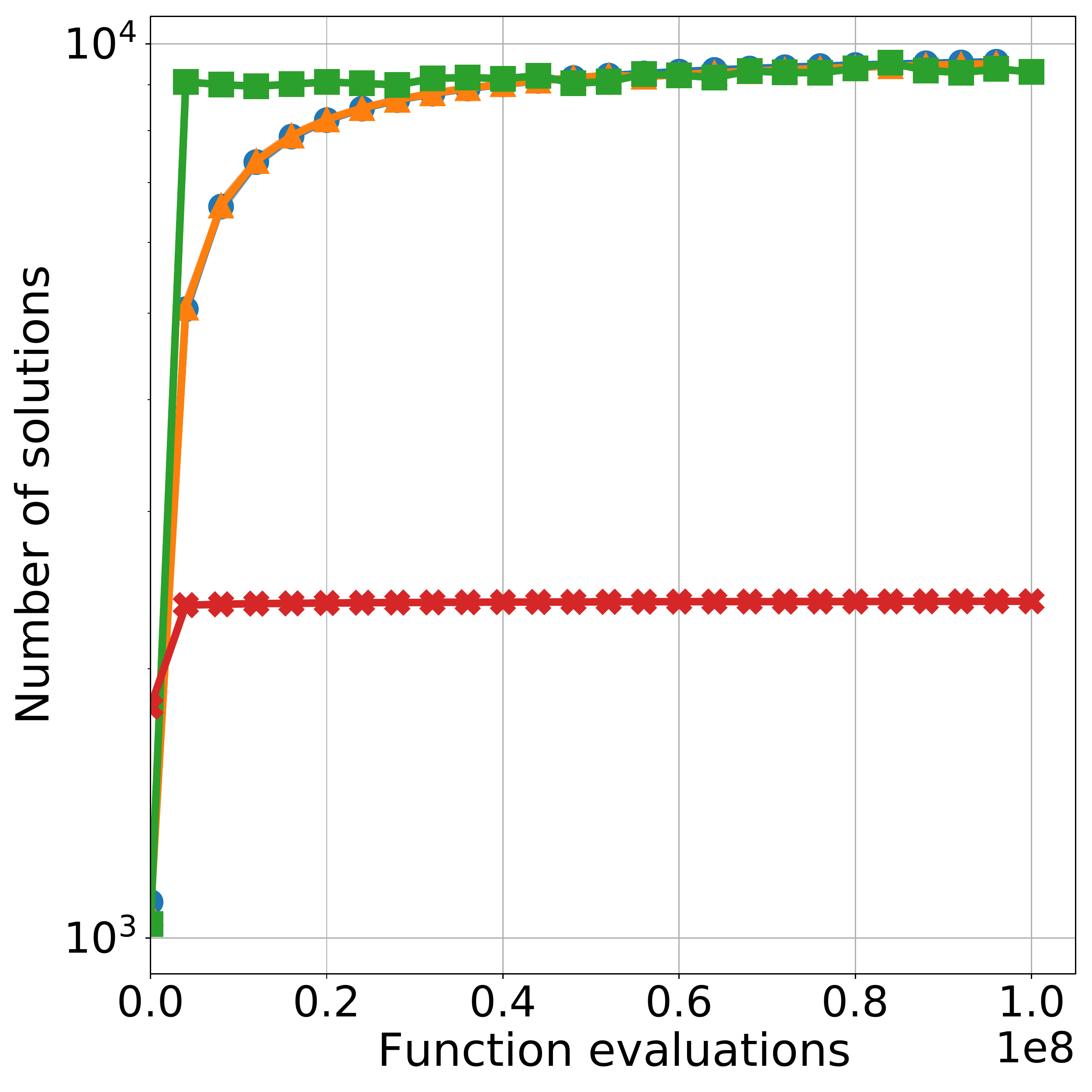}}
\subfloat[Average fitness]{\includegraphics[width=0.30\linewidth]{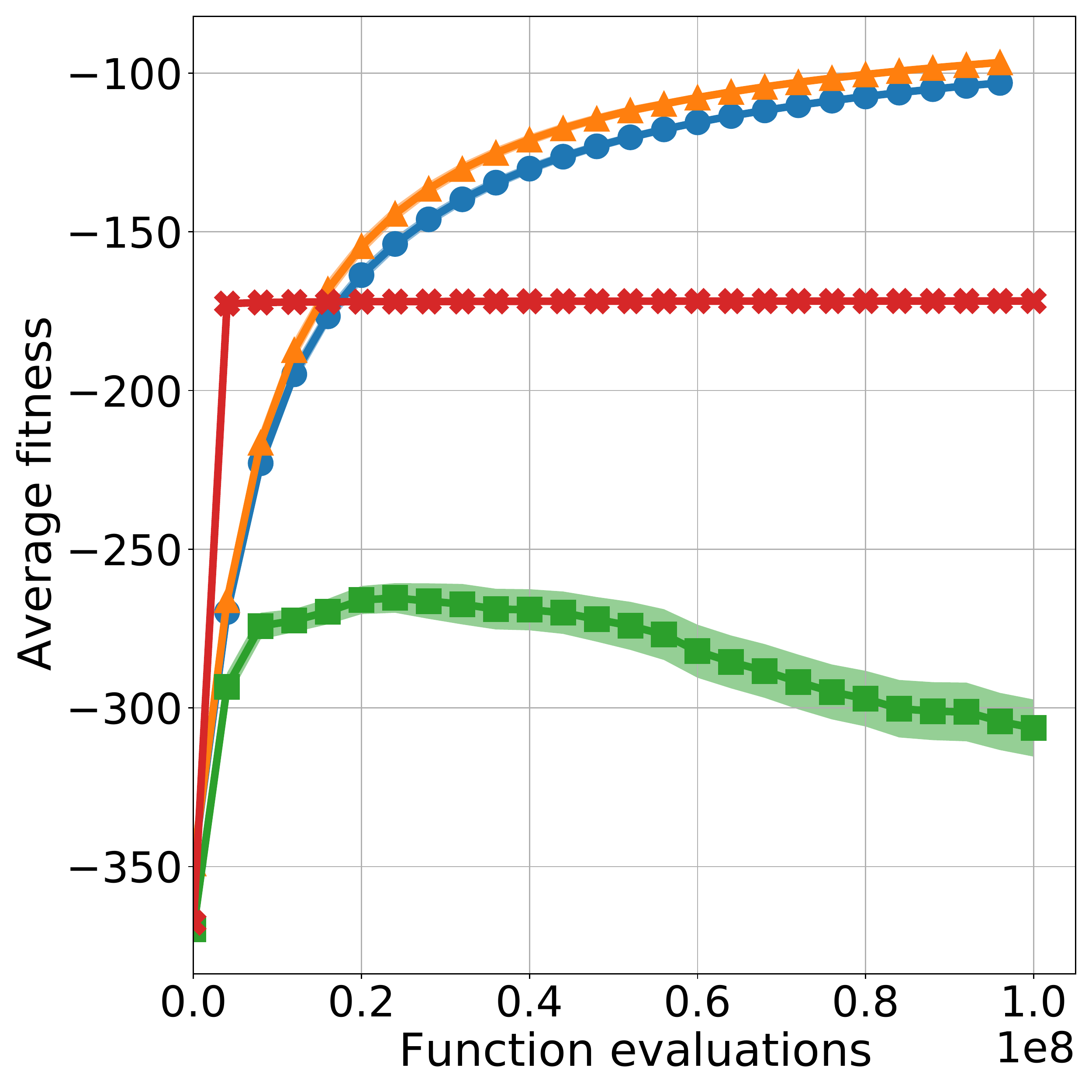}}
\subfloat[Global fitness]{\includegraphics[width=0.30\linewidth]{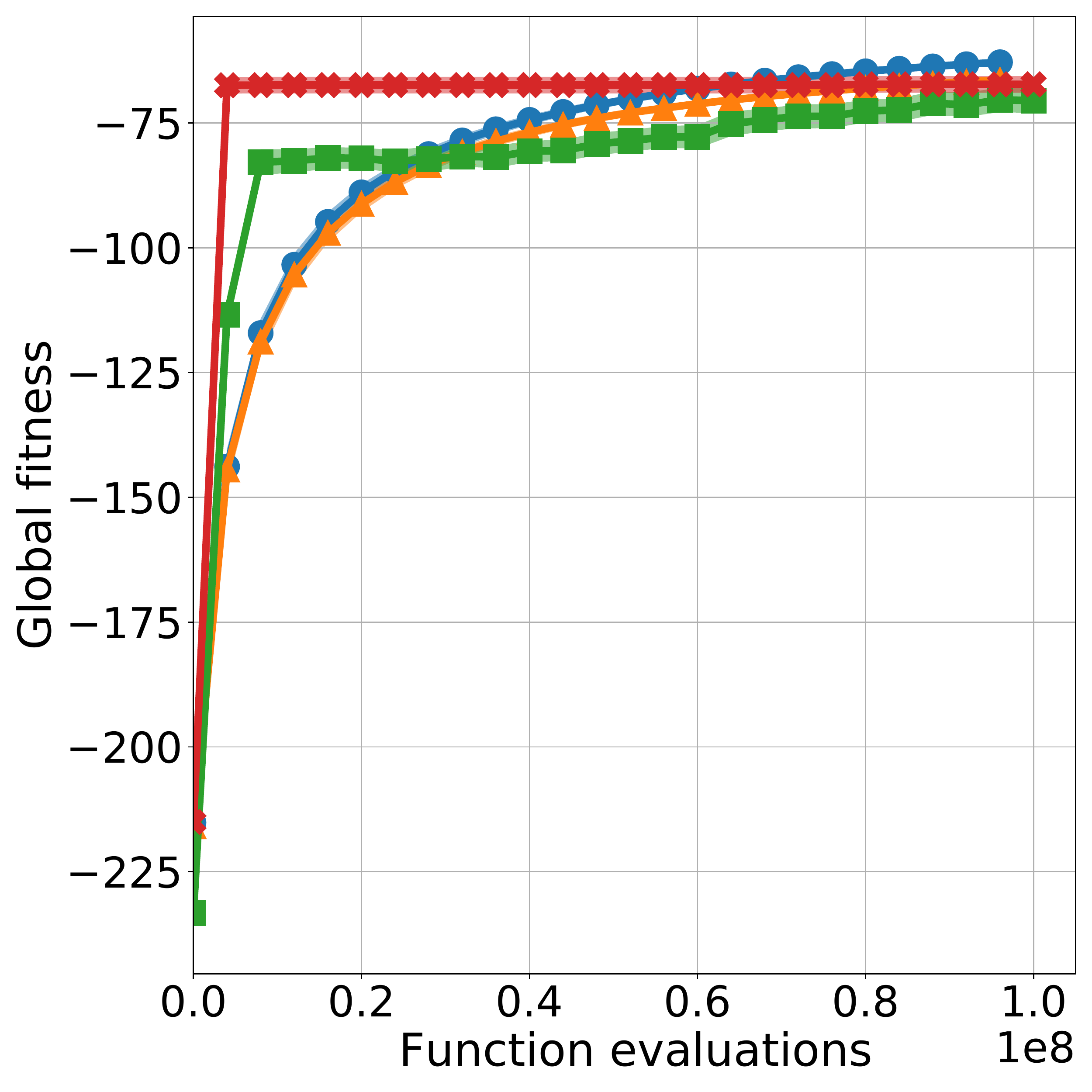}}\\
\caption{Quality-diversity statistics (Mean $\pm$ SE) of the different included QD algorithms across 20 replicates on the Rastrigin function optimisation, including \textbf{(a)} the total number of solutions in the archive; \textbf{(b)} the average fitness across the archive; and \textbf{(c)} the maximal fitness across the archive. For QD-Meta, Mean and SE statistics are aggregated across replicates and the different archives within the meta-population.} \label{fig: evolution-function}
\end{figure}

In the evolution phase, the quality-diversity statistics can be observed from Figure~\ref{fig: evolution-function}. All QD algorithms yield a high number of solutions although there is a difference between CVT-MAP-Elites, which yields just below 3,000 solutions, and AURORA and QD-Meta conditions, which yield about 10,000 solutions. The difference is explained by AURORA setting 10,000 as an explicit target for the proportional container size control and by QD-Meta conditions having a meta-fitness bonus for the number of solutions. All algorithms converge to a global fitness of around $-70$. The largest difference observed is that QD-Meta scores highest on the average fitness, with a score of $-100$, followed by CVT-MAP-Elites with a score of $-170$, and finally AURORA, $-300$. The higher average fitness of QD-Meta conditions is explained by the meta-fitness bonus for covering significant peaks of the fitness landscape (i.e. those peaks that can become top-performers after translation or dimensionality change) rather than covering the behaviour space evenly. The population-based methodology of evolving multiple feature-maps as meta-genotypes is  validated by an analysis comparing the results for QD-Meta Translation with different population sizes $\lambda \in \{1,2,5,10,20\}$ (see Fig. S1 in Supplemental Materials; dynamic parameter control disabled for simplicity), which shows that a population size of 1 has low performance on meta-fitness and QD metrics.

\begin{figure}
\centering
\includegraphics[width=0.25\linewidth]{figures/METAvsCTRL_Rastri_legend.pdf} \\
\subfloat[Dimension test]{\includegraphics[width=0.30\linewidth]{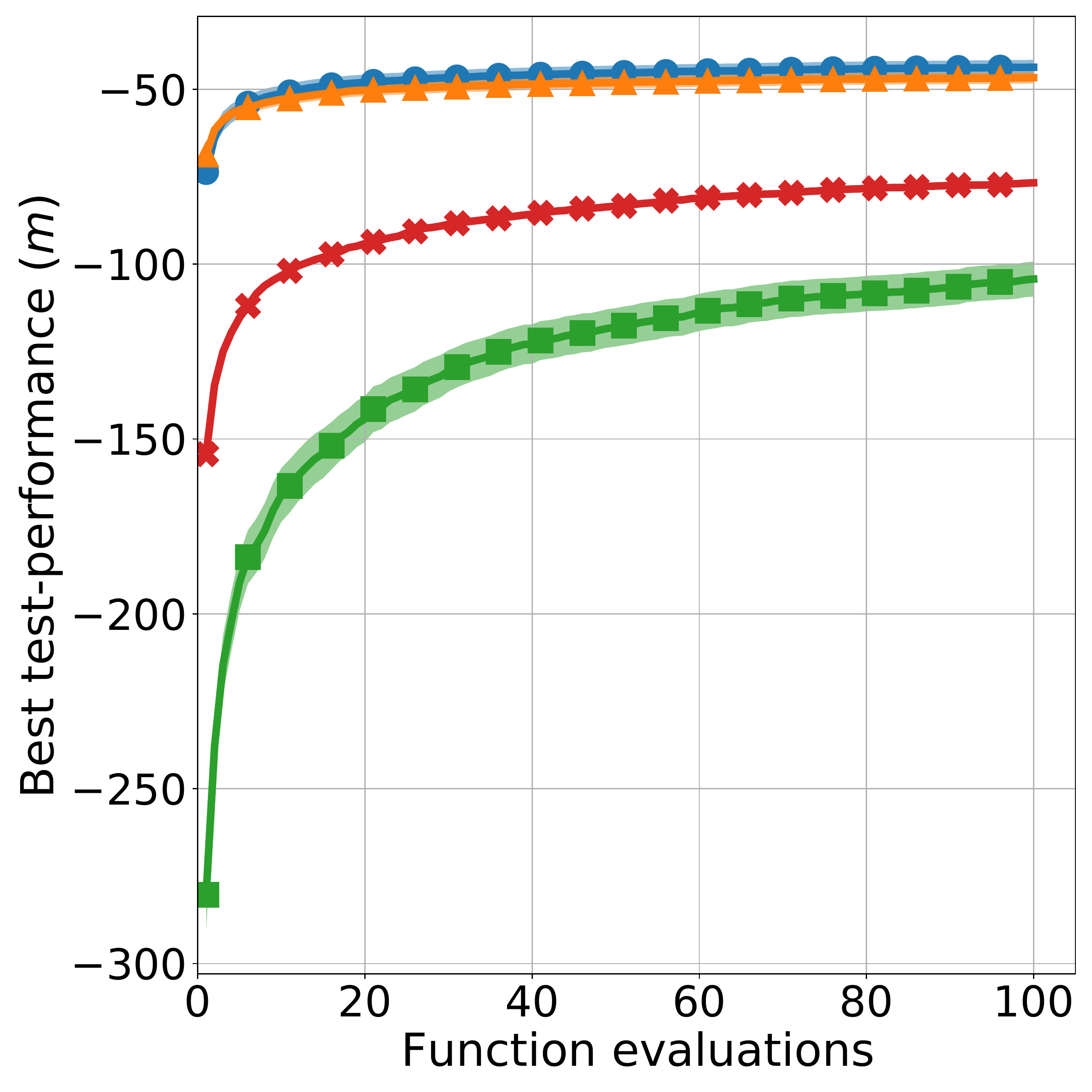}}
\subfloat[Translation test]{\includegraphics[width=0.30\linewidth]{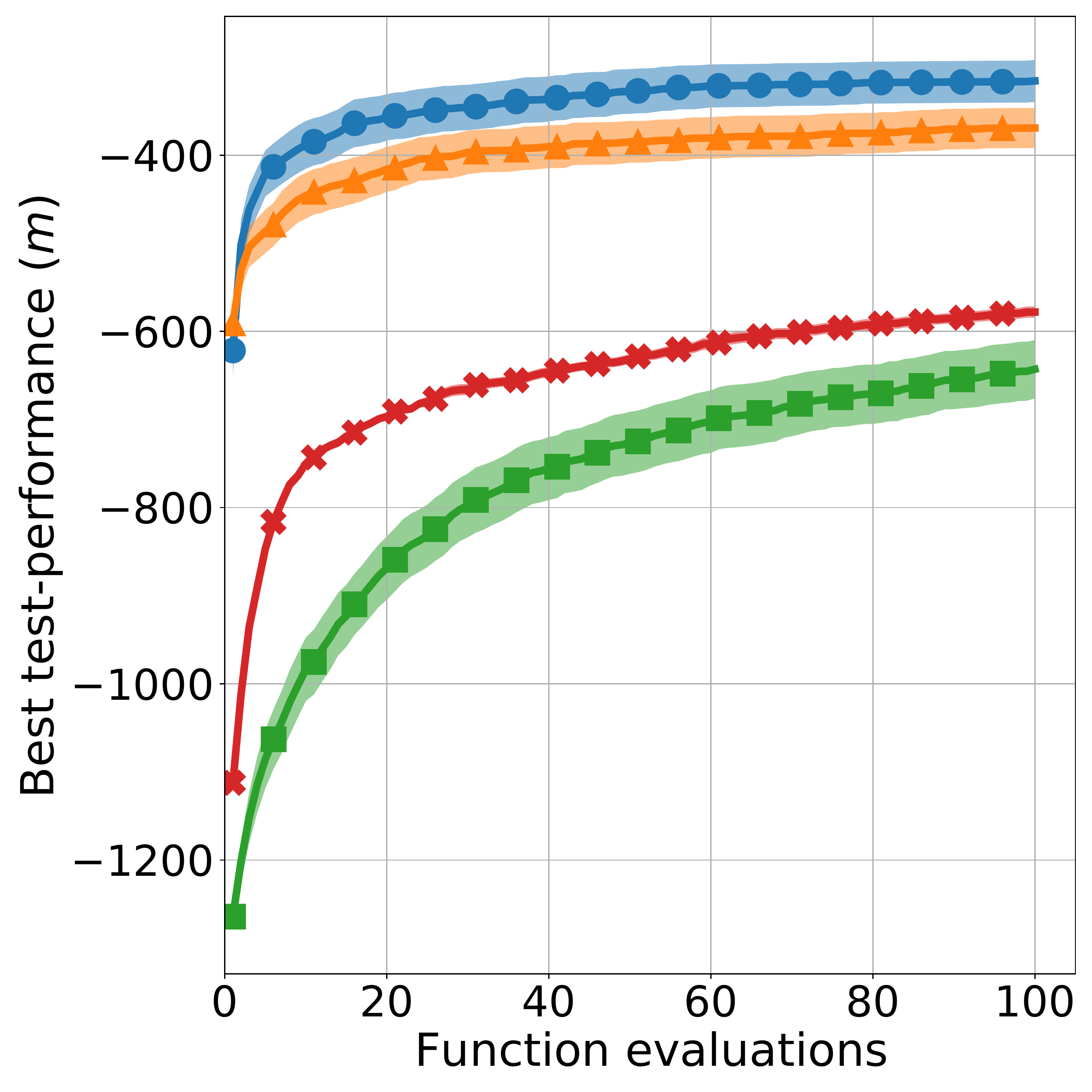}}
\caption{Test performance (Mean $\pm$ SE) of the different included QD algorithms across 20 replicates of the Rastrigin function optimisation benchmark. The $y$-axis shows the best solution so far after performing random search over the behavioural archive for the number of function evaluations indicated on the $x$-axis. For each replicate of QD-Meta, the archive with the highest meta-fitness at the end of meta-evolution is chosen.} \label{fig: adaptation-function}
\end{figure}

In the test phase, a random search across the QD archives allowing at most 100 function evaluations (see Figure~\ref{fig: adaptation-function})  demonstrates that QD-Meta algorithms significantly outperform other algorithms. Both meta-fitness conditions, QD-Meta Dimension and QD-Meta Translation, have a beneficial effect on generalisation, allowing rapid adaptation on both the Dimension test and the Translation test.

\subsection{Hexapod robot locomotion}
In robotics applications, a key challenge is to adapt to environmenal changes within a limited number of trials without trying out low-quality controllers that may cause performance loss or even safety violations. The results of the RHex robot study highlight how QD-Meta can provide archives with a lower number of solutions but a much higher average performance in unforeseen circumstances.

\begin{figure}[htbp!]
\centering
\includegraphics[width=0.40\linewidth]{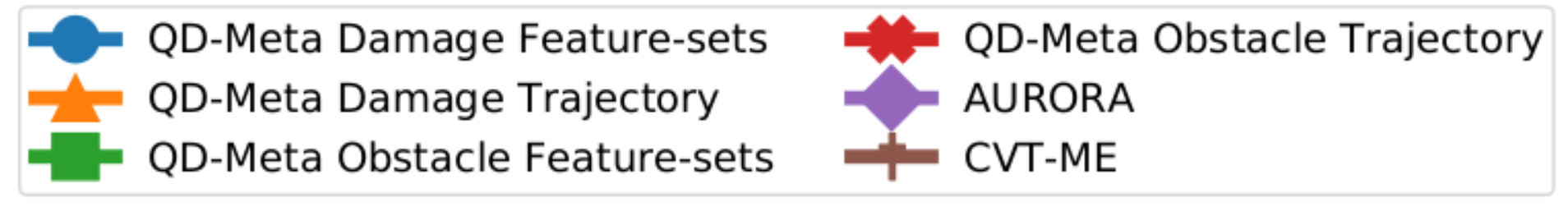} \\
\subfloat[Number of solutions]{\includegraphics[width=0.30\linewidth]{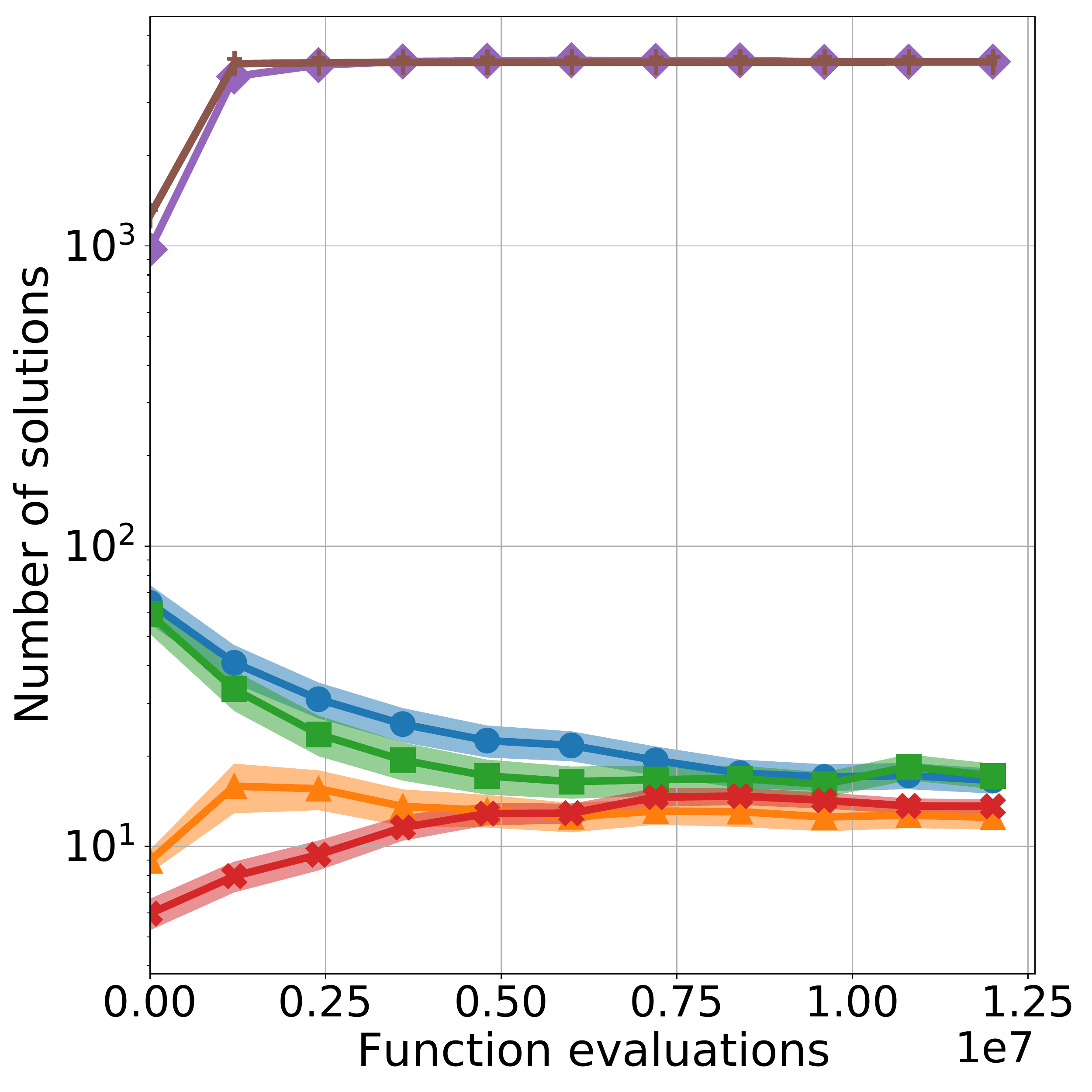}}
\subfloat[Average fitness]{\includegraphics[width=0.30\linewidth]{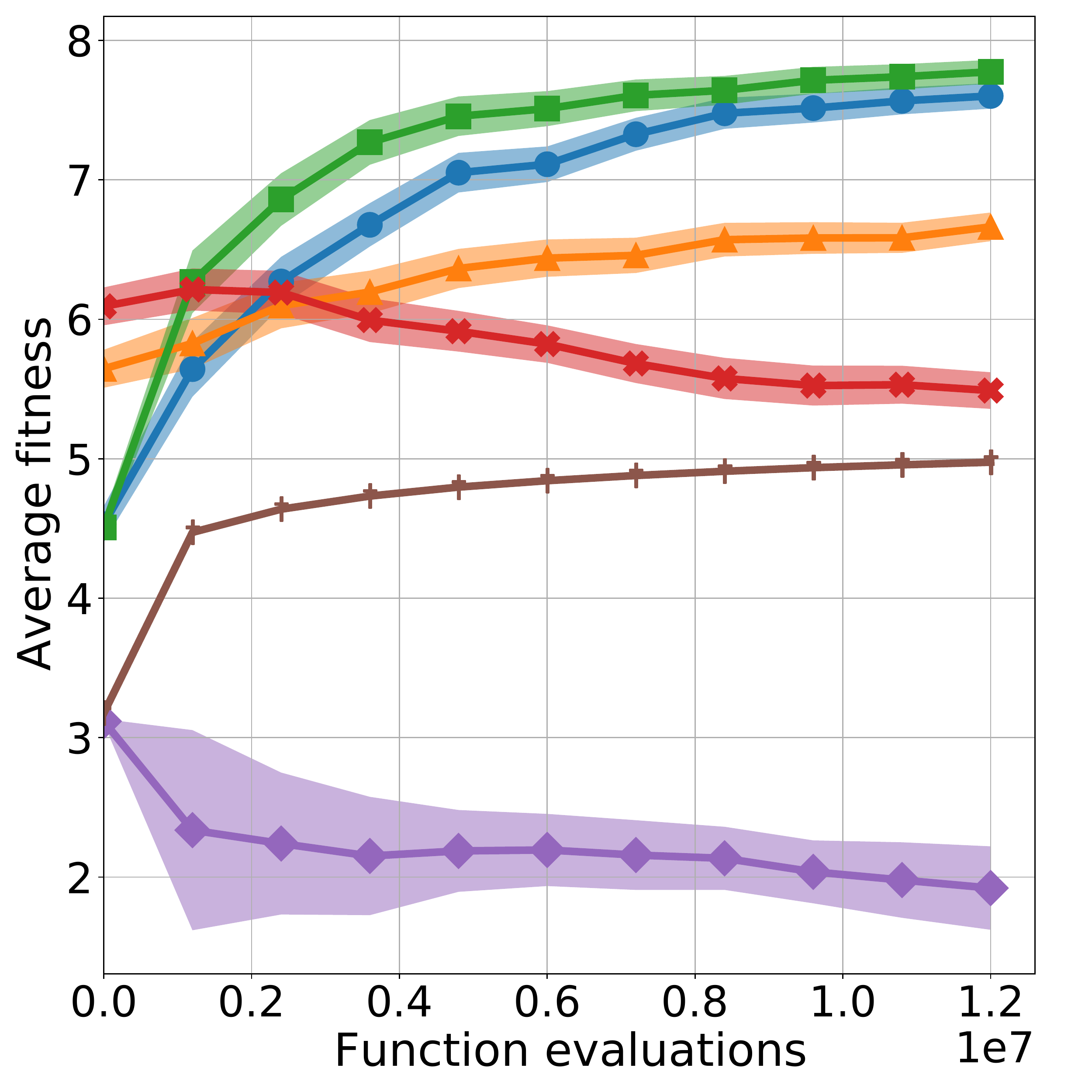}}
\subfloat[Global fitness]{\includegraphics[width=0.30\linewidth]{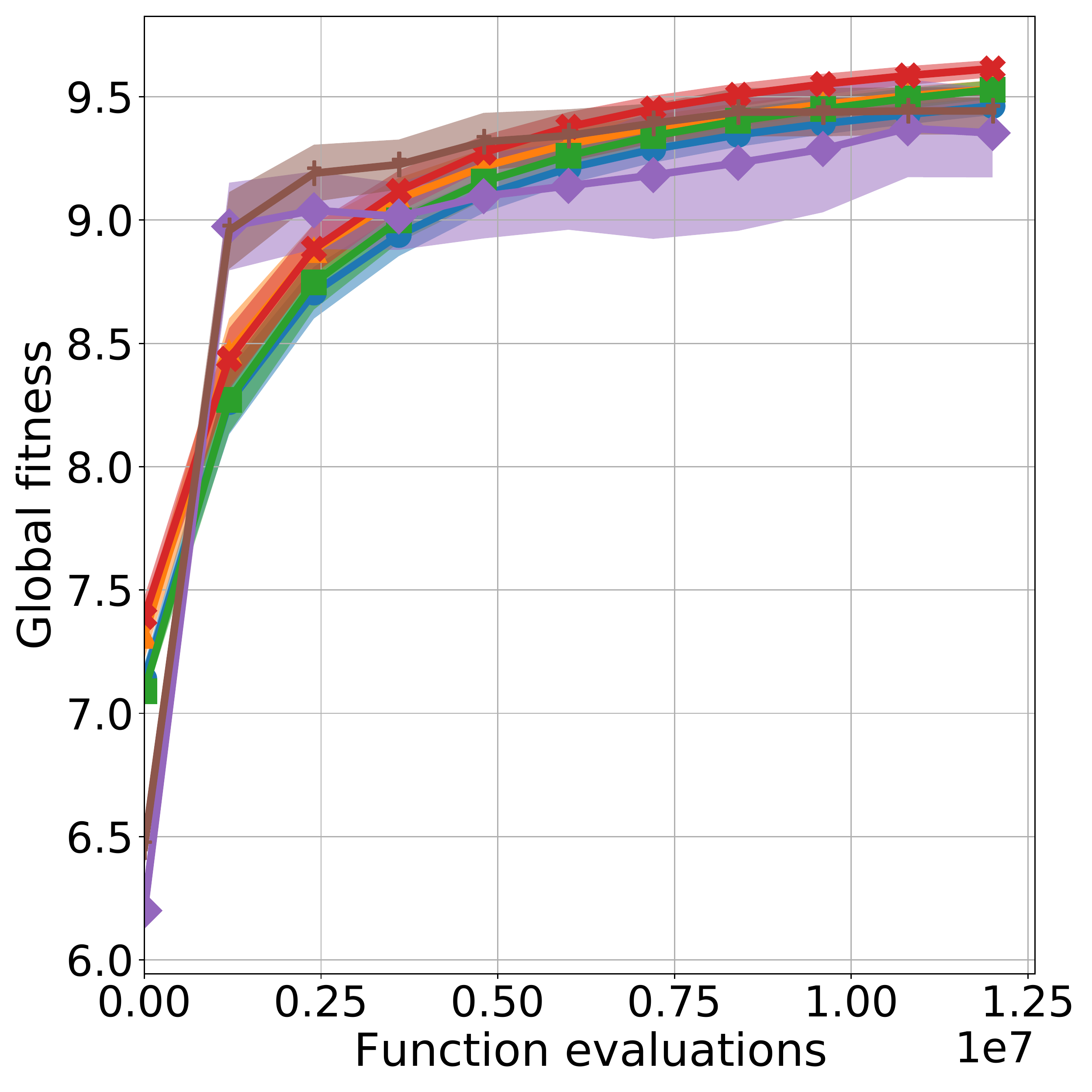}}\\
\caption{Quality-diversity statistics (Mean $\pm$ SE) of the different included QD algorithms across 4 replicates on the RHex robot platform, including \textbf{(a)} the total number of solutions in the archive; \textbf{(b)} the average fitness across the archive; and \textbf{(c)} the maximal fitness across the archive. For QD-Meta, Mean and SE statistics are aggregated across replicates and the different archives within the meta-population.} \label{fig: evolution-rhex}
\end{figure}

\begin{figure*}
\centering
\includegraphics[width=0.40\linewidth]{figures/METAvsCTRL_legend.pdf} \\
\subfloat[Obstacle test]{\includegraphics[width=0.30\linewidth]{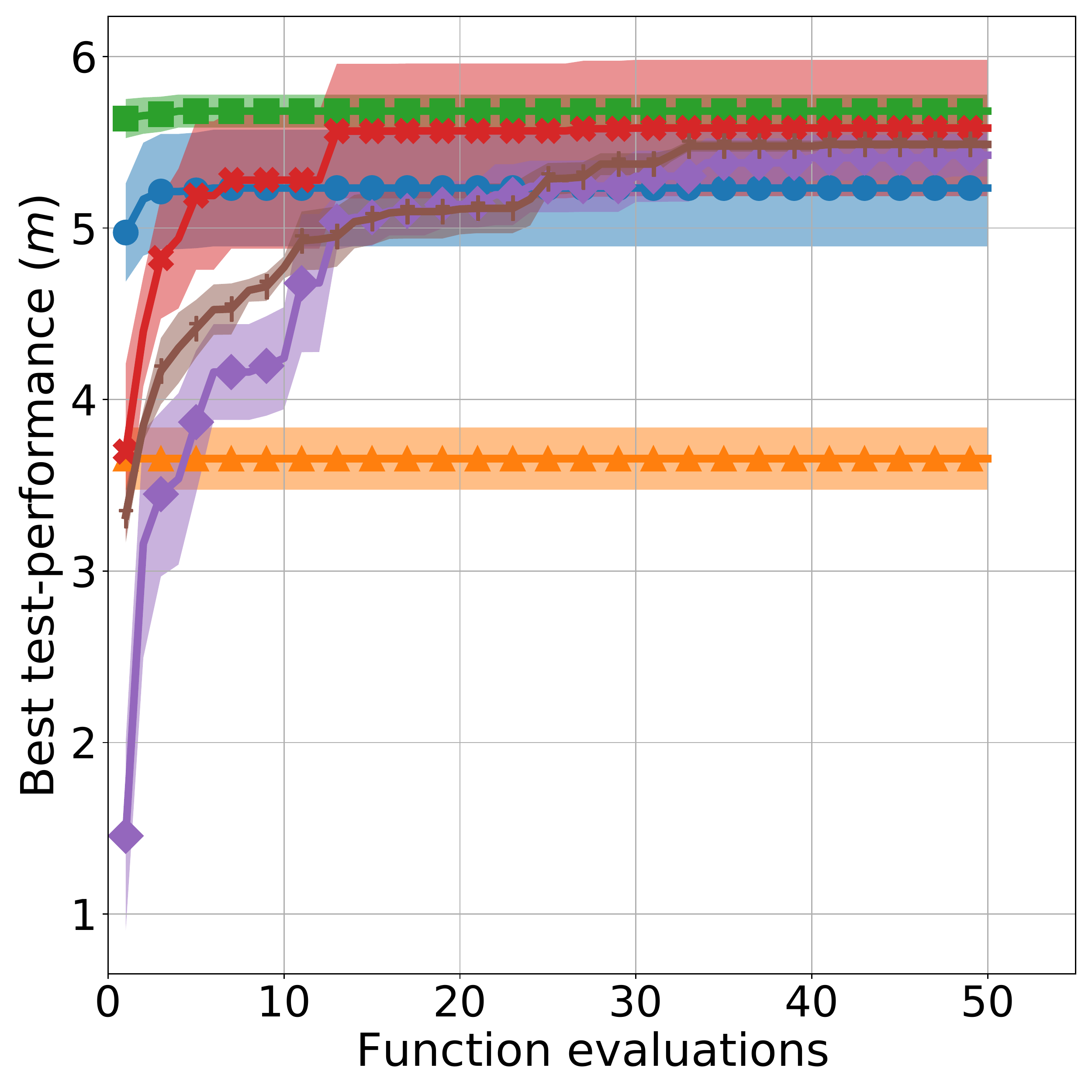}}
\subfloat[Damage test]{\includegraphics[width=0.30\linewidth]{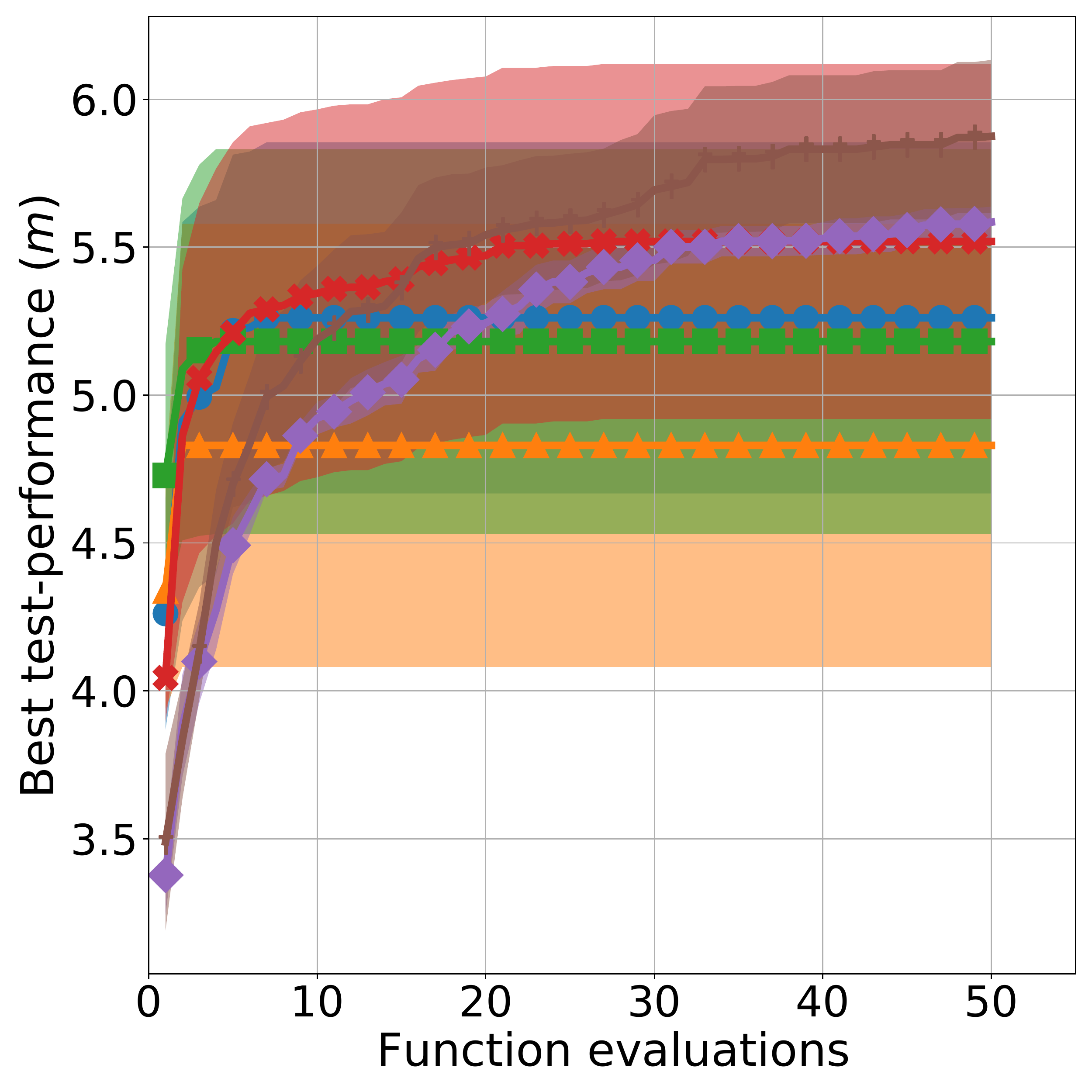}}
\caption{Test performance (Mean $\pm$ SE) of the different included QD algorithms on the RHex robot platform across 4 replicates. The $y$-axis shows the best solution so far after performing random search without replacement over the behavioural archive for the number of function evaluations indicated on the $x$-axis. For comparability, the lines are continued when the archive is exhausted, even though no further function evaluations are performed. For each replicate of QD-Meta, the archive with the highest meta-fitness at the end of meta-evolution is chosen.} \label{fig: adaptation-rhex}
\end{figure*}

As observable in Figure~\ref{fig: evolution-rhex}, rather than increasing the archive coverage, QD-Meta actually decreases the archive coverage over time, yielding around 100 controllers (compare to 3,000 to 4,000 controllers for AURORA and CVT-MAP-Elites). At the same time, QD-Meta increases the average fitness in the normal environment up to  between 6 and 8 meters, outperforming CVT-MAP-Elites ($\SI{5}{m}$) and AURORA ($\SI{2}{m}$). This points out the ability of QD-Meta to determine the optimal archive size for the specified meta-objective: since the behaviour space is optimised to allow high average performance of controllers across different obstacles and damages, QD-Meta results in smaller archives with only high-quality solutions. This effect is beneficial for adapting to a priori unknown environments in two ways. First, QD-Meta archives have improved performance on the generalisation test compared to CVT-MAP-Elites and AURORA (see Table~\ref{tab: test}). This improvement is seen whether or not the damages or obstacles were experienced in the meta-fitness evaluations. Second, in the recovery test, when performing a search across the archive for the best controller for a particular environment, the QD-Meta archives require only a few function evaluations to reach a near-optimal solution (see Figure~\ref{fig: adaptation-rhex}). However, a trade-off of the smaller archives is that QD-Meta archives do not always result in highest maximal end-performance after exhausting all the solutions; with a large number of solutions, even a low-quality archive can eventually yield a high maximum as long as there is one solution performing highly for that particular environmental change. \\

A qualitative analysis using slow-motion video material (see \url{http://tiny.cc/QD-Meta}) demonstrates the different behaviour of QD-Meta Obstacle conditions compared to QD-Meta Damage conditions. In particular, the behaviours evolved by QD-Meta Damage appear to less balanced with more roll motion. In the exemplary videos included for QD-Meta Damage (see videos with QD-Meta-Damage prefix), one of the legs  has its stance phase facing upwards, which is comparable to disabling the leg as if it is damaged. In the exemplary videos included for QD-Meta Obstacle (see videos with QD-Meta-Obstacle prefix), one can observe jumpy behaviours, which help to navigate over  obstacles such as pipes and stairs. The mentioned QD-Meta Damage behaviours are rather specific in comparison; as they seem to disable one or more legs, they result in shaky behaviours that particularly perform well when the disabled leg is damaged but that do not generalise well to other environments. This explains why QD-Meta Damage does not generalise as well as QD-Meta Obstacle to new obstacles or new damages (see Table \ref{tab: test}). In particular, one can observe the larger generalisation drop from Training to Test for Damage conditions and the consistent top ranking of QD-Meta Obstacle Feature-sets.

A further analysis reveals the nature of the benefits of dynamic parameter control with RL. A meta-fitness comparison (see Fig.~S4 in Supplementary materials) finds that the benefit is typically limited although the Obstacle Trajectory condition with RL obtains larger and smoother meta-fitness improvements than the version without RL. By further analysing each run separately and comparing the evolution of the meta-fitness to the  controlled parameter (i.e. the generations per meta-generation; see Fig.~S5 in Supplementary), a pattern is observed in which the number of generations is high initially, declines sharply, and finally levels off or increases slightly as evolution progresses. This is in line with the hypothesis that initially the algorithm should generate a large number of behaviourally diverse and high-performing solutions -- using ME iterations -- to be able to get reliable evaluations of the behaviour space -- using meta-fitness evaluations. Low (resp. high) meta-fitness after the initial phase results in a low (resp. high) ratio of generations per meta-generation. This indicates that: for regions around the global optimum, high-reliability evaluations are required; for local optima, evaluating many feature-maps helps to rapidly escape to a higher-performing region.

\begin{table*}
\caption{Generalisation test performance (Mean $\pm$ SD) on the training and test environments of the RHex robot platform. Bold indicates the best performance for the given environment. Performance is the average across the solutions in the archive and the different included test environments. Mean and standard deviation are aggregated across 4 replicates. For each replicate of QD-Meta, the archive with the highest meta-fitness at the end of meta-evolution is chosen. Underline indicates the obstacles or damages are experienced during meta-evolution.} \label{tab: test}
\resizebox{0.99\textwidth}{!}{
\begin{tabular}{l  l l l l l l l l l l l l l l l l l l }
\toprule
\textbf{Environment}					& \multicolumn{6}{l}{\textbf{Condition}} \\
 \midrule
 & \multicolumn{4}{l}{\textbf{QD-Meta}} & \multicolumn{2}{l}{\textbf{Baseline}} \\
 & \multicolumn{1}{l}{Damage Feature-sets} & \multicolumn{1}{l}{Damage Trajectory} & \multicolumn{1}{l}{Obstacle Feature-sets} & \multicolumn{1}{l}{Obstacle Trajectory} & \multicolumn{1}{l}{AURORA} & \multicolumn{1}{l}{CVT-ME} \\ 
\hline
Training-Damages & $ \underline{6.779 \pm 0.89} $ &$ \underline{\mathbf{6.882 \pm 1.06}}$ &$ 6.876 \pm 1.14 $ &$ 5.093 \pm 1.26 $ &$ 3.072 \pm 0.19 $ &$ 4.451 \pm 0.25 $ \\
Training-Obstacles & $3.789 \pm 0.77$ &$3.017 \pm 0.90$ &$\underline{\mathbf{4.059 \pm 0.66}}$ &$\underline{3.895 \pm 0.45}$ &$2.060 \pm 0.11$ &$ 2.660 \pm 0.11 $ \\
Test-Damages & $ 4.491 \pm 0.99 $ &$ 4.193 \pm 0.66 $ &$\mathbf{4.782 \pm 0.88}$ &$ 4.063 \pm 0.64 $ &$ 3.404 \pm 0.11 $ &$ 3.422 \pm 0.24 $ \\
Test-Obstacles & $ 3.628 \pm 0.50 $ &$ 2.733 \pm 0.61 $ &$\mathbf{4.161 \pm 0.72}$ &$ 2.831 \pm 0.20 $ &$ 2.400 \pm 0.26 $ &$ 2.642 \pm 0.16 $ \\
\bottomrule
\end{tabular}
}
\end{table*}

\section{Discussion}
This paper demonstrates QD-Meta as a promising framework for the automated design of low-dimensional behaviour spaces for QD optimisation. We present results showing that QD-Meta archives improve rapid adaptation to fitness landscapes not experienced during evolution, further establishing QD-Meta as a widely applicable QD framework with state-of-the-art performance. QD-Meta compares favourably to state-of-the-art QD algorithms for automated behaviour spaces, namely AURORA \cite{Grillotti2021a} and CVT-MAP-Elites \cite{Vassiliades2018b}, not only in adaptation to new environments but also on traditional QD metrics; that is, the average fitness and the number of solutions  are  improved when they are rewarded for in the meta-fitness.

The findings  on function optimisation show that QD-Meta allows to evolve archives that are robust to dimensionality decreases and translations. This is of interest on a wide variety of applications as the data that an algorithm is trained on does not always reflect the real distribution of the data or the function may even change dynamically over time. While both meta-fitness formulations in the Rastrigin have a positive effect on both tests, the effect of the drop of dimensionality of the inputs is stronger. One interpretation is that performing dimensionality reductions in the meta-fitness has an effect on the feature-map similar to DropOut \cite{Srivastava2014}, which improves generalisation in neural networks by randomly dropping out the activation of a neuron, typically one of the input neurons.

Traditionally, quality-diversity optimisation has assumed that a large number of solutions is better. However, smaller archives can provide key benefits for adaptation and greater transparency. In this context, QD-Meta has the unique benefit of being able to tailor the behaviour space, including the number of solutions as well as the behavioural features, to a meta-objective rather than being unsupervised \cite{Grillotti2021a} or unadaptive \cite{Vassiliades2018b}. This benefit can be exploited to suit the domain. If an end-user is interested in multi-modal function optimisation domains, and in particular covering the peaks of the fitness function despite dynamic changes, then the meta-fitness could provide a bonus for archive size as in the present Rastrigin experiments. In domains such as multi-legged robot locomotion, the same gait can overcome different types of obstacles near-optimally, and therefore only a few solutions may be required so a meta-fitness bonus for archive size is not necessary. Indeed, in the RHex robot study, QD-Meta demonstrates the benefits of having a lower number of solutions with strong generalisation across different environments: 1) all of the solutions have a high performance even when the environment changes so there is often no need for adaptation; 2) all the controllers can safely be deployed during adaptation without too much risk of low performance or otherwise unsafe behaviours; and 3) as there is only a need to try a dozen solutions, the archives allow rapid adaptation. A further benefit of smaller archives may be in transparency and verification: if the number of solutions in the archive is limited, then the end-user can easily verify and assess the solutions to prevent undesirable consequences from their application. That said, if the number of trialled solutions is not of critical concern but instead the maximal performance, then a meta-fitness bonus for larger archives would be preferable.

In previous work QD-Meta relied on hand-crafted feature-sets. While in most robotics applications feature-sets can be given to the user or constructed with only minor effort, the trajectory of the robot is a generic choice that requires limited to no prior knowledge; for this reason it has been the base-behavioural space of choice in methods such as AURORA and CVT-MAP-Elites. The present study shows that while there is a benefit to feature-sets, QD-Meta can achieve a comparable performance using the robot's trajectories. A difficulty observed is that the meta-optimisation problem becomes more challenging due to the increased dimensionality and the authors recommend research into scaling up the approach to even higher-dimensional base-behavioural spaces, such as (sequences of) raw pixel data. To improve database storage capacity as well as the optimisation of the feature-map in the meta-genotypic space, one suggested approach is to first apply traditional dimensionality reduction methods \cite{Sorzano2014} to yield a low-dimensional ``latent'' base-behavioural space which is then provided as input to the feature-map.

Evolving a meta-population of QD archives rather than a single meta-individual is shown to be essential for the search for high meta-fitness feature-maps, with meta-fitness and QD metrics improving as the meta-population size is increased to 10. Two explanations are proposed for its benefits. First, population-based optimisation algorithms are less prone to get stuck in local optima due to being distributed across the search space; this is crucial in large multi-modal search spaces such as the 522-dimensional space of feature-maps in QD-Meta. Second, each meta-individual comes with its own behaviour space, affecting which elites are selected for reproduction, and therefore QD-Meta generates new solutions from different sources. Beyond the earlier meta-evolution works \cite{Bossens2020a,Bossens2021a}, a similar reasoning can be found in recent works in the literature. Multi-emitter MAP-Elites uses different ``emitters'', each of which provide a different reproduction operator based on different heuristics to explore the search space, for example based on fitness or a random direction in the behaviour space \cite{Cully2021}. Multi-container AURORA maintains a number of feature-maps at the same time to yield improvements to the QD-score, which is the $[0,1]$-normalised archive-summed fitness \cite{Cazenille2021}. These improvements were minor compared to the average fitness improvements in our Rastrigin function optimisation results. Future research is advised to compare and integrate these strategies to yield new insights into QD optimisation.
\section{Conclusion}
This paper demonstrates Quality-Diversity Meta-evolution as a promising framework for quality-diversity optimisation. Quality-diversity metrics as well as tests on changed fitness landscapes demonstrate that QD-Meta outperforms CVT-MAP-Elites as well as AURORA, two state-of-the-art algorithms for automated behaviour spaces from high-dimensional trajectories. The study shows improved adaptation to linear transformations and dimensionality changes in function optimisation as well as adaptation of multilegged locomotion gaits to various robot damages and various obstacle courses. Compared to existing methods, QD-Meta derives its adaptation capability to customising the behaviour space to a generalisation-based meta-objective. The study also demonstrates that QD-Meta can yield similar benefits with high-dimensional observation trajectories without requiring the user to hand-craft feature-sets. For future research, an investigation into further scalability to even higher-dimensional base-behaviour spaces (e.g. raw pixel data streams) is advised, as is a study integrating and comparing multi-emitter and multi-container QD algorithms to QD-Meta variants.

\section*{Acknowledgements}
This work has been supported by the Engineering and Physical Sciences Research Council under the New Investigator Award grant, EP/R030073/1, and the UKRI Trustworthy Autonomous Systems Hub, EP/V00784X/1. The authors acknowledge the use of the IRIDIS High Performance Computing Facility.

\bibliographystyle{IEEEtran}
\bibliography{bibliography}

\end{document}


\thispagestyle{empty}
\pagestyle{empty}
\maketitle 

\section{Effect of population size on meta-evolution in the Rastrigin function optimisation benchmark}
Manipulation of the population size demonstrates that using $\lambda=1$ yields meta-fitness of only 900,000, which is about 50\% of the other settings with $\lambda \in \{2,5,10,20\}$, which have nearly 2,000,000 in meta-fitness. Similarly, the average fitness and number of solutions are also decreased when taking $\lambda=1$. Between different settings of $\lambda > 1$,  $\lambda=10$ is consistently ranked on top.
\begin{figure*}[htbp!]
\centering
\includegraphics[width=0.3\linewidth]{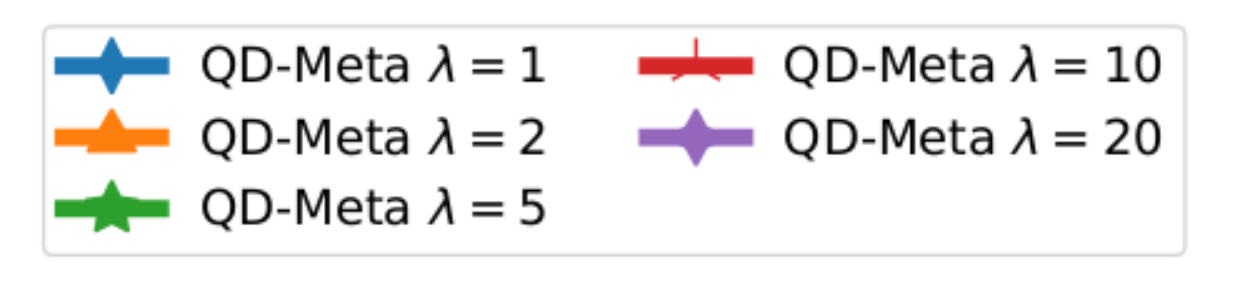}\\
\subfloat[Number of solutions]{\includegraphics[width=0.24\linewidth]{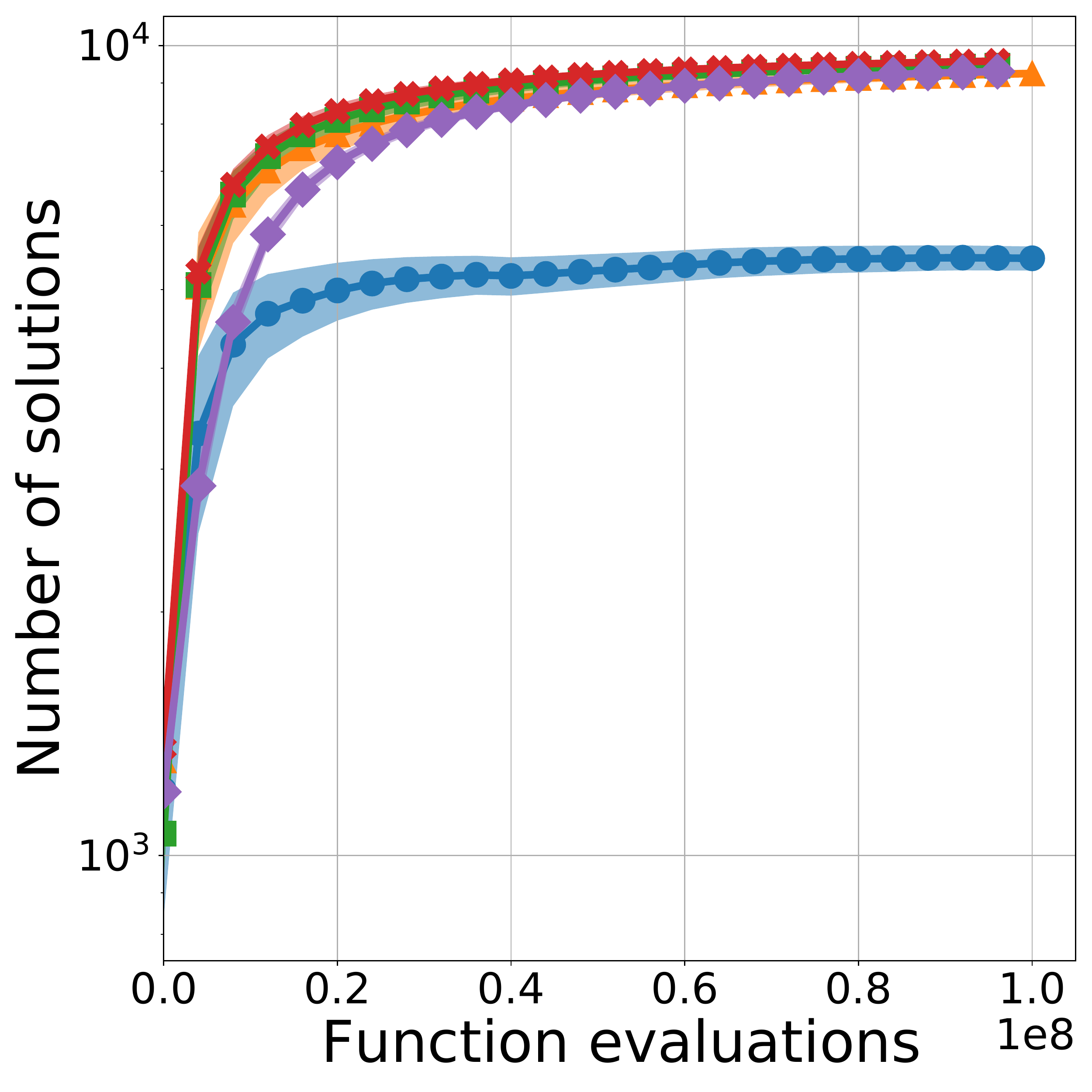}}
\subfloat[Average fitness]{\includegraphics[width=0.24\linewidth]{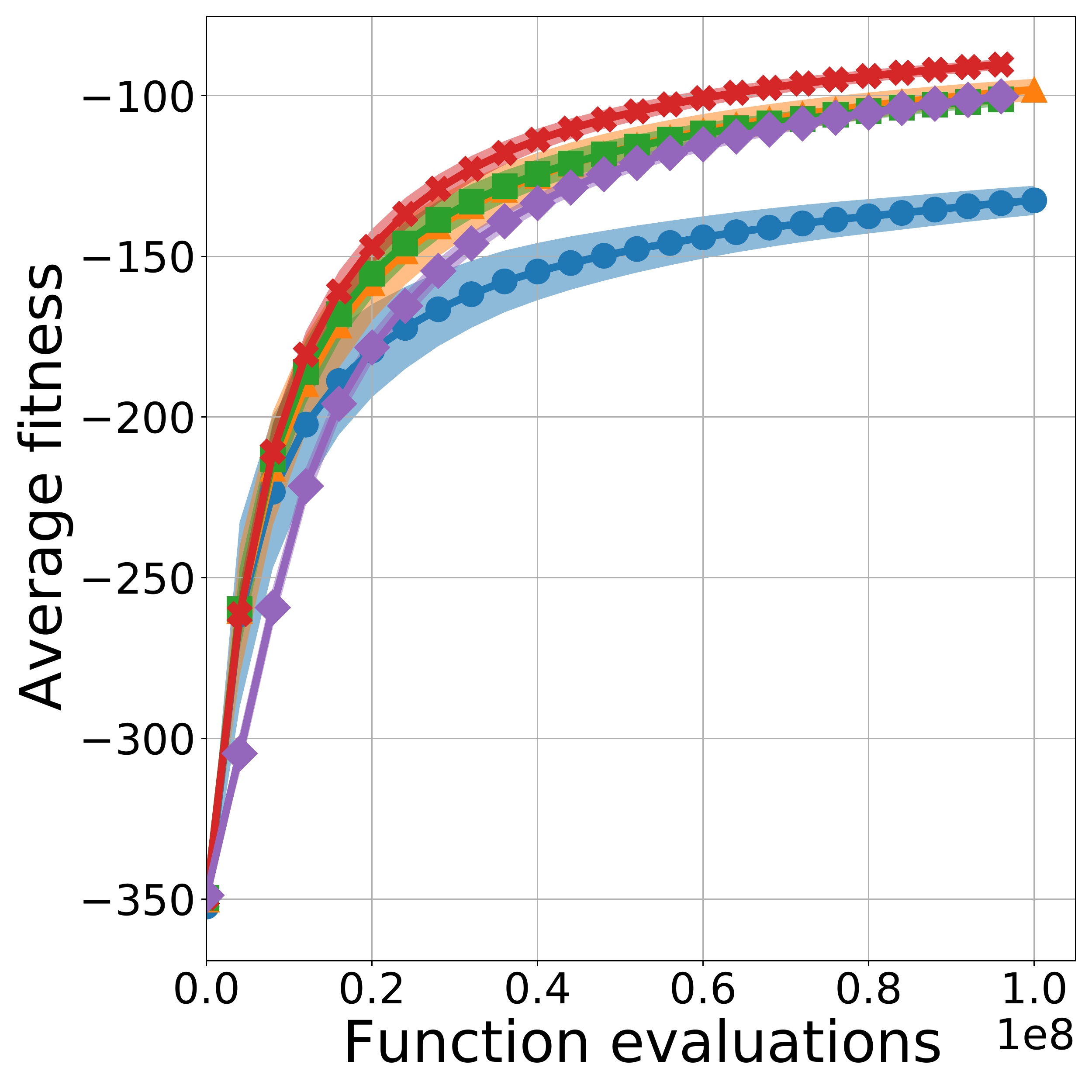}} 
\subfloat[Global fitness]{\includegraphics[width=0.24\linewidth]{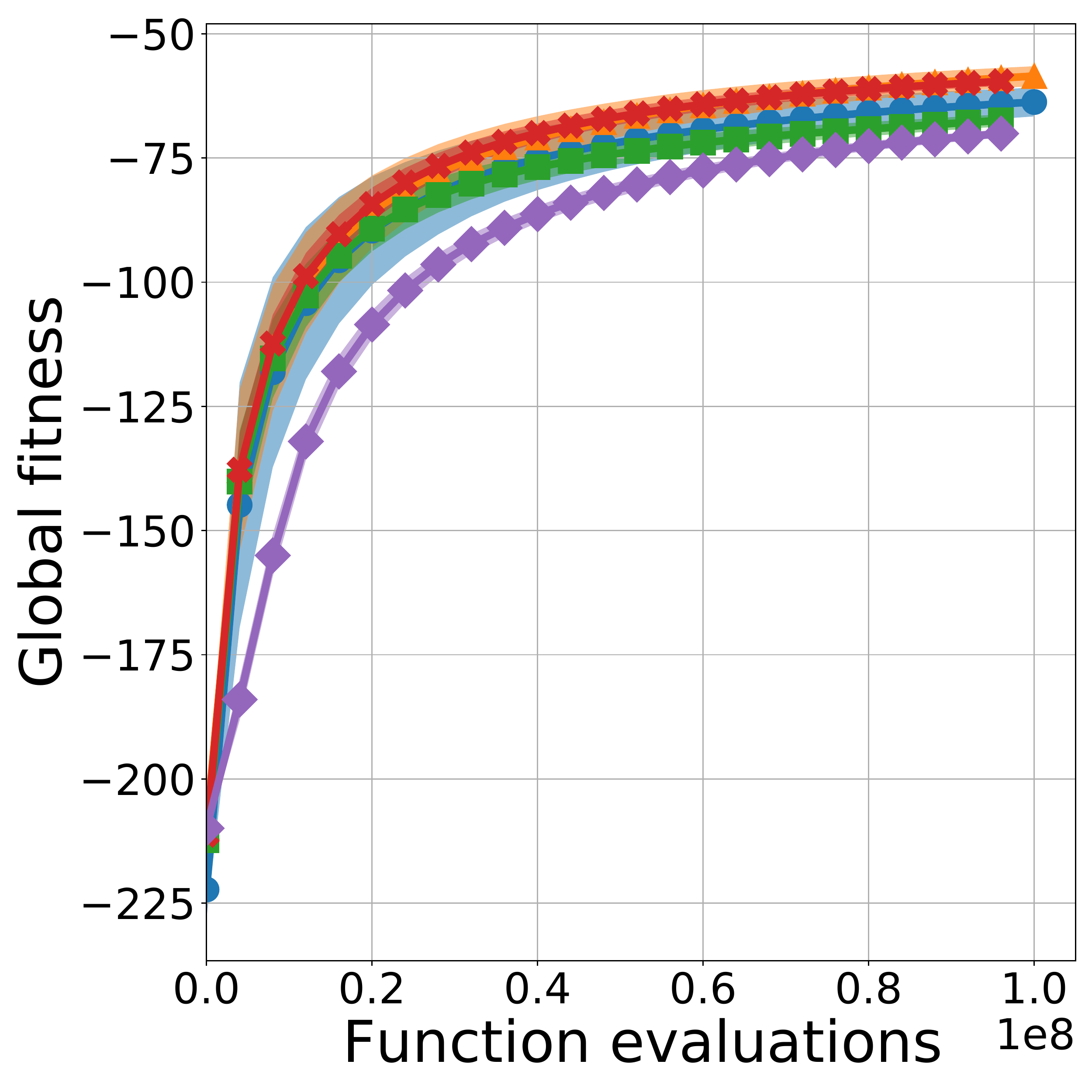}}
\subfloat[Meta-fitness]{\includegraphics[width=0.24\linewidth]{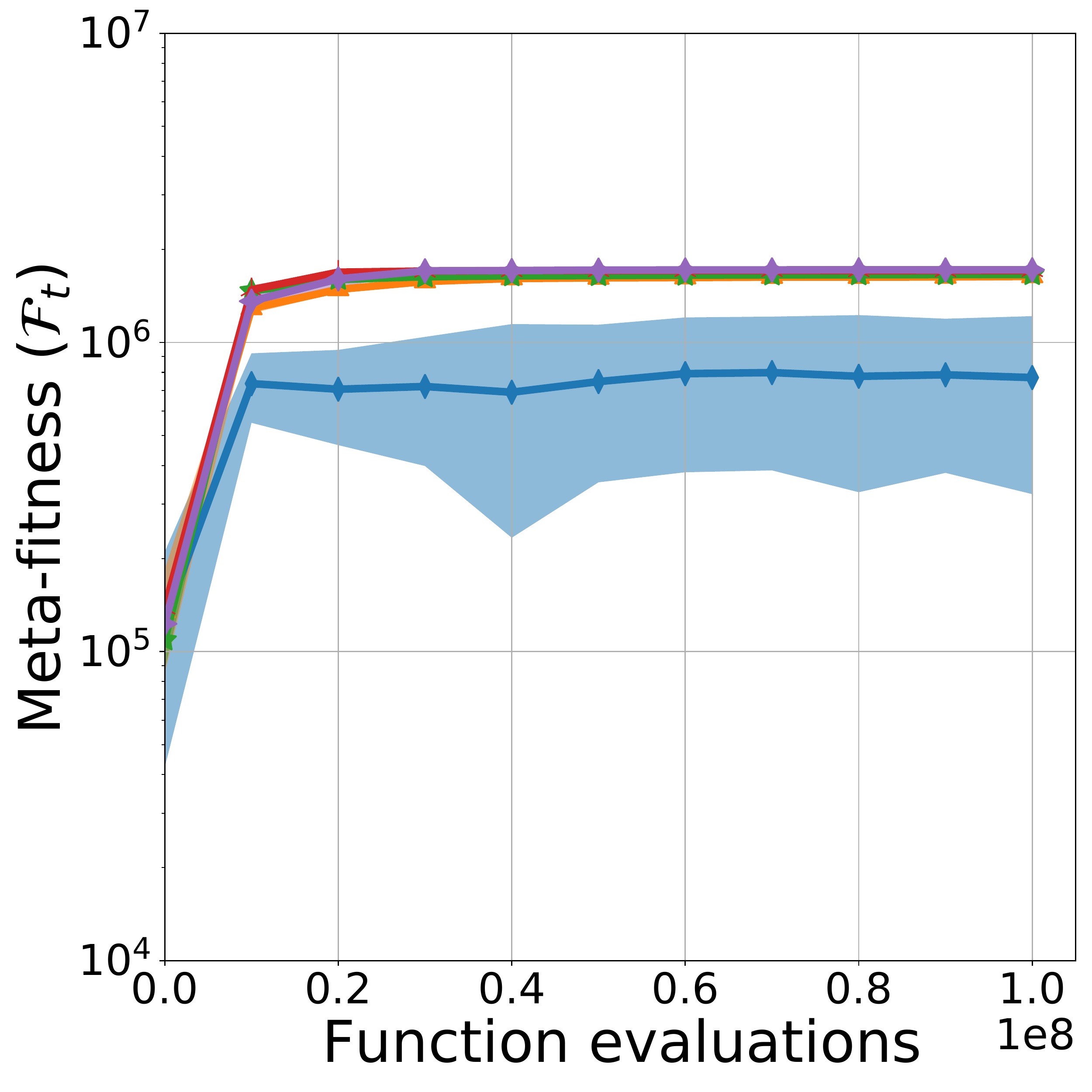}} \\
\caption{\small{Effect of meta-population size, $\lambda \in \{1,3,5,10,20\}$, on meta-fitness (Mean $\pm$ SE) in Rastrigin function optimisation. $x$-axis is the number of function evaluations. $y$-axis is the meta-fitness average across the archives in the meta-population, with the mean and standard error being aggregated across 20 replicates.}} \label{fig: poprecovery_m}
\end{figure*}

\newpage 

\section{RHex hexapod robot locomotion benchmark}
\subsection{Full set of obstacle courses}
\begin{figure}[htbp!]
\subfloat[Sphere]{\includegraphics[width=0.19\linewidth]{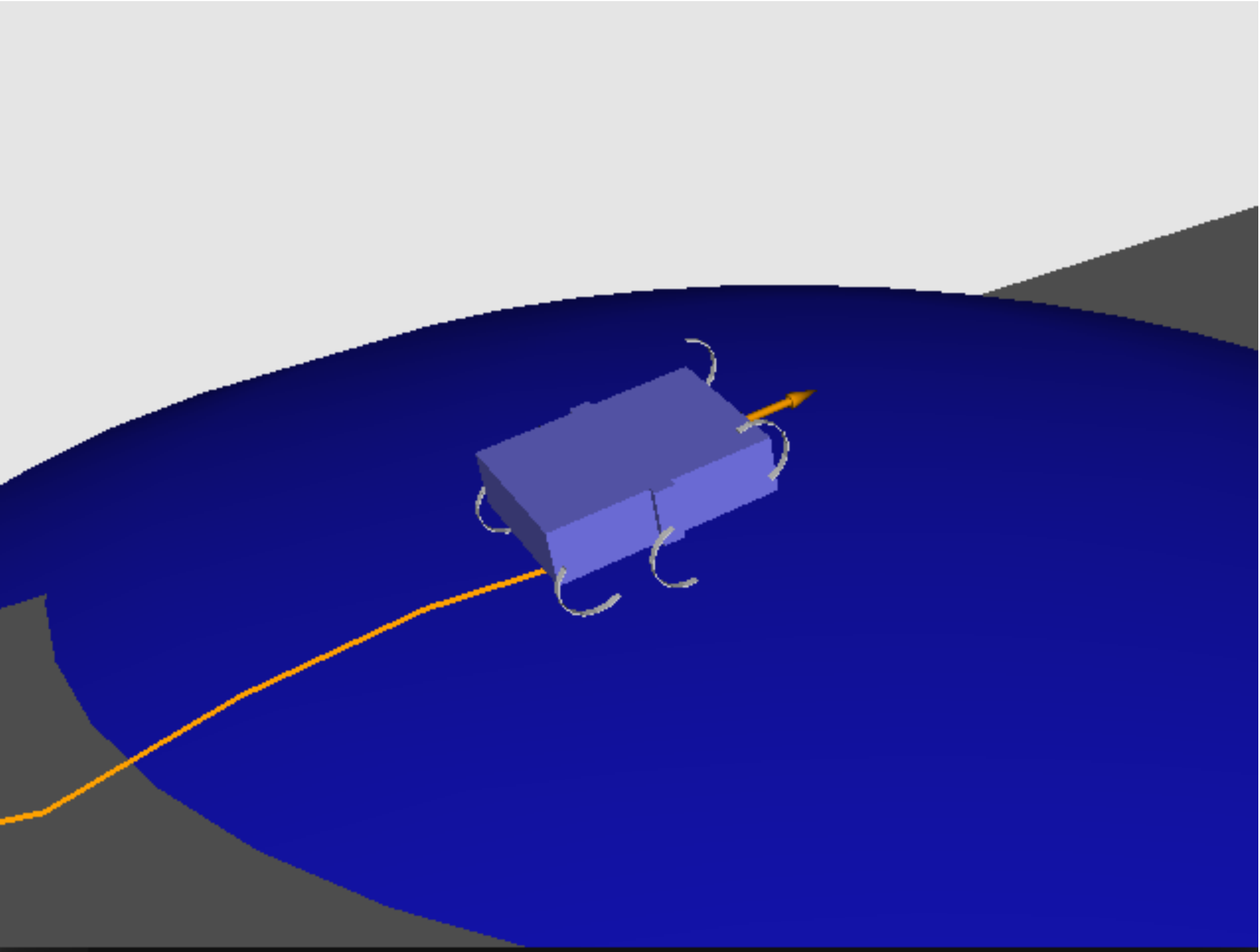}} \hfill
\subfloat[Stairs]{\includegraphics[width=0.19\linewidth]{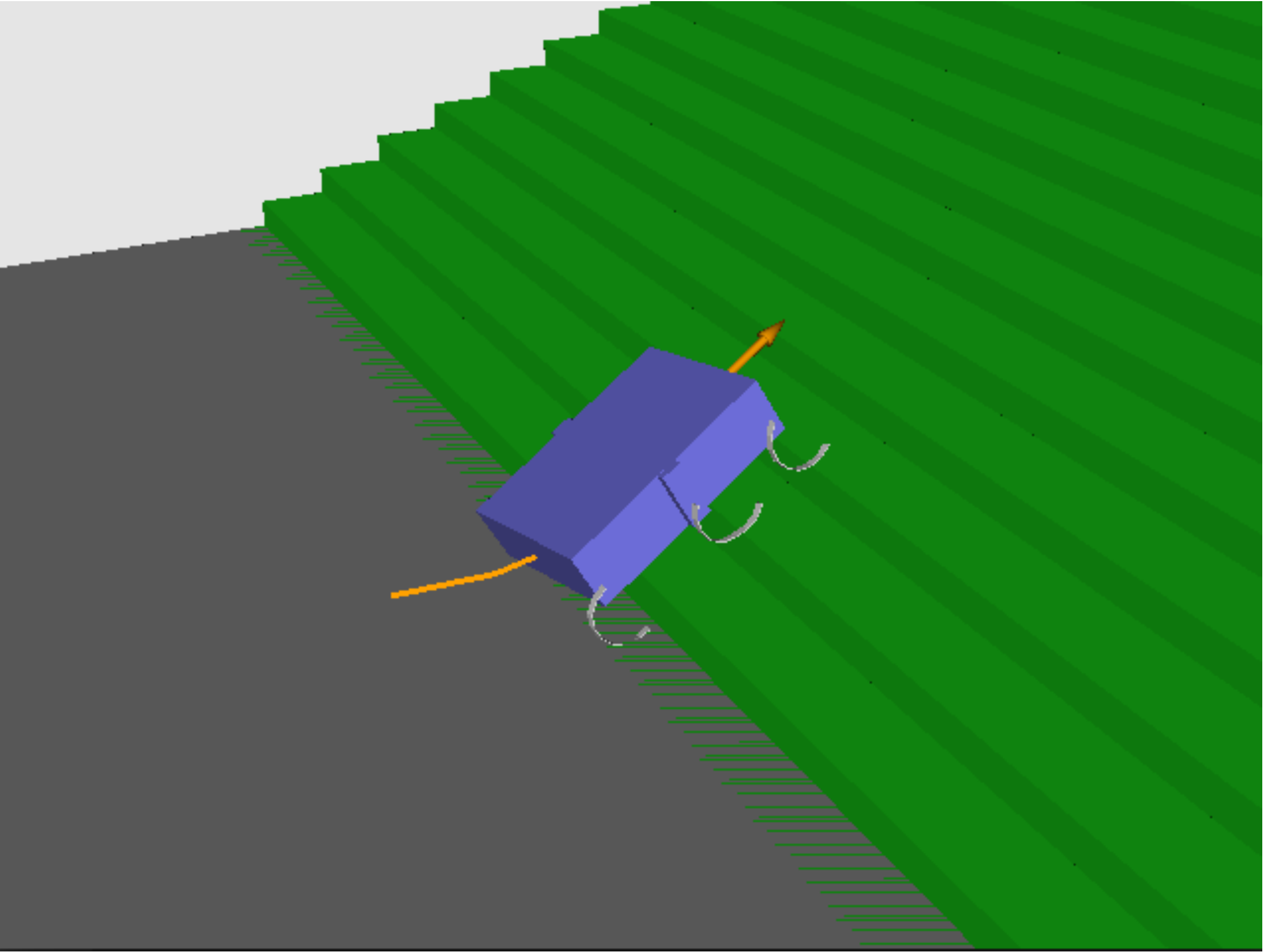}} \hfill
\subfloat[Slope]{\includegraphics[width=0.19\linewidth]{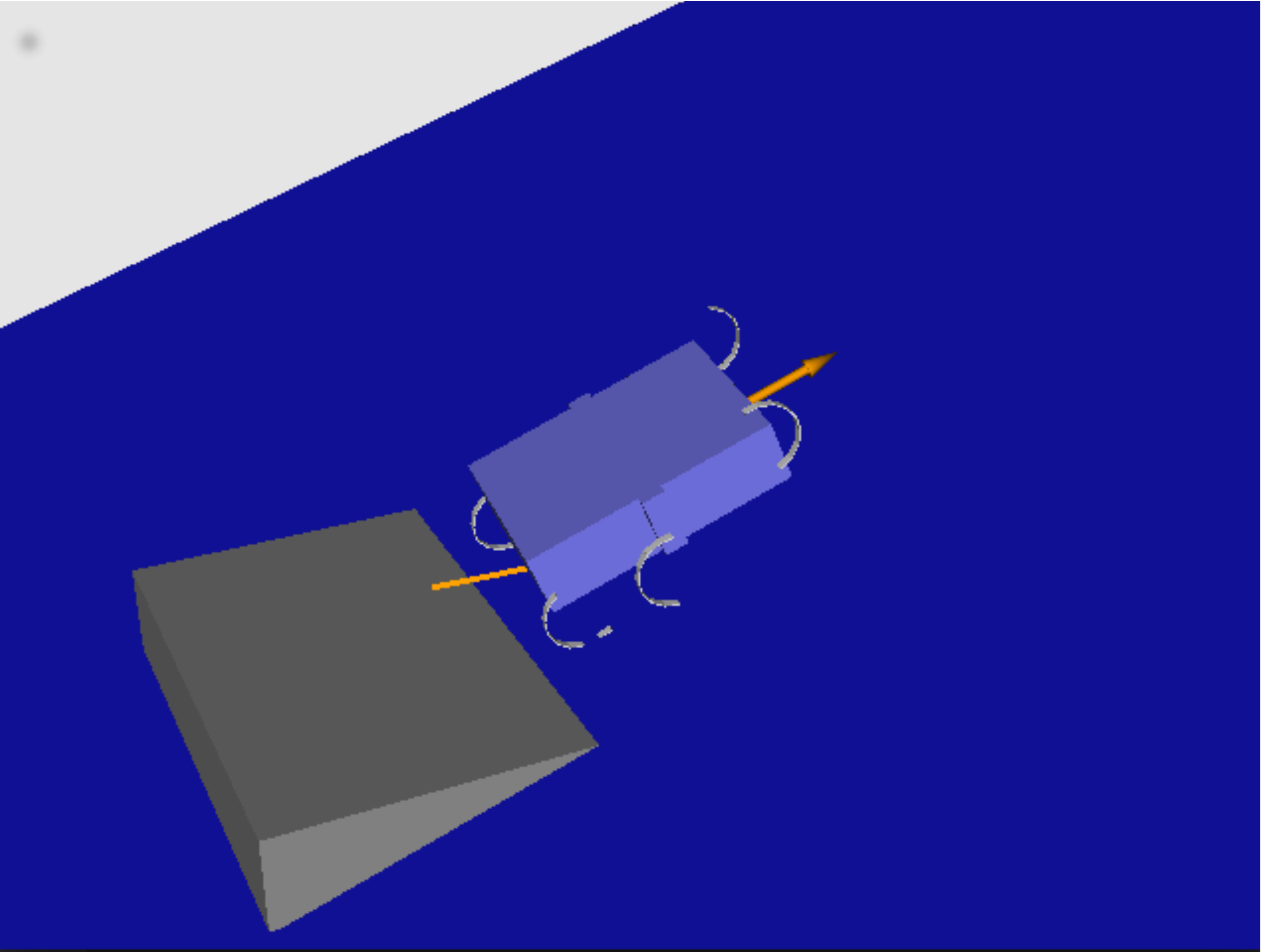}} \hfill
\subfloat[Rubble]{\includegraphics[width=0.19\linewidth]{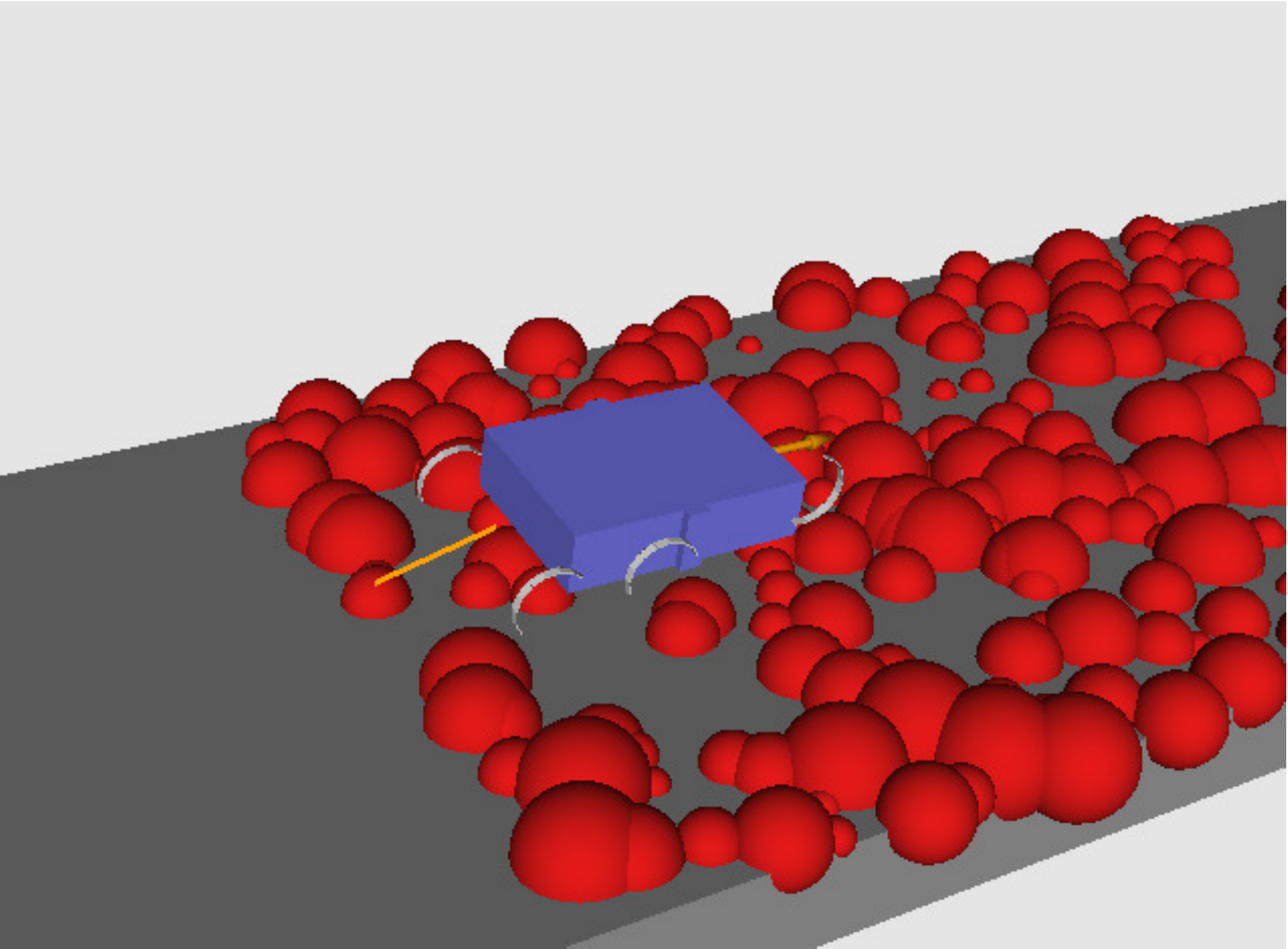}} \hfill
\subfloat[Down-and-up stairs]{\includegraphics[width=0.19\linewidth]{figures/obstacles/obstacle_downstairs.pdf}}\\
\subfloat[Many pipes]{\includegraphics[width=0.19\linewidth]{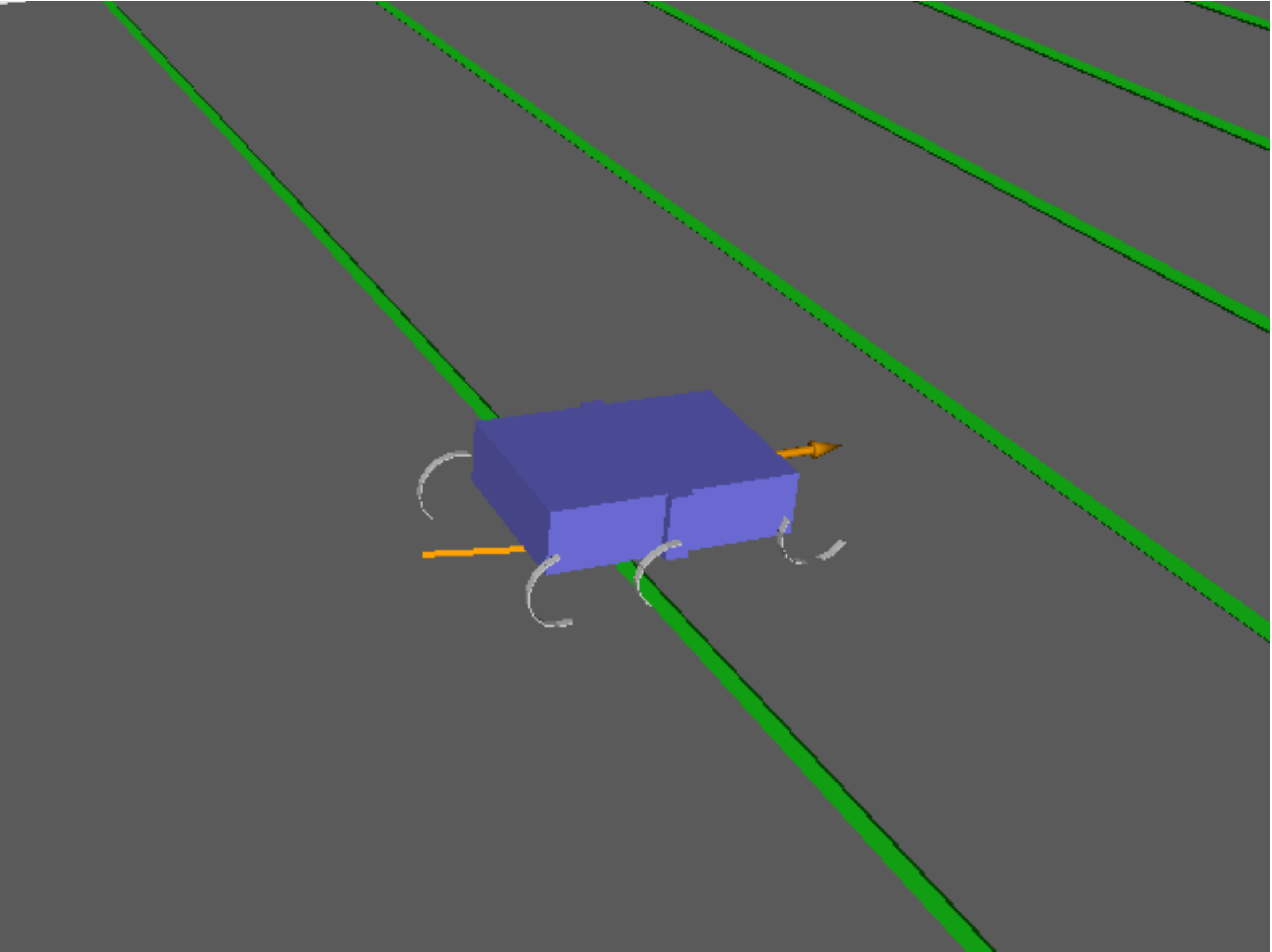}} \hfill
\subfloat[Ditch]{\includegraphics[width=0.19\linewidth]{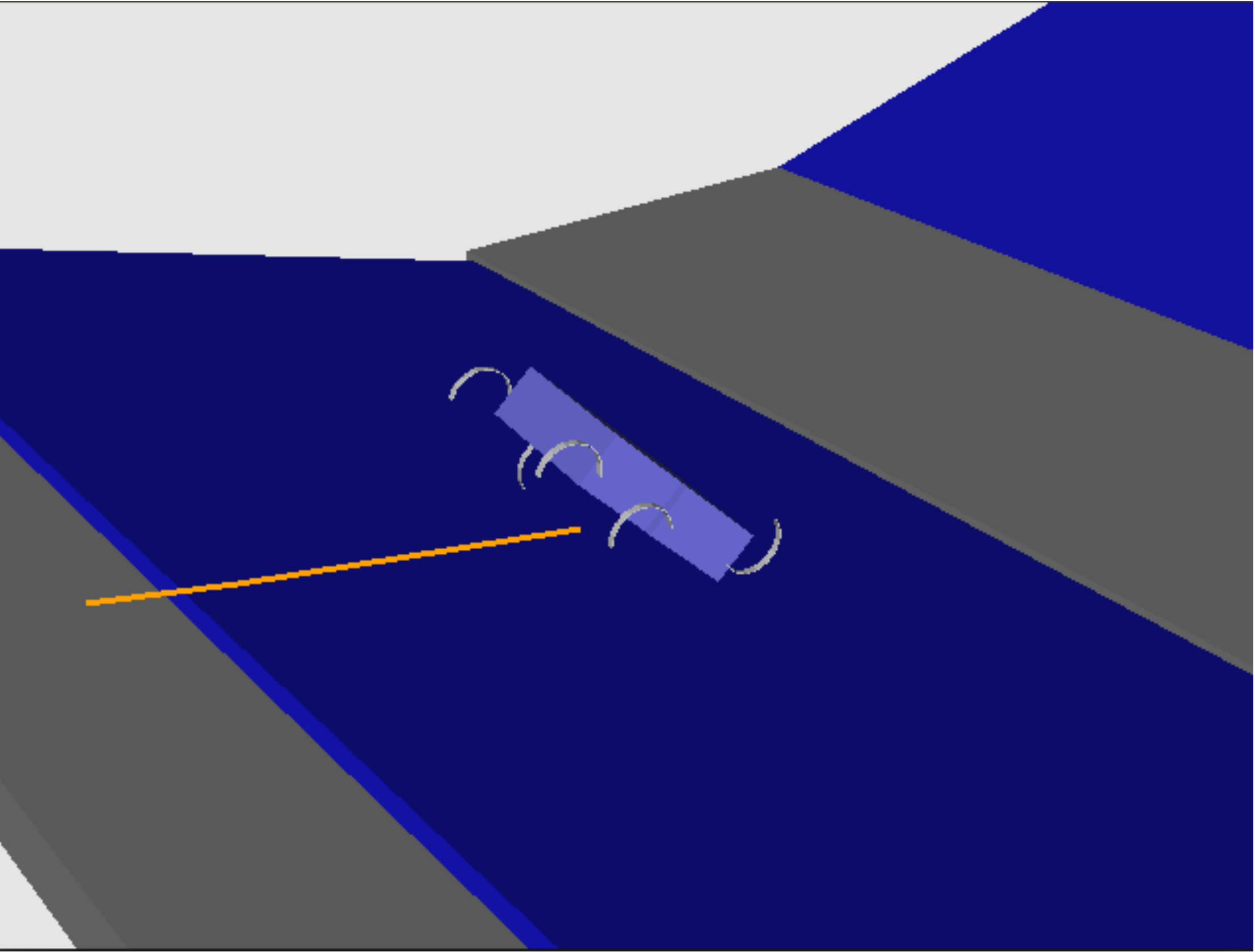}} \hfill
\subfloat[Thick pipe]{\includegraphics[width=0.19\linewidth]{figures/obstacles/obstacle_thickpipe.pdf}} \hfill
\subfloat[Thin pipe]{\includegraphics[width=0.19\linewidth]{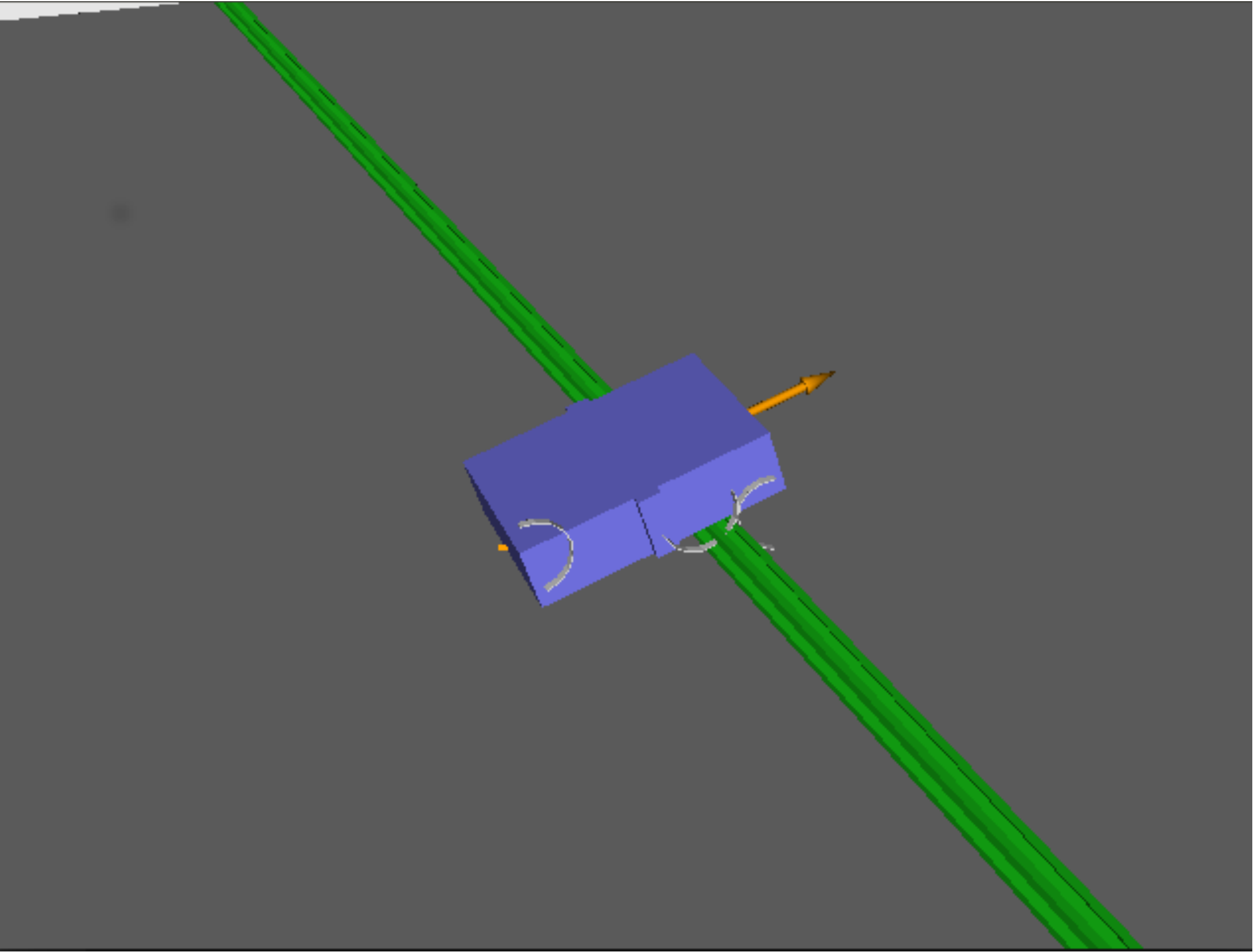}}\\
\caption{All obstacles used for the simulated obstacle course. At the start of an evolutionary run of QD-Meta, 5 of these obstacle courses are selected randomly to be used in the meta-fitness evaluation. For each run, the remaining 4 obstacle courses, which are not experienced during (meta-)evolution, are used in the test phase.} \label{fig: obstacle-course}
\end{figure}

Fig.~\ref{fig: obstacle-course} shows the full set of obstacle courses. For each evolutionary replicate, one partition of the set is used for meta-fitness evaluation in the QD-Meta Obstacle condition while the remaining partition of the set is used for the test phase of all algorithms.
\subsection{Results for dynamic parameter control in QD-Meta}
\begin{figure*}[htbp!]
\centering
\includegraphics[width=0.40\linewidth]{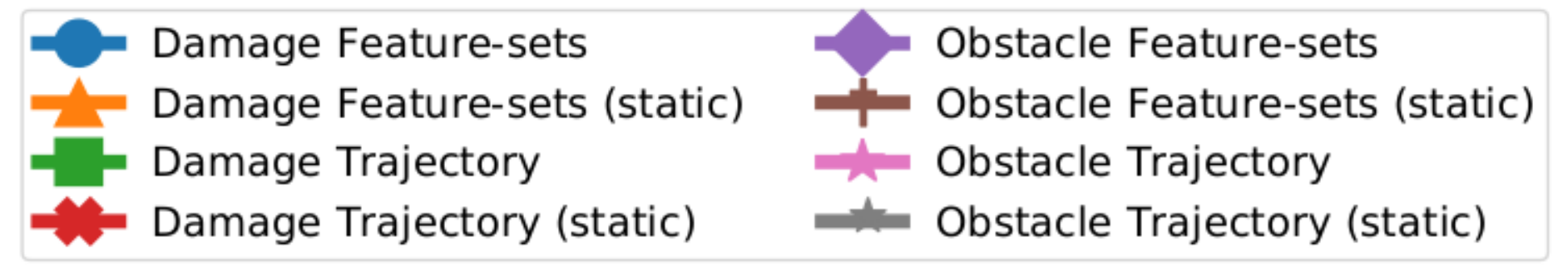}
\subfloat[Number of solutions]{\includegraphics[width=0.30\linewidth]{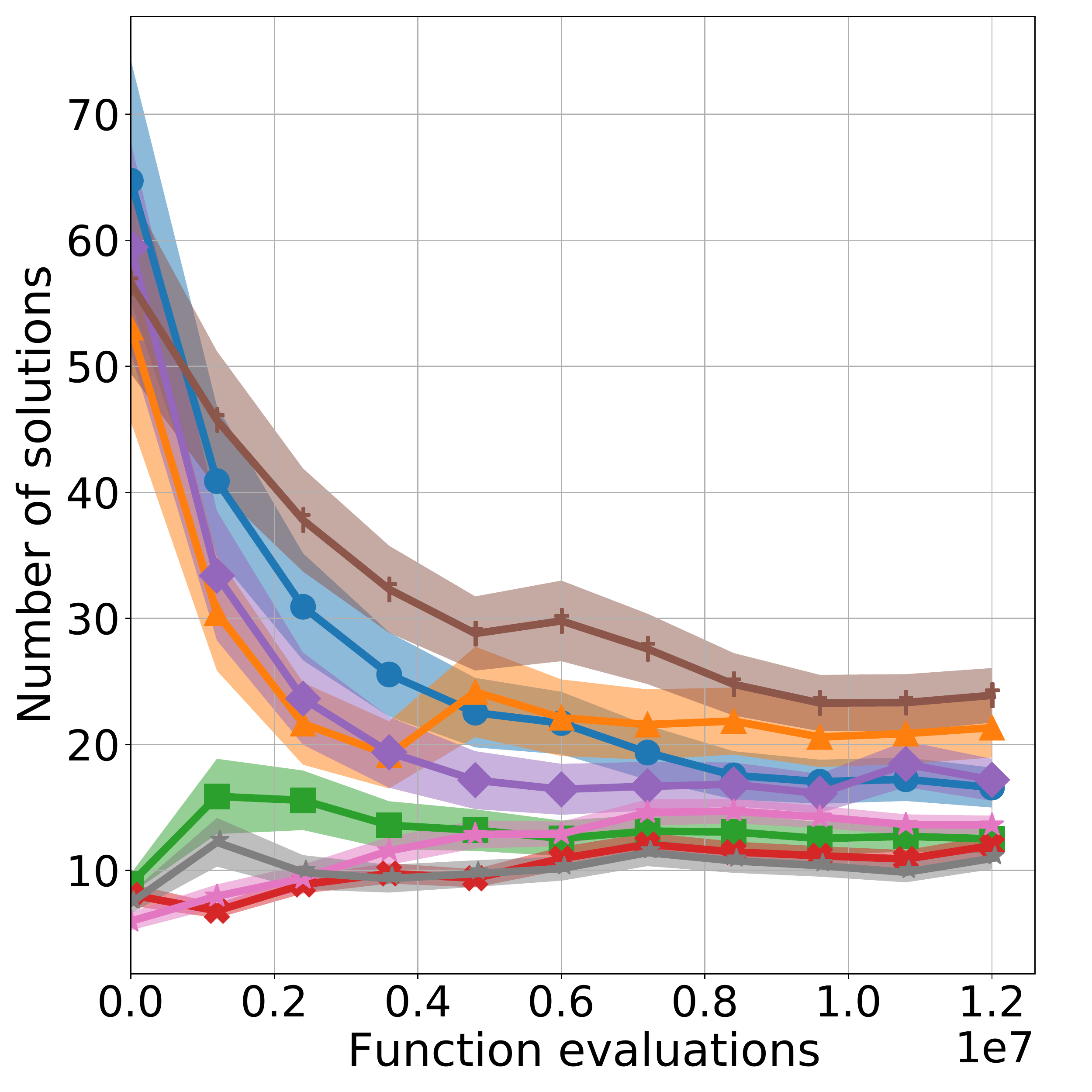}} 
\subfloat[Average fitness]{\includegraphics[width=0.30\linewidth]{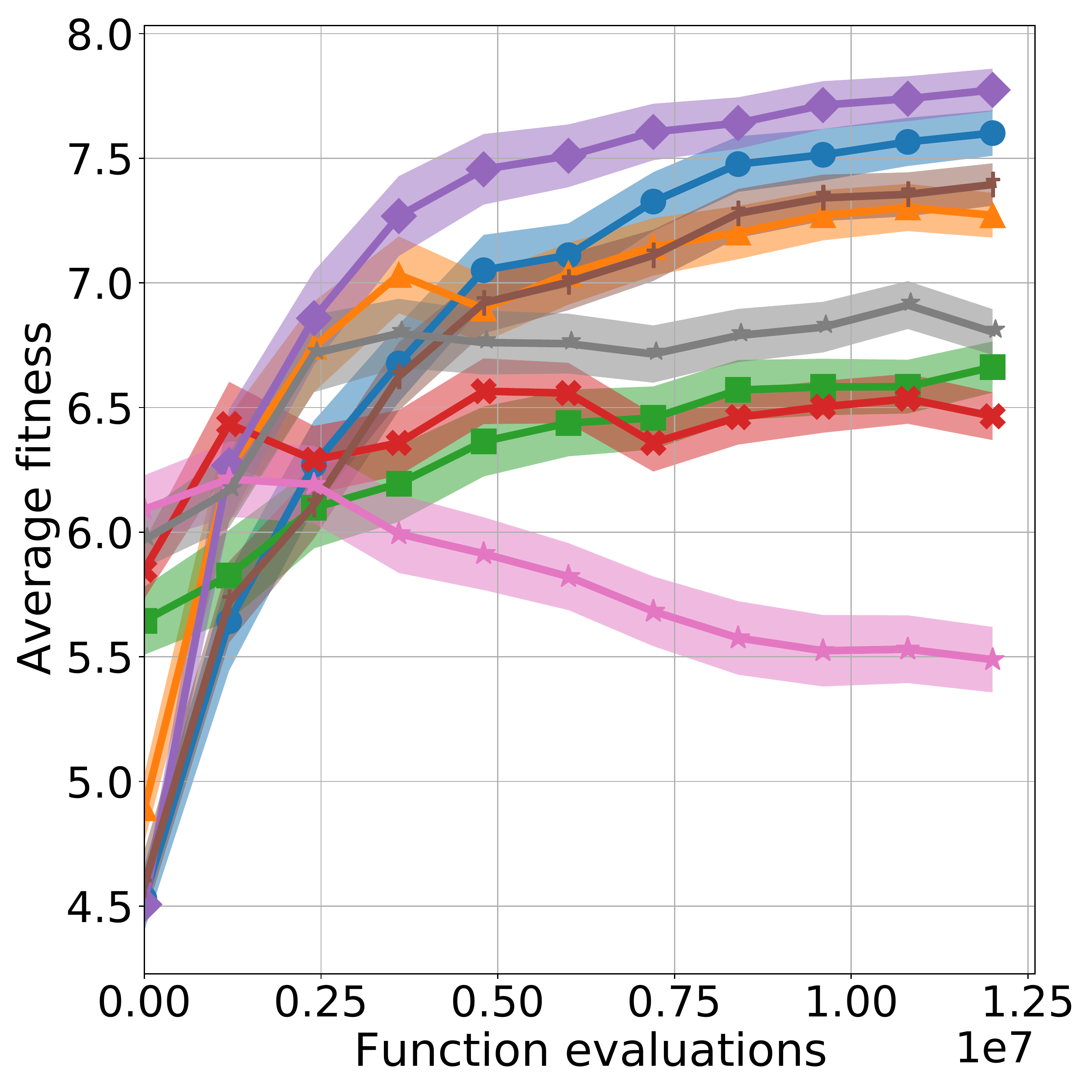}} 
\subfloat[Global fitness]{\includegraphics[width=0.30\linewidth]{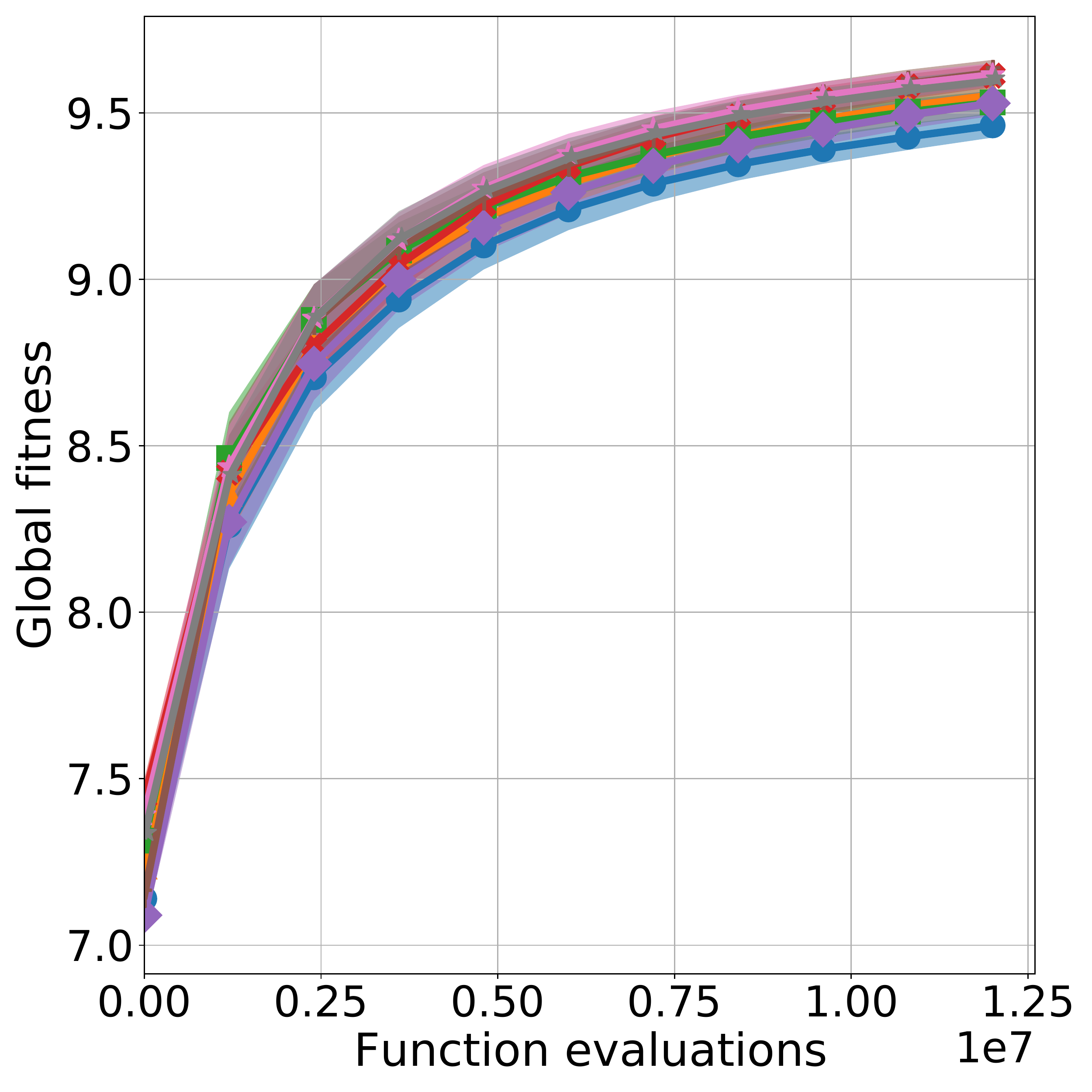}} 
\caption{\small{Quality-diversity statistics (Mean $\pm$ SE) of the different QD-Meta conditions across 4 replicates on the RHex robot platform, including \textbf{(a)} the total number of solutions in the archive; \textbf{(b)} the average fitness across the archive; and \textbf{(c)} the maximal fitness across the archive. Dynamic parameter control with RL is used by default while the suffix \textbf{static} indicates that no dynamic parameter control is used.}} \label{fig: QD-stats-meta}
\end{figure*}

\begin{figure*}[htbp!]
\hspace{1.5cm} 
\includegraphics[width=0.40\linewidth]{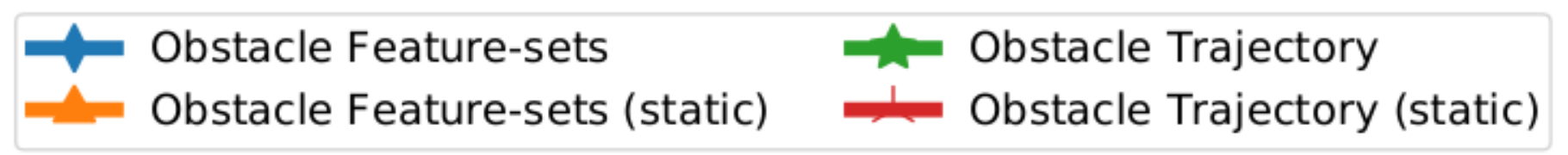} \hspace{1.5cm}
\includegraphics[width=0.40\linewidth]{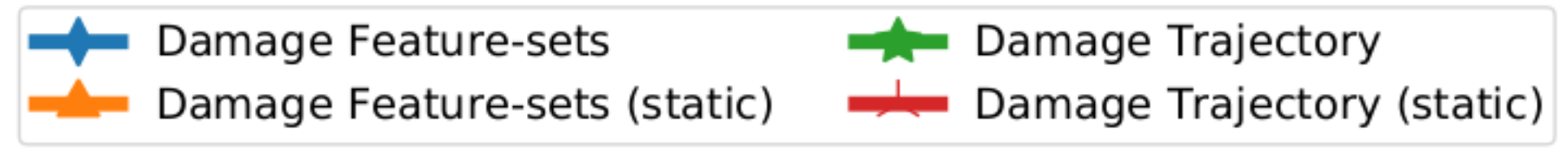} \\
\centering
\subfloat[Obstacle]{\includegraphics[width=0.46\linewidth]{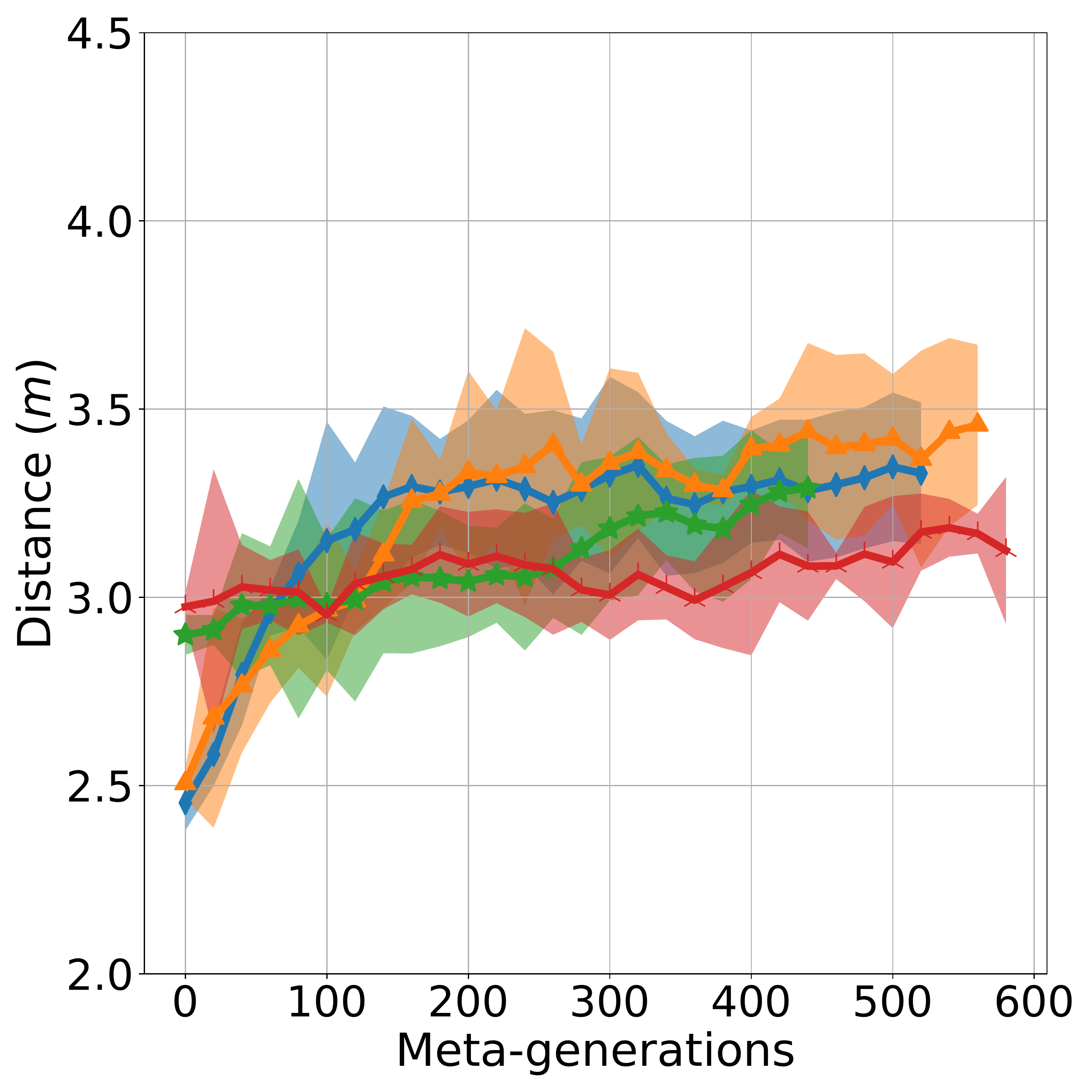}}
\subfloat[Damage]{\includegraphics[width=0.46\linewidth]{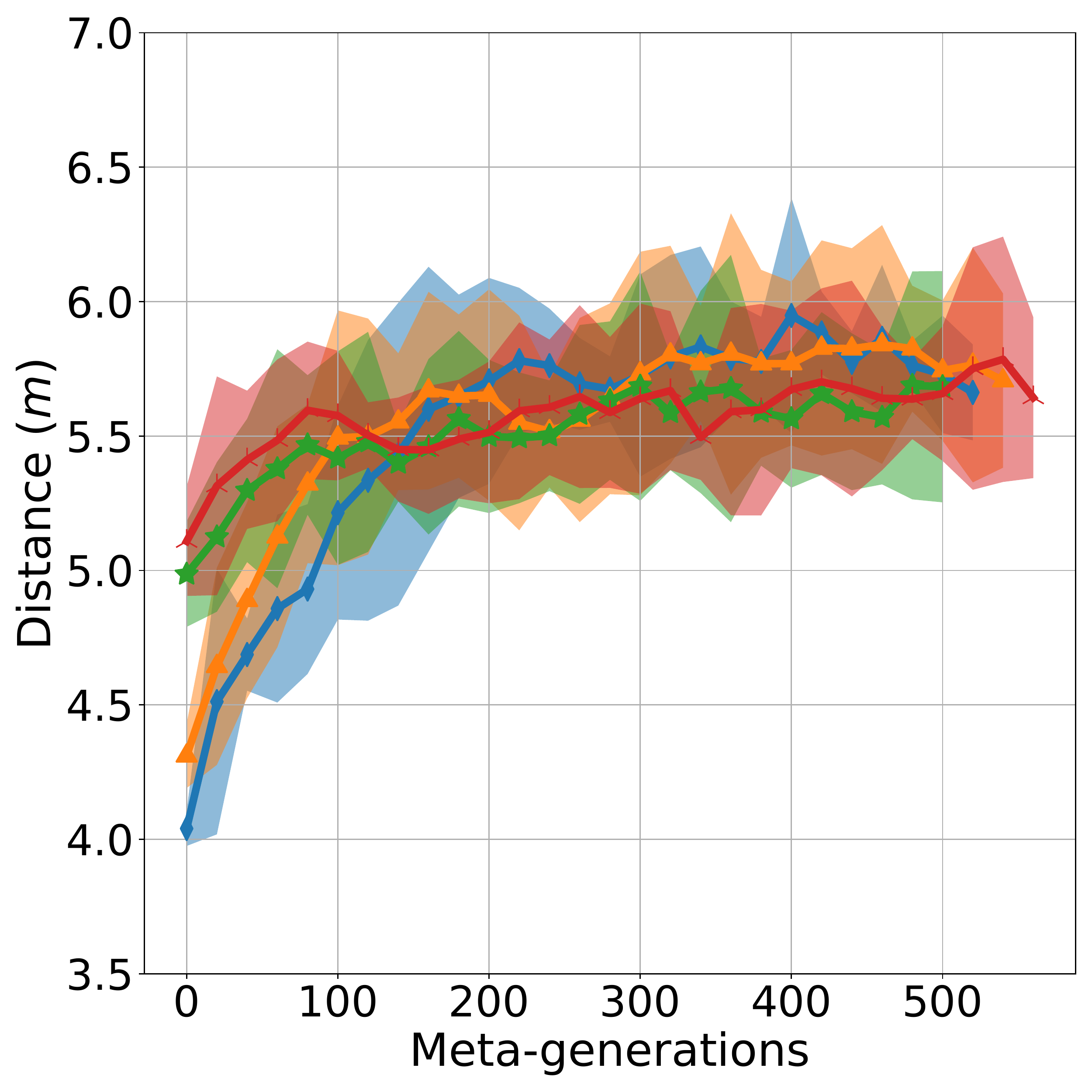}}
\caption{\small{Meta-fitness (Mean $\pm$ SE) evolution of different QD-Meta conditions over meta-generations. $x$-axis is the number of meta-generations and $y$-axis is the meta-fitness average across the archives in the meta-population, with the mean and standard error being aggregated across 4 replicates. The meta-fitness represents a change of environment in the RHex robot platform: in the QD-Meta Obstacle condition (left), the archive is assessed on various obstacle courses; in the QD-Meta Damage condition (right), the archive is assessed on different kinds of damages injected to the robot's legs one-by-one. To better display the trend over time, the meta-fitness is further smoothed over time by computing the running average.  Dynamic parameter control with RL is used by default while the suffix \textbf{static} indicates that no dynamic parameter control is used.}} \label{fig: meta-fitness}
\end{figure*}

\begin{figure*}[htbp!]
\centering
\includegraphics[width=0.22\linewidth]{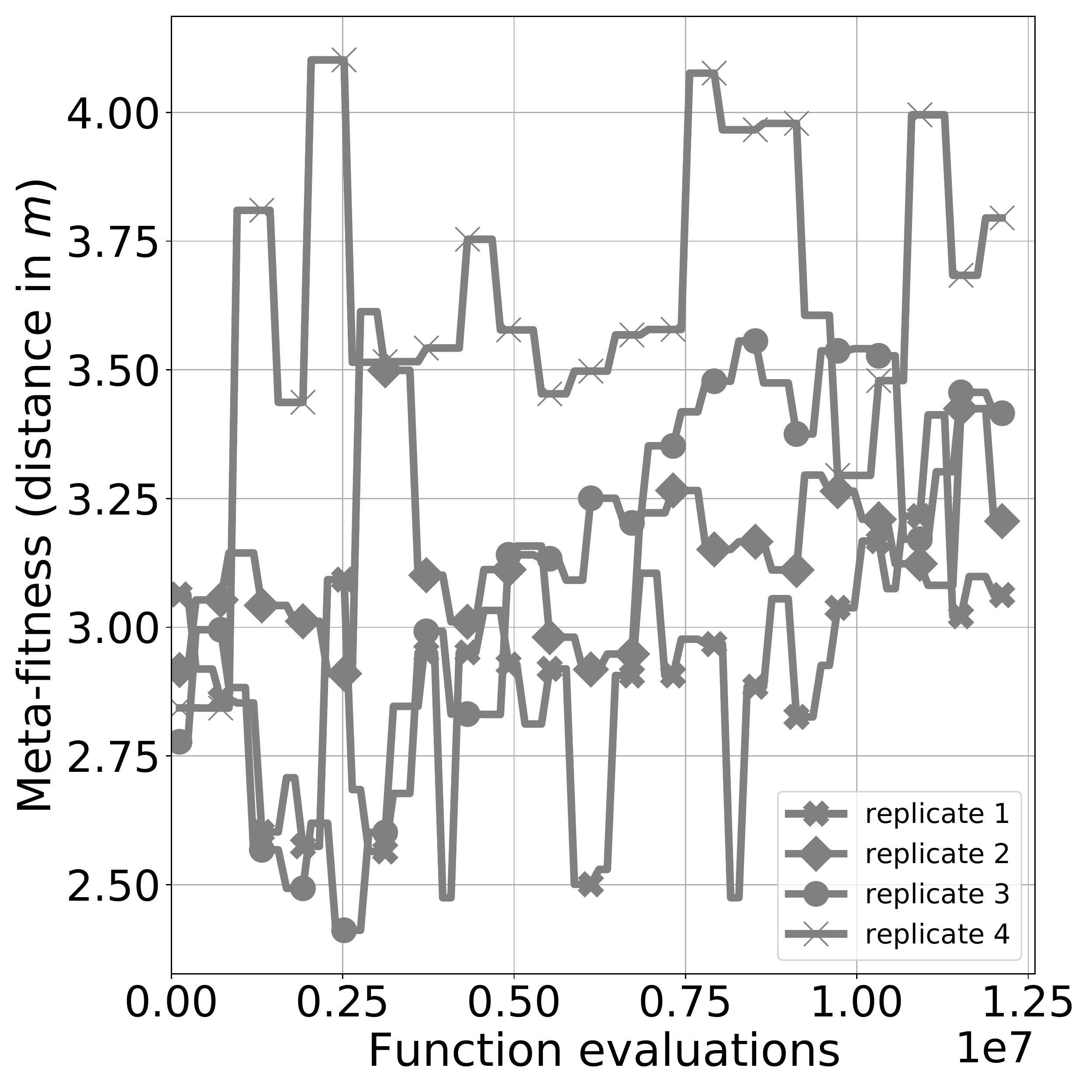}
\includegraphics[width=0.22\linewidth]{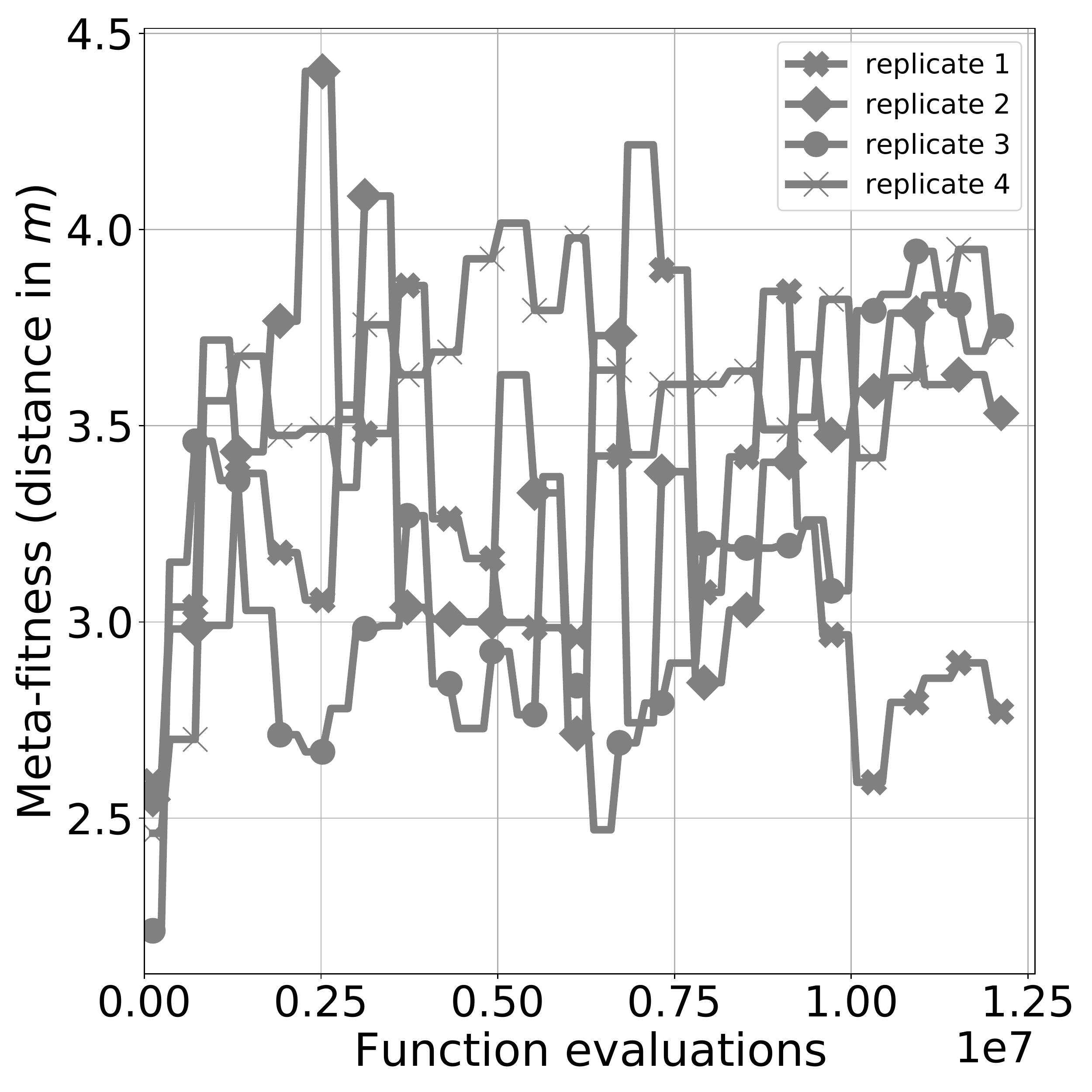}
\includegraphics[width=0.22\linewidth]{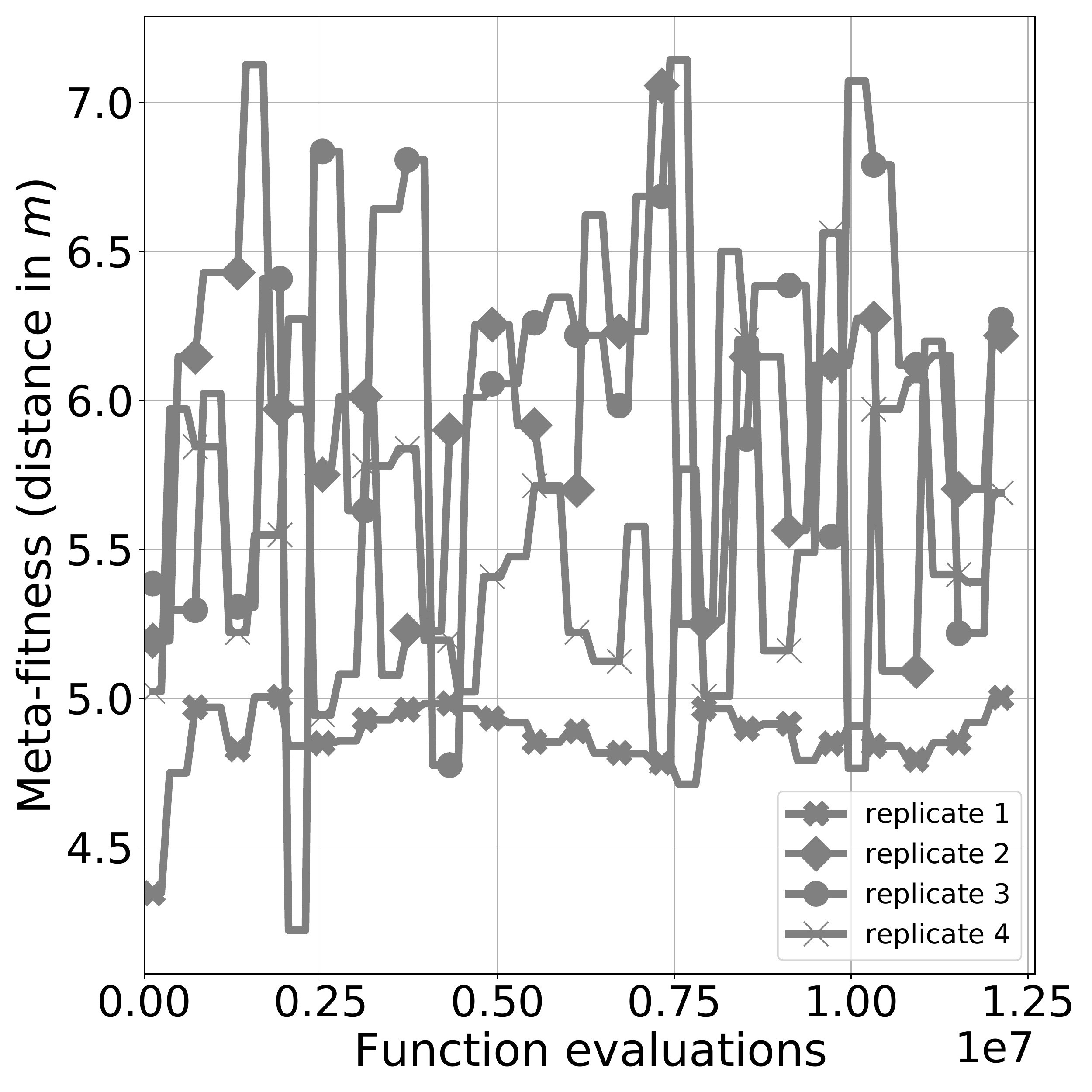}
\includegraphics[width=0.22\linewidth]{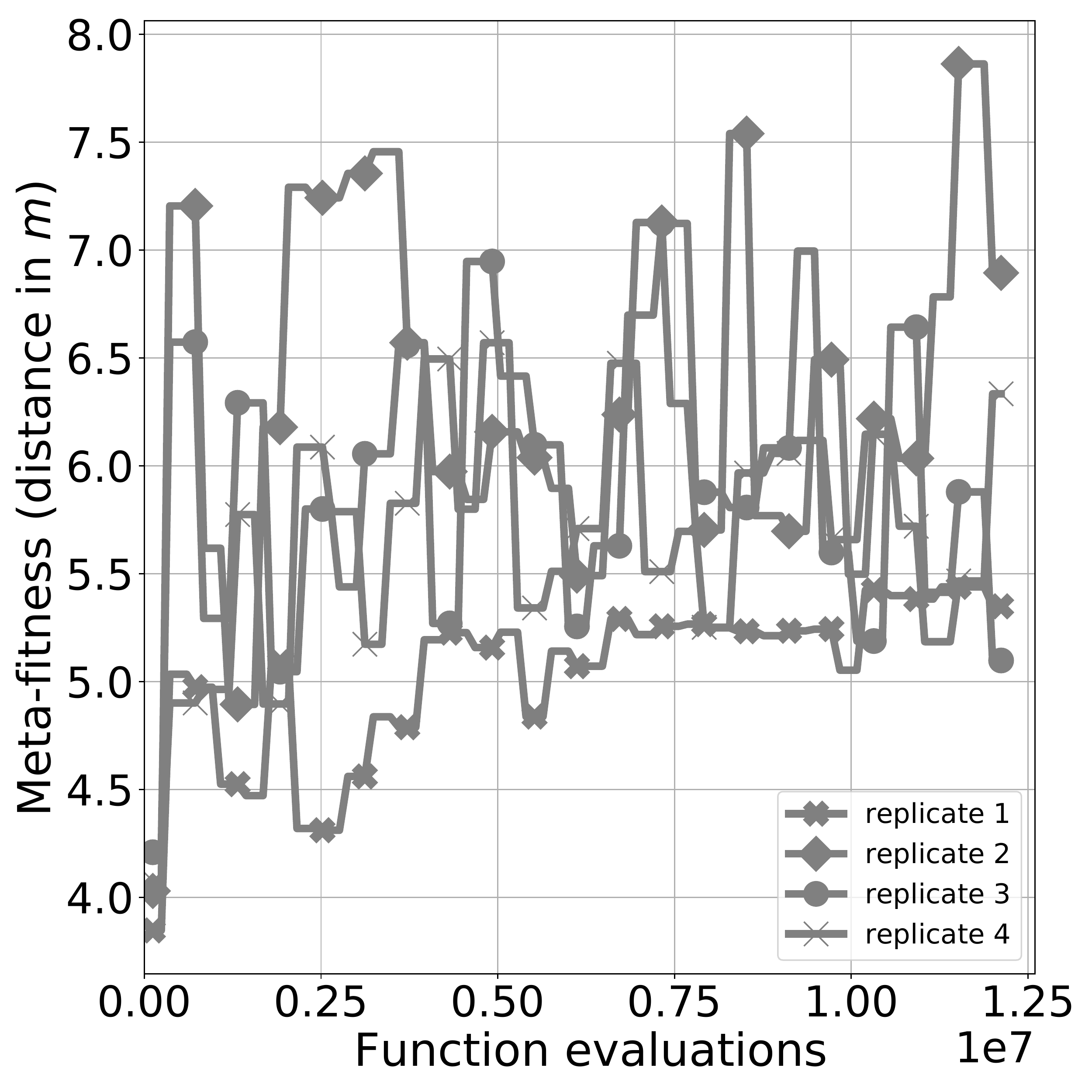} \\
\subfloat[Obstacle Trajectory]{\includegraphics[width=0.22\linewidth]{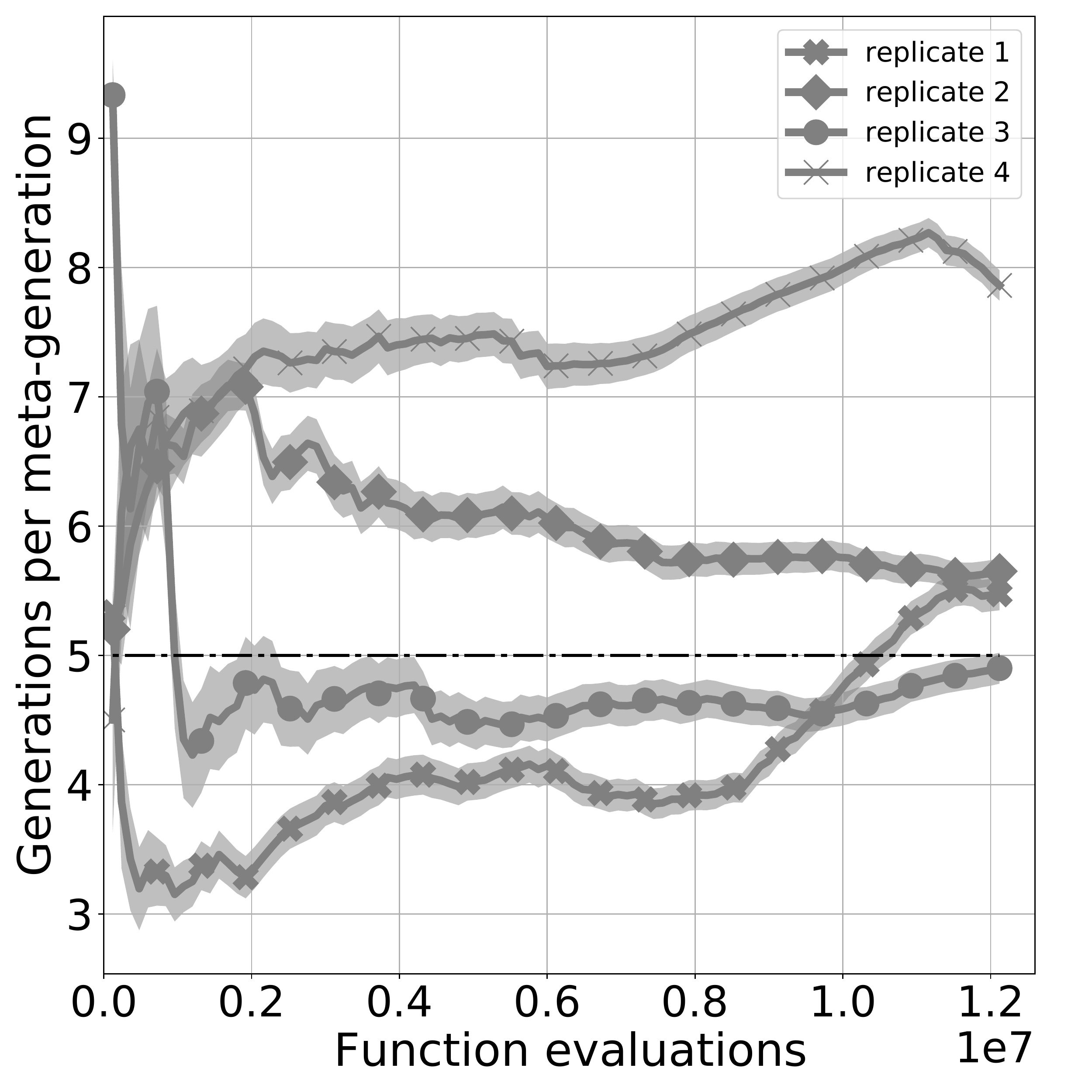}}
\subfloat[Obstacle Feature-sets]{\includegraphics[width=0.22\linewidth]{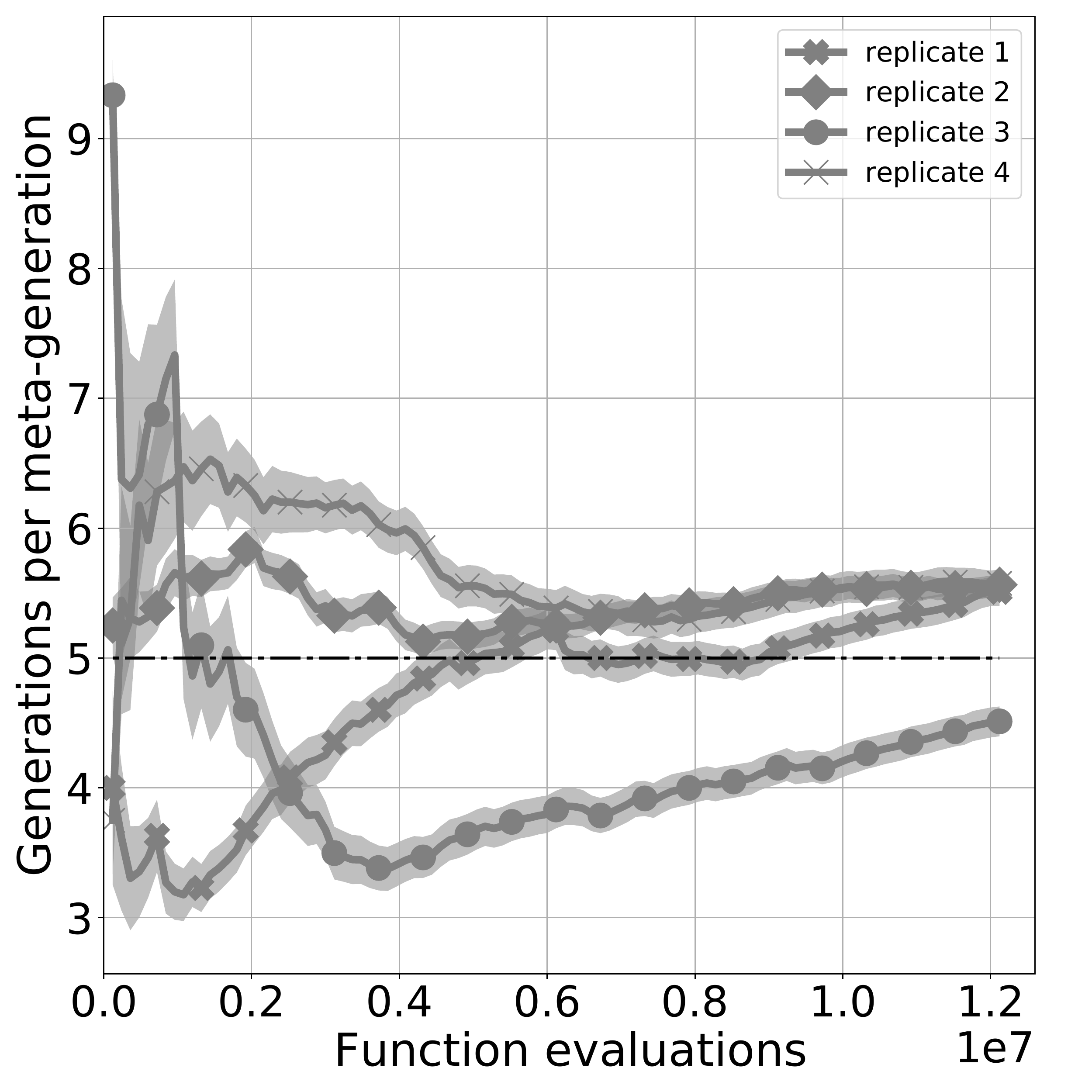}}
\subfloat[Damage Trajectory]{\includegraphics[width=0.22\linewidth]{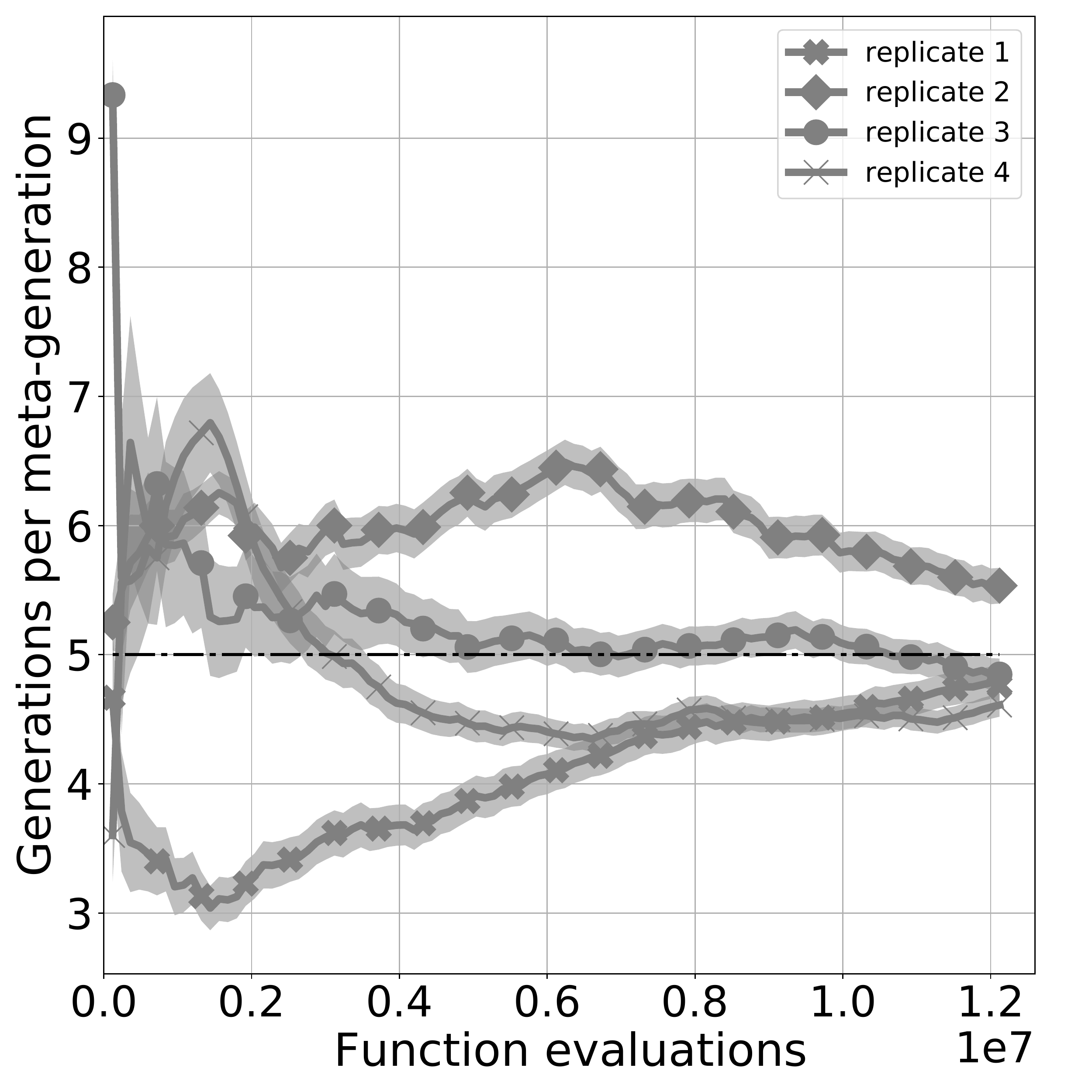}}
\subfloat[Damage Feature-sets]{\includegraphics[width=0.22\linewidth]{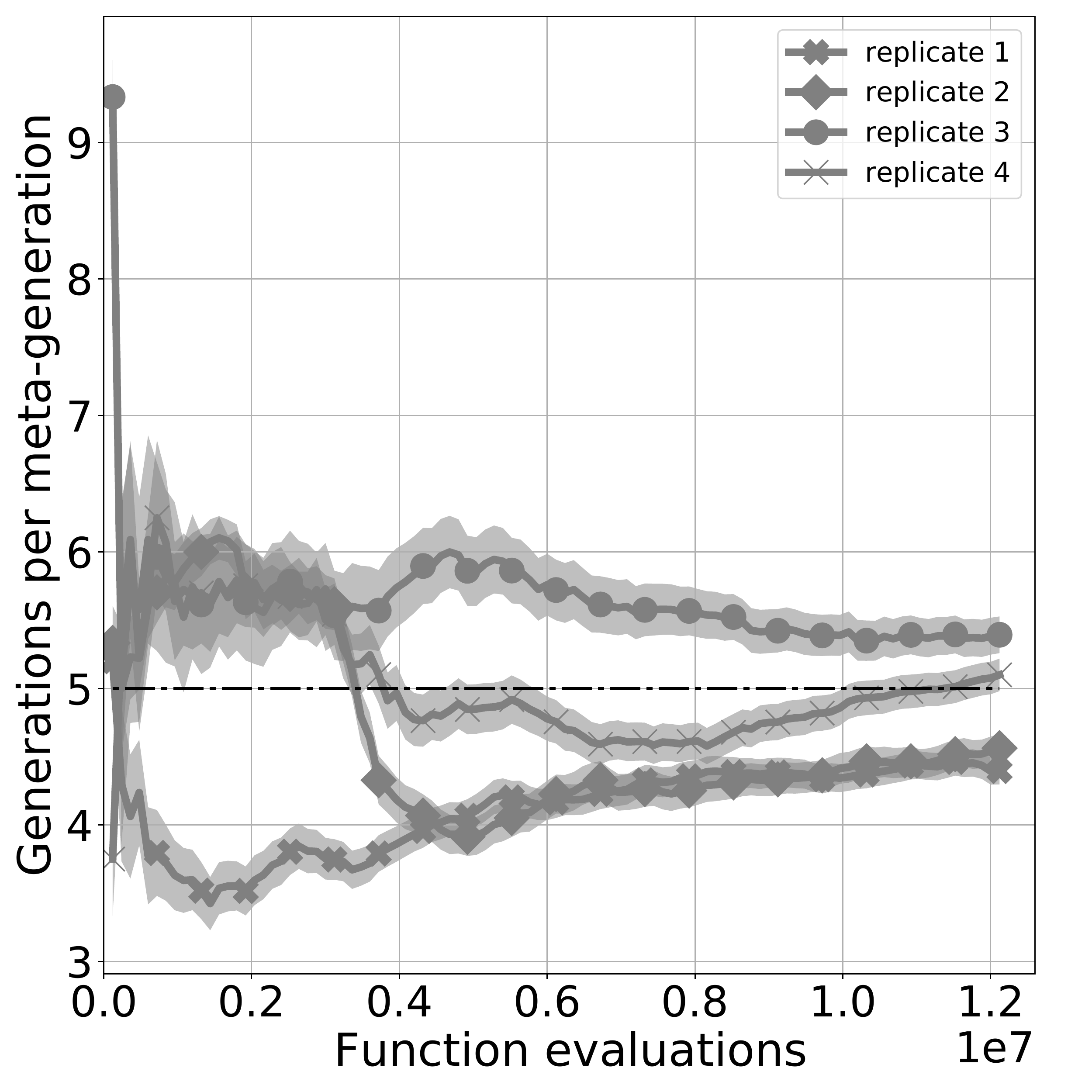}}
\caption{\small{The relation between the dynamically controlled generations per meta-generation parameter and the meta-fitness is shown for different QD-Meta conditions on the Rhex hexapod robot locomotion benchmark. Top panels show for each replicate the average meta-fitness across the meta-population while bottom panels show the corresponding changes to the number of generations per meta-generation (Mean $\pm$ SE aggregated over the meta-generations within the time bin).}} \label{fig: parameter-control}
\end{figure*}

Below results provide further results on the effect of the RL-based dynamic parameter control of the number of generations per meta-generation. This includes the QD-metrics (see Fig.~\ref{fig: QD-stats-meta}), the meta-fitness evolution (see Fig.~\ref{fig: meta-fitness}), and the relation of the number of generations per meta-generation with the meta-fitness over time (see Fig.~\ref{fig: parameter-control}).

\bibliography{library} 
\bibliographystyle{IEEEtran}